\documentclass[letterpaper,twoside,twocolumn,final]{IEEEtran}
\usepackage[noadjust]{cite}

\usepackage[dvips]{graphicx}
\usepackage{multirow,multicol}

\usepackage[tbtags]{amsmath}
\usepackage{amsbsy}
\usepackage{amssymb}
\usepackage{amsfonts}
\usepackage{bbm}

\usepackage{graphicx}
\usepackage{subfig}
\usepackage{pstricks}
\usepackage{pst-node}
\usepackage{pstricks-add}
\usepackage{url}
\usepackage{textcomp}
\usepackage{algorithmic}
\usepackage{algorithm}
\usepackage{balance}

\newtheorem{remark}{Remark}

\newtheorem{example}{Example}

{\begin{list}               
    {$\bullet$ \hfill}{
        \setlength{\leftmargin}{\parindent}
        \setlength{\parsep}{0.04\baselineskip}
        \setlength{\itemsep}{0.5\parsep}
        \setlength{\labelwidth}{\leftmargin}
        \setlength{\labelsep}{0em}}
    }
{\end{list}}

\providecommand{\eref}[1]{\eqref{#1}}  
\providecommand{\cref}[1]{Chapter~\ref{#1}}

\providecommand{\fref}[1]{Figure~\ref{#1}}
\providecommand{\tref}[1]{Table~\ref{#1}}

\providecommand{\R}{\ensuremath{\mathbb{R}}}

\providecommand{\E}{\ensuremath{\mathbb{E}}}

\providecommand{\inprod}[1]{\langle#1\rangle}

\providecommand{\bydef}{\overset{\text{def}}{=}}

\renewcommand{\vec}[1]{\ensuremath{\boldsymbol{#1}}}
\providecommand{\mat}[1]{\ensuremath{\boldsymbol{#1}}}

\providecommand{\calF}{\mathcal{F}}

\providecommand{\calL}{\mathcal{L}}

\providecommand{\mA}{\mat{A}}
\providecommand{\mB}{\mat{B}}

\providecommand{\mD}{\mat{D}}

\providecommand{\mI}{\mat{I}}

\providecommand{\mS}{\mat{S}}

\providecommand{\mU}{\mat{U}}

\providecommand{\mW}{\mat{W}}
\providecommand{\mX}{\mat{X}}
\providecommand{\mY}{\mat{Y}}

\providecommand{\va}{\vec{a}}
\providecommand{\vb}{\vec{b}}

\providecommand{\ve}{\vec{e}}

\providecommand{\vr}{\vec{r}}

\providecommand{\vu}{\vec{u}}
\providecommand{\vv}{\vec{v}}
\providecommand{\vw}{\vec{w}}
\providecommand{\vx}{\vec{x}}
\providecommand{\vy}{\vec{y}}
\providecommand{\vz}{\vec{z}}

\providecommand{\mLambda}{\mat{\Lambda}}

\providecommand{\mPhi}{\mat{\Phi}}

\providecommand{\valpha}{\vec{\alpha}}

\providecommand{\vvarphi}{\vec{\varphi}}

\providecommand{\vwtilde}{\boldsymbol{\widetilde{w}}}

\providecommand{\vzero}{\vec{0}}

\providecommand{\Var}{\mathrm{Var}}

\newcommand{\subjectto}{\mathop{\mathrm{subject\, to}}}
\newcommand{\argmin}[1]{\mathop{\underset{#1}{\mbox{argmin}}}}

\newcommand{\minimize}[1]{\mathop{\underset{#1}{\mathrm{minimize}}}}

\input{customFigureMacros}
\title{Depth Reconstruction from Sparse Samples: Representation, Algorithm, and Sampling}
\author{Lee-Kang~Liu,~\IEEEmembership{Student Member,~IEEE}, Stanley~H.~Chan,~\IEEEmembership{Member,~IEEE} and Truong~Q.~Nguyen,~\IEEEmembership{Fellow,~IEEE}
\thanks{L. Liu and T. Nguyen are with Department of Electrical and Computer Engineering, University of California at San Diego, La Jolla, CA 92093, USA. Emails: l7liu@ucsd.edu and tqn001@eng.ucsd.edu}
\thanks{S. Chan is with School of Electrical and Computer Engineering and Department of Statistics, Purdue University, West Lafayette, IN 47907, USA. Email: stanleychan@purdue.edu}
\thanks{This work was supported in part by the National Science Foundation under grant CCF-1065305. Preliminary material in this paper was presented at the 39th IEEE International Conference on Acoustics, Speech and Signal Processing (ICASSP), Florence, May 2014.}
}
\graphicspath{{pix/}}
\begin{document}
\maketitle

\begin{abstract}
The rapid development of 3D technology and computer vision applications have motivated a thrust of methodologies for depth acquisition and estimation. However, most existing hardware and software methods have limited performance due to poor depth precision, low resolution and high computational cost. In this paper, we present a computationally efficient method to recover dense depth maps from sparse measurements. We make three contributions. First, we provide empirical evidence that depth maps can be encoded much more sparsely than natural images by using common dictionaries such as wavelets and contourlets. We also show that a combined wavelet-contourlet dictionary achieves better performance than using either dictionary alone. Second, we propose an alternating direction method of multipliers (ADMM) for depth map reconstruction. A multi-scale warm start procedure is proposed to speed up the convergence. Third, we propose a two-stage randomized sampling scheme to optimally choose the sampling locations, thus maximizing the reconstruction performance for any given sampling budget. Experimental results show that the proposed method produces high quality dense depth estimates, and is robust to noisy measurements. Applications to real data in stereo matching are demonstrated.
\end{abstract}


\section{Introduction}
The rapid development of 3D technology has created a new wave of visualization and sensing impacts to the digital signal processing community. From remote sensing \cite{Lefsky_Cohen_Parker_2002} to preserving historical heritages \cite{Agarwal_Snavely_Simon_2009}, and from rescue \cite{Burion_2004} to 3D laparoscopic surgery \cite{Khoshabeh_Juang_Talamini_2012,Chan_Khoshabeh_Gibson_2011}, the footprints of 3D have been influencing a broad spectrum of the technological frontiers.

The successful development of 3D signal processing is fundamentally linked to a system's ability to acquire depth. To date, there are two major classes of depth acquisition techniques: hardware solutions and computational procedures. Hardware devices are usually equipped with active sensors such as time-of-flight camera \cite{Foix_Alenya_Torras_2011} and LiDAR \cite{Schwarz_2010}. While being able to produce high quality depth maps, these hardware systems have high instrumentation cost. Moreover, the data acquisition time of the devices is long (10 fps as opposed to 60fps on standard cameras \cite{Niclass_Soga_Matsubara_Kato_Kagami_2013}). Although speeding up is possible, spatial resolution has to be traded off in return.

An alternative solution to acquiring depth is to estimate depth using a set of computational procedures. This class of computational methods, broadly referred to as disparity estimation algorithms \cite{Mei_Sun_Zhou_2011,Klaus_Sormann_Karner_2006,Wang_Zheng_2008,Yang_Wang_Yang_2009}, estimates the depth by computing the disparities between a pair of stereo images through their corresponding matching features \cite{Heikkila_Silven_1997,Zhang_1999}. Disparity estimation algorithms usually work well under well conditioned environments, but they could be sensitive to illumination, noise, stereo camera alignments, and other camera factors. Thus, the effective number of reliable features that one can use for disparity estimation is actually much fewer than the number of pixels of the image \cite{Feris_Gemmell_Toyama_2002,Ke_Sukthankar_2004}.

\subsection{Scope and Contributions}
The objective of this paper is to present a sampling and reconstruction framework to improve and speed up the depth acquisition process. The key idea is to carefully select a sparse subset of spatial samples and use an optimization algorithm to reconstruct the final dense depth map.

The three major contributions of this paper are as follows.

\emph{1) Representation} (Section III). In order to reconstruct the depth map, we must first define an appropriate representation. We show that, as opposed to natural images, depth maps can be well approximated using a sparse subset of wavelet atoms. Moreover, we show that a combined dictionary of wavelets and contourlets can further improve the reconstruction quality.

\emph{2) Algorithm} (Section IV). We propose a fast numerical algorithm based on the alternating direction method of multipliers (ADMM). We derive novel splitting strategies that allow one to solve a sequence of parallelizable subproblems. We also present a multiscale implementation that utilizes the depth structures for efficient warm starts.

\emph{3) Sampling} (Section V). We propose an efficient spatial sampling strategy that maximizes the reconstruction performance. In particular, we show that for a fixed sampling budget, a high quality sampling pattern can be obtained by allocating random samples with probabilities in proportional to the magnitudes of the depth gradients.

\subsection{Related Work}
The focus of this paper lies in the intersection of two closely related subjects: depth enhancement and compressed sensing. Both subjects have a rich collection of prior works but there are also limitations which we should now discuss.

The goal of depth enhancement is to improve the resolution of a depth map. Some classical examples include Markov Random Field (MRF) \cite{Diebel_Thrun_2005}, bilateral filter \cite{Yang_Yang_Davis_Nister_2007}, and other approaches \cite{Li_Xue_Sun_2012,Aodha_Campbell_Nair_2012}. One limitation of these methods is that the low-resolution depth maps are sampled uniformly. Also, it is usually assumed that a color image of the scene is available. In contrast, our proposed method is applicable to \emph{any} non-uniformly sampled low-resolution depth map and does not require color images. Thus, the new method allows for a greater flexibility for the enhancement.

Compressed sensing (CS) is a popular mathematical framework for sampling and recovery \cite{Candes_Wakin_2008}. In many cases, CS methods assume that \emph{natural images} exhibit sparse structures in certain domains, \emph{e.g.}, wavelet. However, as will be discussed in Section III of this paper, natural images are indeed \emph{not} sparse. If we compare natural images to depth maps, the latter would show a much sparser structure than the former. Furthermore, the theory of combined bases \cite{Donoho_Huo_2001,Elad_Bruckstein_2002} shows that a pair of incoherent bases are typically more effective for signal recovery. Yet, the application of these theories to depth maps is not fully explored.

The most relevant paper to our work is perhaps \cite{Hawe_Kleinsteuber_Diepold_2011}. However, our work has two advantages. First, we propose a new ADMM algorithm for the reconstruction task (Section IV). We show that the ADMM algorithm is significantly more efficient than the subgradient method proposed in \cite{Hawe_Kleinsteuber_Diepold_2011}. Second, we present a sampling scheme to choose optimal sampling patterns to improve the depth reconstruction (Section V), which was not discussed in \cite{Hawe_Kleinsteuber_Diepold_2011}.

We should also mention a saliency-guided CS method proposed in \cite{Schwartz_Wong_Clausi_2013, Schwartz_Wong_Clausi_2012}. In these two papers, the spatial sampling is achieved by a mixing-plus-sampling process, meaning that the unknown pixels are filtered and then sub-sampled. The filtering coefficients are constructed through a pre-defined saliency map and certain density functions (\emph{e.g.}, Gaussian-Bernoulli). In our work, the mixing process is \emph{not} required so that depth values are sampled without filtering. This makes our proposed method applicable to disparity estimation where mixing cannot be used (otherwise it will defeat the purpose of reconstructing dense depth maps from a few estimated values.)

Finally, advanced computational photography techniques are recently proposed for fast depth acquisition, \emph{e.g.}, \cite{Kirmani_Colaco_Wong_2012, Kirmani_Venkatraman_Shin_2013}. However, the problem settings of these works involve hardware designs and are thus different from this paper.

The rest of the paper is organized as follows. After elaborating the problem and clarifying notations in Section II, we discuss the representation of depth maps in Section III. A fast reconstruction algorithm is presented in Section IV. In Section V we discuss the design of optimal sampling patterns. Experimental results are shown in Section VI, and a concluding remark is given in Section VII. 
\section{Notations and Problem Formulation}
In this section we introduce notations and elaborate on the problem formulation.

\subsection{Depth and Disparity}
The type of data that we are interested in studying is the depth map. Depth can be directly measured using active sensors, or inferred from the disparity of a pair of stereo images. Since the correspondence between depth and disparity is unique by simple geometry \cite{Hartley_2004}, in the rest of the paper we shall use depth and disparity interchangeably.

\subsection{Sampling Model}
Let $\vx \in \R^{N}$ be an $N \times 1$ vector representing a disparity map. For simplicity we assume that $\vx$ is normalized so that $0 \le x_j \le 1$ for $j=1,\ldots,N$.

To acquire a set of spatial samples, we define a diagonal matrix $\mS \in \R^{N \times N}$ with the $(j,j)$th entry being
\begin{equation}
S_{jj} \bydef
\begin{cases}
1, &\quad \mbox{ with probability $p_j$,}\\
0, &\quad \mbox{ with probability $1-p_j$,}
\end{cases}
\label{eq:Sii}
\end{equation}
where $\{p_j\}_{j=1}^N$ is a sequence of pre-defined probabilities. Specific examples of $\{p_j\}_{j=1}^N$ will be discussed below. For now, we only require $\{p_j\}_{j=1}^N$ to satisfy two criteria: (1) for each $j = 1,\ldots,N$, $p_j$ must be bounded so that $0 \le p_j \le 1$; (2) the average of the probabilities must achieve a target \emph{sampling ratio} $\xi$:
\begin{equation}
\frac{1}{N}\sum_{j=1}^N p_j = \xi,
\end{equation}
where $0 < \xi < 1$. 

\begin{example}
If $p_j = \xi$ for all $j$, then the sampling pattern $\mS$ is a diagonal matrix with uniformly random entries. This sampling pattern corresponds to a uniform sampling without filtering in the classical compressed sensing, e.g., \cite{Candes_Wakin_2008}.
\end{example}

\begin{example}
If $p_j = 1$ for $j \in \Omega_1$ and $p_j = 0$ for $j \in \Omega_0$, where $\Omega_1$ and $\Omega_0$ are two pre-defined sets such that $|\Omega_1| = \xi N$ and $|\Omega_0| = (1-\xi)N$, then $\mS$ is a deterministic sampling pattern. In particular, if $\Omega_1$ and $\Omega_0$ are designed so that the indices are uniformly gridded, then $\mS$ will become the usual down-sampling operator.
\end{example}

With $\mS$, we define the sampled disparity map as
\begin{equation}
\vb = \mS\vx.
\end{equation}
Note that according to our definition of $\mS$, the sampled disparity $\vb \in \R^{N \times 1}$ will contain zeros, \emph{i.e.}, $b_j = 0$ if $S_{jj}=0$. Physically, this corresponds to the situation where the unsampled pixels are marked with a value of zero.

\begin{remark}
Since $\mS$ is a random diagonal matrix, readers at this point may have concerns about the overall number of samples which is also random. However, we argue that such randomness has negligible effects for the following reason. For large $N$, standard concentration inequality guarantees that the average number of ones in $\mS$ stays closely to $\xi N$. In particular, by Bernstein's inequality \cite{Chung_Lu_2006} we can show that for $\varepsilon > 0$,
\begin{equation}
\Pr\left( \left|\frac{1}{N}\sum_{j=1}^N S_{jj} - \xi \right| > \varepsilon \right) \le 2\exp\left\{-\frac{N\varepsilon^2}{1/2 + 2\varepsilon/3}\right\}.
\end{equation}
Therefore, although the sampling pattern in our framework is randomized, the average number of samples is concentrated around $\xi N$ for large $N$.
\end{remark}

\subsection{Representation Model}
To properly formulate the reconstruction problem, we assume that the disparity map can be efficiently represented as a linear combination of basis vectors $\{\vvarphi_i\}_{i=1}^M$:
\begin{equation}
\vx = \sum_{i=1}^M \inprod{\vx,\vvarphi_i}\vvarphi_i,
\label{eq:linear combination}
\end{equation}
where $\inprod{\cdot,\cdot}$ denotes the standard inner product. Defining $\alpha_i \bydef \inprod{\vx,\vvarphi_i}$ as the $i$th basis coefficient, $\valpha \bydef [\alpha_1,\ldots,\alpha_M]^T$, and $\mPhi \bydef [\vvarphi_1,\ldots,\vvarphi_M]$, the relationship in \eref{eq:linear combination} can be equivalently written as $\vx = \mPhi\valpha$.

The reconstruction problem can be posed as an optimization problem in which the goal is to seek a sparse vector $\valpha \in \R^M$ such that the observed samples $\vb$ are best approximated. Mathematically, we consider the problem
\begin{equation}
\minimize{\valpha} \;\; \frac{1}{2}\| \mS \mPhi \valpha - \vb \|_2^2 + \lambda \|\valpha\|_1,
\label{eq:problem P0}
\end{equation}
where $\lambda > 0$ is a regularization parameter, and $\|\cdot\|_{1}$ is the $\ell_1$-norm of a vector.

In this paper, we are mainly interested in two types of $\mPhi$ --- the wavelet frame and the contourlet frame \cite{Do_Vetterli_2005}. Frames are  generalizations of the standard bases in which $M$, the number of bases, can be more than $N$, the dimension of $\vx$. Moreover, for any frame $\mPhi$, it holds that $\mPhi\mPhi^T = \mI$. Therefore, $\vx = \mPhi\valpha$ if and only if $\valpha = \mPhi^T\vx$. Using this result, we can equivalently express \eref{eq:problem P0} as
\begin{equation}
\minimize{\vx} \;\; \frac{1}{2}\| \mS \vx - \vb \|_2^2 + \lambda \|\mPhi^T\vx\|_1.
\label{eq:problem P1}
\end{equation}

\begin{remark}
In compressed sensing literature, \eref{eq:problem P0} is known as the synthesis problem and \eref{eq:problem P1} is known as the analysis problem \cite{Elad_Milanfar_Rubinstein_2007}. Furthermore, the overall measurement matrix $\mS\mPhi$ in \eref{eq:problem P0} suggests that if $p_j = \xi$ for all $j$, then $\mS\mPhi$ corresponds to the partial orthogonal system as discussed in \cite{Candes_Plan_2011}. In this case, the restricted isometry property (RIP) holds \cite{Baraniuk_Davenport_DeVore_2008} and exact recovery can be guaranteed under appropriate assumptions of sparsity and number of measurements. For general $\{p_j\}_{j=1}^N$, establishing RIP is more challenging, but empirically we observe that the optimization produces reasonable solutions.
\end{remark}

\subsection{Penalty Functions}
As discussed in \cite{Hawe_Kleinsteuber_Diepold_2011}, \eref{eq:problem P1} is not an effective formulation because the $\ell_1$ norm penalizes \emph{both} the approximation (lowpass) and the detailed (highpass) coefficients. In reality, since disparity maps are mostly piecewise linear functions, the lowpass coefficients should be maintained whereas the highpass coefficients are desirable to be sparse. To this end, we introduce a binary diagonal matrix $\mW \in \R^{M \times M}$ where the $(j,j)$th entry is 0 if $j$ is an index in the lowest passband, and is 1 otherwise. Consequently, we modify the optimization problem as
\begin{equation}
\minimize{\vx} \;\; \frac{1}{2}\| \mS \vx - \vb \|_2^2 + \lambda \|\mW\mPhi^T\vx\|_1.
\label{eq:problem P2}
\end{equation}

Finally, it is desirable to further enforce smoothness of the reconstructed disparity map. Therefore, we introduce a total variation penalty so that the problem becomes
\begin{equation}
\minimize{\vx} \;\; \frac{1}{2}\| \mS \vx - \vb \|_2^2 + \lambda \|\mW\mPhi^T\vx\|_1 + \beta\|\vx\|_{TV}.
\label{eq:problem P3}
\end{equation}
Here, the total variation norm is defined as
\begin{equation}
\|\vx\|_{TV} \bydef \|\mD_x \vx\|_1 + \|\mD_y\vx\|_1,
\end{equation}
where $\mD = [\mD_x; \, \mD_y]$ is the first-order finite difference operator in the horizontal and vertical directions. The above definition of total variation is known as the anisotropic total variation. The same formulation holds for isotropic total variation, in which $\|\vx\|_{TV} = \sum_{j=1}^N  \sqrt{[\mD_x \vx]_j^2 + [\mD_y \vx]_j^2}$.

The problem in \eref{eq:problem P3} is generalizable to take into account of a combination of $L$ dictionaries. In this case, one can consider a sum of $L$ penalty terms as
\begin{equation}
\minimize{\vx} \;\; \frac{1}{2}\| \mS \vx - \vb \|_2^2 + \sum_{\ell=1}^L \lambda_{\ell} \|\mW_{\ell}\mPhi_{\ell}^T\vx\|_1 + \beta\|\vx\|_{TV}.
\label{eq:problem P4}
\end{equation}
For example, in the case of combined wavelet and contourlet dictionaries, we let $L = 2$.

\section{Sparse Representation of Disparity Map}
The choice of the dictionary $\mPhi$ in \eref{eq:problem P4} is an important factor for the reconstruction performance. In this section we discuss the general representation problem of disparity maps. We show that disparity maps can be represented more sparsely than natural images. We also show that a combined wavelet-contourlet dictionary is more effective in representing disparity maps than using the wavelet dictionary alone.

\subsection{Natural Images vs Depth Data}
Seeking effective representations for \emph{natural images} is a well-studied subject in image processing \cite{Aharon_Elad_Bruckstein_2006,Mairal_Elad_Sapiro_2008,Mallat_2008,Po_Do_2006, Do_Vetterli_2005,Candes_Donoho_2002, Candes_Donoho_2004,LePennec_Mallat_2005}. However, representations of \emph{disparity maps} seems to be less studied. For example, it is unclear how sparse can a predefined dictionary (\emph{e.g.}, wavelets) encode disparity maps as compared to natural images.

\begin{figure}[h]
\centering
\begin{tabular}{cccc}
\includegraphics[width=0.24\linewidth]{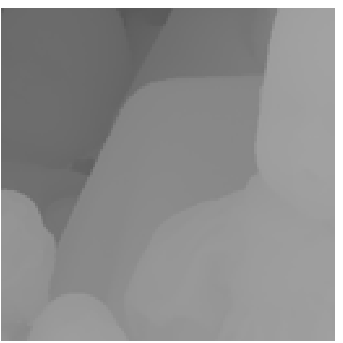}& \hspace{-0.5cm}
\includegraphics[width=0.24\linewidth]{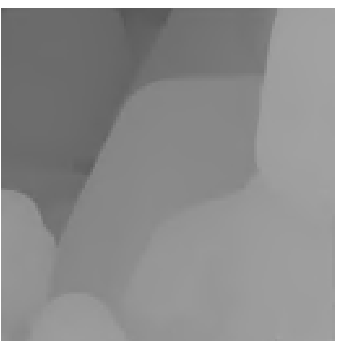}& \hspace{-0.5cm}
\includegraphics[width=0.24\linewidth]{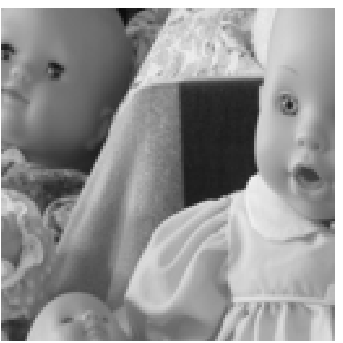}& \hspace{-0.5cm}
\includegraphics[width=0.24\linewidth]{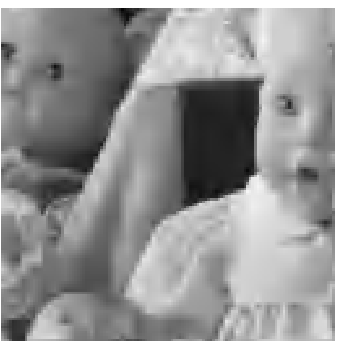}\\
(a) Original & \hspace{-0.5cm} (b) Approx. & \hspace{-0.5cm}(c) Original & \hspace{-0.5cm}(d) Approx. \\
 disparity & \hspace{-0.5cm} disparity & \hspace{-0.5cm} view & \hspace{-0.5cm} view\\
 &   \hspace{-0.5cm} (50.25 dB) & & \hspace{-0.5cm} (29.29 dB)
\end{tabular}
\caption{PSNR values of approximating a disparity patch and a image patch using the leading $5\%$ of the wavelet coefficients.}
\label{fig:wavelet disparity vs image}
\end{figure}

To address this question, we consider a $128 \times 128$ cropped patch from a gray-scaled image and the corresponding patch in the disparity map. For each of the image and the disparity, we apply the wavelet transform with Daubechies $5/3$ filter and 5 decomposition levels. Then, we truncate the wavelet coefficients to the leading $5 \%$ coefficients with the largest magnitudes. The reconstructed patches are compared and the results are shown in \fref{fig:wavelet disparity vs image}.

The result indicates that for the same number of wavelet coefficients, the disparity map can be synthesized with significantly lower approximation error than the image. While such result is not surprising, the big difference in the PSNRs provides evidence that reconstruction of disparity maps from sparse samples should achieve better results than that of natural images.

\subsection{Wavelet vs Contourlet}
The above results indicate that wavelets are efficient representations for disparity maps. Our next question is to ask if some of the dictionaries would do better than other dictionaries. In this section, we discuss how a combined wavelet-contourlet dictionary can improve the wavelet dictionary.

\subsubsection{Evaluation Metric}
To compare the performance of two dictionaries, it is necessary to first specify which metric to use. For the purpose of reconstruction, we compare the mean squared error (MSE) of the reconstructed disparity maps obtained by feeding different dictionaries into \eref{eq:problem P4}. For any fixed sampling pattern $\mS$, we say that a dictionary $\mPhi_1$ is better than another dictionary $\mPhi_2$ if the reconstruction result using $\mPhi_1$ has a lower MSE than using $\mPhi_2$, for the best choice of parameters $\lambda_1$, $\lambda_2$ and $\beta$. Note that in this evaluation we do not compare the sparsity of the signal using different dictionaries. In fact, sparsity is not an appropriate metric because contourlets typically require 33\% more coefficients than wavelets \cite{Vetterli_Kovacevic_1995}, but contourlets have a better representation of curves than wavelets.

\subsubsection{Comparison Results}
We synthetically create a gray-scaled image consisting of a triangle overlapping with an ellipse to simulate a disparity map. We choose the uniformly random sampling pattern $\mS$ so that there is no bias caused by a particular sampling pattern.

As parameters are concerned, we set $\lambda_1 = 4\times 10^{-5}$ and $\beta = 2\times 10^{-3}$ for the single wavelet dictionary model ($L=1$), and $\lambda_1 = 4\times 10^{-5}$, $\lambda_2 = 2\times 10^{-4}$ and $\beta=2\times 10^{-3}$ for the combined dictionary model ($L=2$). The choices of these parameters are discussed in Section IV-C.

\begin{figure}[h]
\centering
\includegraphics[width=0.95\linewidth]{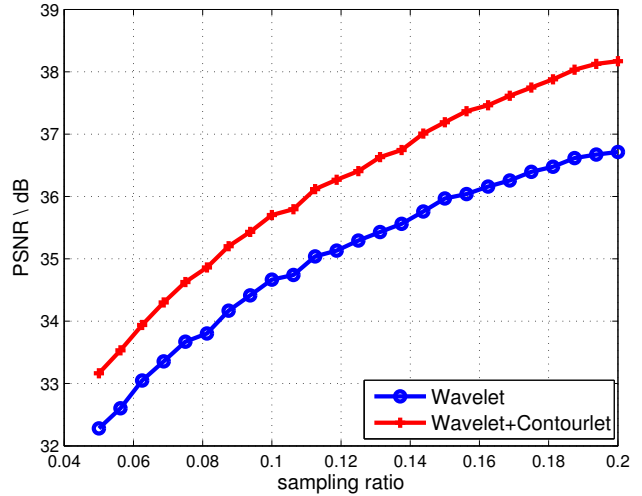}
\caption{ADMM reconstruction result as a function of sampling ratio $\xi$. Each point on the curves is averaged over 20 independent Monte-Carlo trials. The PSNR evaluates the performance of solving \eref{eq:problem P4} using different combinations of dictionaries.}
\label{fig:tri_ellip_psnr}
\end{figure}

\begin{figure}[h]
\centering
\begin{tabular}{cc}
\includegraphics[width = 0.35\linewidth]{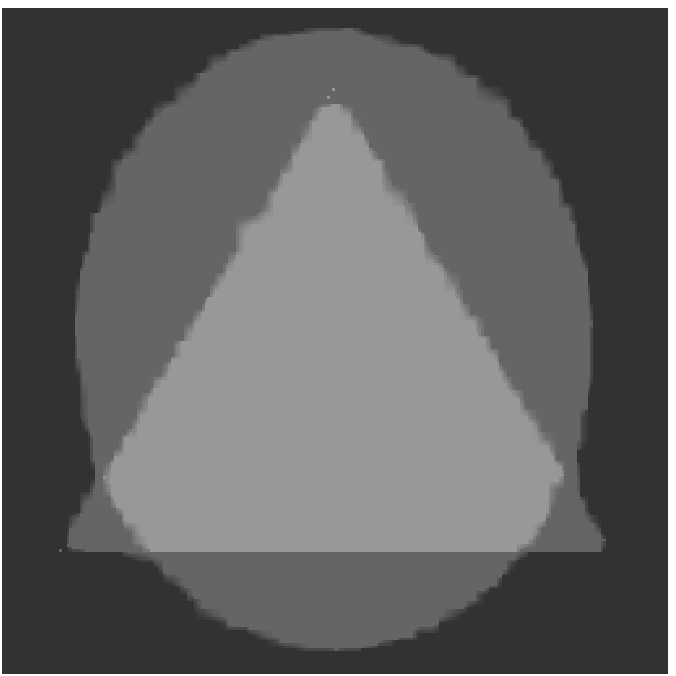}&
\includegraphics[width = 0.35\linewidth]{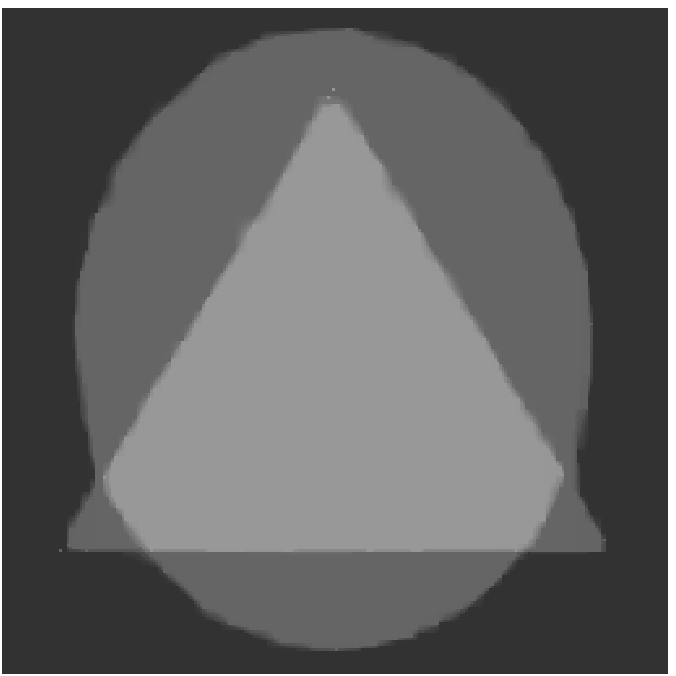}\\
(a) Wavelet, 34.77 dB & (b) Combined, 35.86 dB
\end{tabular}
\caption{Snapshot of the comparison between wavelet dictionary and a combined wavelet-contourlet dictionary at $\xi = 0.1$.}
\label{fig:tri_ellip_figs}
\end{figure}

Using the proposed ADMM algorithm (See Section IV), we plot the performance of the reconstruction result as a function of the sampling ratio. For each point of the sampling ratio, we perform a Monte-Carlo simulation over 20 independent trials to reduce the fluctuation caused by the randomness in the sampling pattern. The result in \fref{fig:tri_ellip_psnr} indicates that the combined dictionary is consistently better than the wavelet dictionary alone. A snapshot of the result at $\xi = 0.1$ is shown in \fref{fig:tri_ellip_figs}. As observed, the reconstruction along the edges of the ellipse is better in the combined dictionary than using wavelet alone.

\section{Reconstruction Algorithm}
In this section we present an alternating direction method of multipliers (ADMM) algorithm to solve \eref{eq:problem P4}. The ADMM algorithm has a tight connection with the proximal operator presented by Moreau in the 60's \cite{Moreau_1965}, and later by Eckstein and Bertsekas \cite{Eckstein_Bertsekas_1992} in the 90's. The application of ADMM to image deconvolution was first mentioned in \cite{Yang_Zhang_Yin_2009}. For brevity we skip the introduction of ADMM algorithm because comprehensive tutorials are easily accessible \cite{Han_Yuan_2012, Boyd_Parikh_Chu_Peleato_Eckstein_2011}. Instead, we highlight the unique contributions of this paper, which include a particular operator splitting strategy and a multiscale implementation.

For notational simplicity we consider a single dictionary so that $L=1$. Generalization to $L>1$ is straight forward. Also, in our derivation we focus on the anisotropic total variation so that $\|\vx\|_{TV} = \|\mD_x \vx\|_1 + \|\mD_y \vx\|_1$. Extension to isotropic total variation follows the same idea as presented in \cite{Chan_Khoshabeh_Gibson_2011}.

\vspace{-0.2cm}
\subsection{ADMM and Operator Splitting}
A central question about ADMM algorithms is which of the variables should be splitted so that the subsequent subproblems can be efficiently solved. Inspecting \eref{eq:problem P4}, we observe that there are many possible choices. For example, we could split the quadratic term in \eref{eq:problem P4} by defining an auxiliary variable $\vu = \mS\vx$, or we could keep the quadratic term without a split. In what follows, we present an overview of our proposed splitting method and discuss the steps in subsequent subsections.

We start the ADMM algorithm by introducing three auxiliary variables $\vr = \vx$, $\vu_{\ell} = \mPhi_{\ell}\vx$, and $\vv = \mD\vx$. Consequently, we rewrite the optimization problem as
\begin{equation}
\begin{array}{cl}
\minimize{\vx,\vr,\vu_{\ell},\vv} &\; \frac{1}{2}\|\vb-\mS\vr \|^2 + \lambda_{\ell} \|\mW_{\ell}\vu_{\ell}\|_1 + \beta \|\vv\|_{1}\\
\subjectto                        &\; \vr = \vx, \quad \vu_{\ell} = \mPhi_{\ell}^T\vx, \quad \vv=\mD\vx.
\end{array}
\label{eq:Intermediate_Lagranian}
\end{equation}

The ADMM algorithm is a computational procedure to find a stationary point of \eref{eq:Intermediate_Lagranian}. The idea is to consider the augmented Lagrangian function defined as
\begin{align}
&\calL\left(\vx,\vu_{\ell},\vr,\vv,\vw,\vy_{\ell},\vz\right) \notag\\
&\quad=\frac{1}{2}\|\vb - \mS\vr \|^2 + \lambda_{\ell}\|\mW_{\ell}\vu_{\ell}\|_1 + \beta\|\vv\|_1 \label{eq:ADMM_Lagrangian} \\
&\quad\quad-\vw^{T}\left(\vr - \vx \right) - \vy_{\ell}^{T}\left(\vu_{\ell} - \mPhi_{\ell}^T\vx \right) - \vz^{T}\left(\vv - \mD\vx \right) \notag\\
&\quad\quad+ \frac{\mu}{2}\|\vr - \vx \|^2 + \frac{\rho_{\ell}}{2}\|\vu_{\ell} - \mPhi_{\ell}^T\vx \|^2 + \frac{\gamma}{2}\|\vv - \mD\vx \|^2. \notag
\end{align}
In \eref{eq:ADMM_Lagrangian}, the vectors $\vw$, $\vy_{\ell}$ and $\vz$ are the Lagrange multipliers; $\lambda_{\ell}$ and $\beta$ are the regularization parameters, and $\mu$, $\rho_{\ell}$ and $\gamma$ are the internal half quadratic penalty parameters. The stationary point of the augmented Lagrangian function can be determined by solving the following sequence of subproblems
\small
\begin{align*}
\vx^{(k+1)}         & = \argmin{\vx}   \calL \left(\vx,\vu_{\ell}^{(k)},\vr^{(k)},\vv^{(k)},\vw^{(k)},\vy_{\ell}^{(k)},\vz^{(k)}\right),\\
\vu_{\ell}^{(k+1)}  & = \argmin{\vu_{\ell}} \calL \left(\vx^{(k+1)},\vu_{\ell},\vr^{(k)},\vv^{(k)},\vw^{(k)},\vy_{\ell}^{(k)},\vz^{(k)}\right),\\
\vr^{(k+1)}         & = \argmin{\vr} \calL \left(\vx^{(k+1)},\vu_{\ell}^{(k+1)},\vr,\vv^{(k)},\vw^{(k)},\vy_{\ell}^{(k)},\vz^{(k)}\right),\\
\vv^{(k+1)}         & = \argmin{\vv}   \calL \left(\vx^{(k+1)},\vu_{\ell}^{(k+1)},\vr^{(k+1)},\vv,\vw^{(k)},\vy_{\ell}^{(k)},\vz^{(k)}\right),
\end{align*}
\normalsize
and the Lagrange multipliers are updated as
\begin{subequations}
\begin{align}
\vy_{\ell}^{(k+1)}  & = \vy_{\ell}^{(k)} - \rho_{\ell}\left(\vu_{\ell}^{(k+1)} - \mPhi_{\ell}^T\vx^{(k+1)}\right),\label{eq:y1_solution}\\
\vw^{(k+1)}         & = \vw^{(k)} - \mu\left(\vr^{(k+1)} - \vx^{(k+1)}\right),\label{eq:y2_solution}\\
\vz^{(k+1)}         & = \vz^{(k)} - \gamma\left(\vv^{(k+1)} - \mD\vx^{(k+1)}\right).\label{eq:z_solution}
\end{align}
\end{subequations}
We now discuss how each subproblem is solved.

\vspace{-1ex}
\subsection{Subproblems}
\subsubsection{$\vx$-subproblem}
The $\vx$-subproblem is obtained by dropping terms that do not involve $\vx$ in \eref{eq:ADMM_Lagrangian}. This yields
\begin{align}
\vx^{(k+1)}
&= \argmin{\vx} - \vw^{T}\left(\vr -\vx \right) -\vy_{\ell}^T\left(\vu_{\ell} - \mPhi_{\ell}^T\vx \right)  \notag\\
&\quad  - \vz^{T}\left(\vv -\mD\vx \right) + \frac{\mu}{2}\|\vr - \vx \|^2  \label{eq:x-subproblem}\\
&\quad  + \frac{\rho_{\ell}}{2}\|\vu_{\ell} - \mPhi_{\ell}^T\vx \|^2 + \frac{\gamma}{2}\|\vv - \mD\vx\|^2. \notag
\end{align}
Problem \eref{eq:x-subproblem} can be solved by considering the first-order optimality condition, which yields a normal equation
\begin{align}
&\left(\rho_{\ell}\mPhi_{\ell}\mPhi_{\ell}^T + \mu \mI + \gamma \mD^{T}\mD \right)\vx^{(k+1)} \label{eq:x-subproblem normal}\\
&\quad\quad\quad = \mPhi_{\ell}\left(\rho_{\ell} \vu_{\ell} - \vy_{\ell}\right) + \left(\mu\vr - \vw\right) +  \mD^{T}\left(\gamma\vv-\vz\right). \notag
\end{align}
The matrix in \eref{eq:x-subproblem normal} can be simplified as $(\rho_{\ell}+\mu)\mI+\gamma\mD^T\mD$, because for any frame $\mPhi_{\ell}$, it holds that $\mPhi_{\ell}\mPhi_{\ell}^T = \mI$. Now, since the matrix $\mD^T\mD$ is a circulant matrix, the matrix $(\rho_{\ell}+\mu)\mI+\gamma\mD^T\mD$ is diagonalizable by the Fourier transform. This leads to a closed form solution as
\begin{equation}
\vx^{(k+1)} = \calF^{-1}\left[ \frac{\calF(\mbox{RHS})}{(\rho_{\ell} + \mu)\mI + \gamma |\calF(\mD)|^2 }\right],\label{eq:x-subproblem solution}
\end{equation}
where RHS denotes the right hand side of \eref{eq:x-subproblem normal}, $\calF(\cdot)$ denotes the 2D Fourier transform, $\calF^{-1}(\cdot)$ denotes the 2D inverse Fourier transform, and $|\calF(\mD)|^2$ denotes the magnitude square of the eigenvalues of the differential operator $\mD$.

\begin{remark}
If we do not split the quadratic function $\|\vb - \mS\vx \|^2$ using $\vr = \vx$, then the identity matrix $\mu\mI$ in \eref{eq:x-subproblem normal} would become $\mu\mS^T\mS$. Since $\mS$ is a diagonal matrix containing 1's and 0's, the matrix $\rho_{\ell}\mPhi_{\ell}\mPhi_{\ell}^T + \mu \mS^T\mS + \gamma \mD^{T}\mD$ is not diagonalizable using the Fourier transform.
\end{remark}

\subsubsection{$\vu_{\ell}$-subproblem}
The $\vu_{\ell}$-subproblem is given by
\begin{equation}
\min_{\vu_{\ell}} \;\; \lambda_{\ell}\|\mW_{\ell}\vu_{\ell} \|_1 - \vy_{\ell}^{T}\left(\vu_{\ell} - \mPhi_{\ell}^T\vx \right) + \frac{\rho_{\ell}}{2}\|\vu_{\ell}-\mPhi_{\ell}^T\vx \|^2.
\label{eq:u1_subproblem}
\end{equation}
Since $\mW_{\ell}$ is a diagonal matrix, \eref{eq:u1_subproblem} is a separable optimization consisting of a sum of scalar problems. By using the standard shrinkage formula \cite{Chan_Khoshabeh_Gibson_2011}, one can show that the closed-form solution of \eref{eq:u1_subproblem} exists and is given by
\begin{equation}
\vu_{\ell}^{(k+1)} = \max \left( \left|\valpha_{\ell} + \frac{\vy_{\ell}}{\rho_{\ell}}\right| - \frac{\lambda_{\ell} \vwtilde_{\ell}}{\rho_{\ell}}, 0 \right)\cdot \mathrm{sign}\left( \valpha_{\ell} + \frac{\vy_{\ell}}{\rho_{\ell}} \right),
\label{eq:u1_subproblem_solution}
\end{equation}
where $\vwtilde_{\ell} = \mbox{diag}(\mW_{\ell})$ and $\valpha_{\ell}=\mPhi_{\ell}^T\vx$.

\begin{remark}
If we do not split using ${\vu_{\ell}} = \mPhi_{\ell}^T\vx$, then the ${\vu_{\ell}}$-subproblem is not separable and hence the shrinkage formula cannot be applied. Moreover, if we split ${\vu_{\ell}} = \mW_{\ell}\mPhi_{\ell}^T\vx$, \emph{i.e.}, include $\mW_{\ell}$, then the $\vx$-subproblem will contain $\mPhi_{\ell}\mW_{\ell}\mPhi_{\ell}^T$, which is not diagonalizable using the Fourier transform.
\end{remark}

\subsubsection{$\vr$-subproblem}
The $\vr$-subproblem is the standard quadratic minimization problem:
\begin{equation}
\min_{\vr} \; \frac{1}{2}\|\mS\vr-\vb\|^2 - \vw^T\left( \vr -\vx\right) + \frac{\mu}{2}\|\vr - \vx \|^2.
\label{eq:u2_subproblem}
\end{equation}
Taking the first-order optimality yields a normal equation
\begin{equation}
\left(\mS^T\mS+\mu\mI \right) \vr = \left({\mS^T\vb+\vw + \mu\vx }\right).
\label{eq:u2_subproblem_solution}
\end{equation}
Since $\mS$ is a diagonal binary matrix, \eref{eq:u2_subproblem_solution} can be evaluated via an element-wise computation.

\begin{remark}
\eref{eq:u2_subproblem_solution} shows that our splitting strategy of using $\vr = \vx$ is particularly efficient because $\mS$ is a diagonal matrix. If $\mS$ is a general matrix, \emph{e.g.}, i.i.d. Gaussian matrix in \cite{Dai_Milenkovis_2009}, then solving \eref{eq:u2_subproblem_solution} will be less efficient.
\end{remark}

\subsubsection{$\vv$-subproblem}
The $\vv$-subproblem is the standard total variation problem:
\begin{equation}
\min_{\vv} \; \beta\|\vv\|_1-\vz^{T}\left(\vv - \mD\vx \right) + \frac{\gamma}{2}\|\vv-\mD\vx \|^2.
\label{eq:v_subproblem}
\end{equation}
The solution is given by
\begin{equation}
\vv^{(k+1)} = \max\left( \left |\mD\vx+\frac{\vz}{\gamma} \right | - \frac{\beta}{\gamma} , 0 \right) \cdot \mathrm{sign} \left( \mD\vx+\frac{\vz}{\gamma} \right).
\label{eq:v_subproblem_solution}
\end{equation}

The overall ADMM algorithm is shown in Algorithm \ref{alg:ADMM_algorithm}.

\begin{algorithm}
  \caption{ADMM Algorithm}
  \label{alg:ADMM_algorithm}
  \begin{algorithmic}[1]
  \REQUIRE {$\vb$,$\mS$}
   \STATE $\vx^{(0)} = \mS\vb$, ${\vu_{\ell}^{(0)}} = \mPhi_{\ell}^{T}\vx^{(0)}$, $\vr^{(0)} = \vx^{(0)}$, $\vv^{(0)} = \mD\vx^{(0)}$
   \WHILE{$\|\vx^{(k+1)}-\vx^{(k)} \|_2 / \|\vx^{(k)} \|_2 \ge  \mathrm{tol}$}
            \STATE Solve $\vx$-subproblem by \eref{eq:x-subproblem solution}.
            \STATE Solve ${\vu_{\ell}}$, $\vr$ and $\vv$ subproblems by \eref{eq:u1_subproblem_solution}, \eref{eq:u2_subproblem_solution} and \eref{eq:v_subproblem_solution}.
            \STATE Update multipliers by \eref{eq:y1_solution}, \eref{eq:y2_solution} and \eref{eq:z_solution}.
   \ENDWHILE
   \STATE \textbf{return} $\vx^{\ast} \gets \vx^{(k+1)} $
  \end{algorithmic}
\end{algorithm}

\subsection{Parameters}

\begin{table}[h]
\centering
\normalsize
\begin{tabular}{cll}
\hline
Parameter   & Functionality         & Values \\
\hline
$\lambda_1$ & Wavelet sparsity      &   $4\times10^{-5}$ \\
$\lambda_2$ & Contourlet sparsity   &   $2\times10^{-4}$ \\
$\beta$     & Total variation       &   $2\times10^{-3}$\\
$\rho_1$    & Half quad. penalty for Wavelet        &   $0.001$\\
$\rho_2$    & Half quad. penalty for Contourlet     &   $0.001$\\
$\mu$       & Half quad. penalty for $\vr = \vx$    &   $0.01$\\
$\gamma$    & Half quad. penalty for $\vv = \mD\vx$ &   $0.1$\\
\hline
\end{tabular}
\caption{Summary of Parameters.}
\label{table:parameters}
\vspace{-2ex}
\end{table}

The regularization parameters ($\lambda_{\ell}$, $\beta$) and internal half quadratic penalty parameters ($\rho_{\ell}$, $\mu$, $\gamma$) are chosen empirically. Table \ref{table:parameters} provides a summary of the parameters we use in this paper. These values are the typical values we found over a wide range of images and testing conditions. For detailed experiments of the parameter selection process, we refer the readers to our supplementary technical report in \cite{Liu_Chan_Nguyen_2015}.

\subsection{Convergence Comparison}
Since \eref{eq:problem P4} is convex, standard convergence proof of ADMM applies (c.f. \cite{Boyd_Parikh_Chu_Peleato_Eckstein_2011}). Thus, instead of repeating the convergence theory, we compare our proposed algorithm with a subgradient algorithm proposed by Hawe et al. \cite{Hawe_Kleinsteuber_Diepold_2011}.

To set up the experiment, we consider the uniformly random sampling pattern $\mS$ with sampling ratios $\xi = 0.1, 0.15, 0.2$. For both our algorithm and the subgradient algorithm proposed in \cite{Hawe_Kleinsteuber_Diepold_2011}, we consider a single wavelet dictionary using Daubechies wavelet ``db2'' with 2 decomposition levels. Other choices of wavelets are possible, but we observe that the difference is not significant.

\begin{figure}[ht!]
\centering
\includegraphics[width=\linewidth]{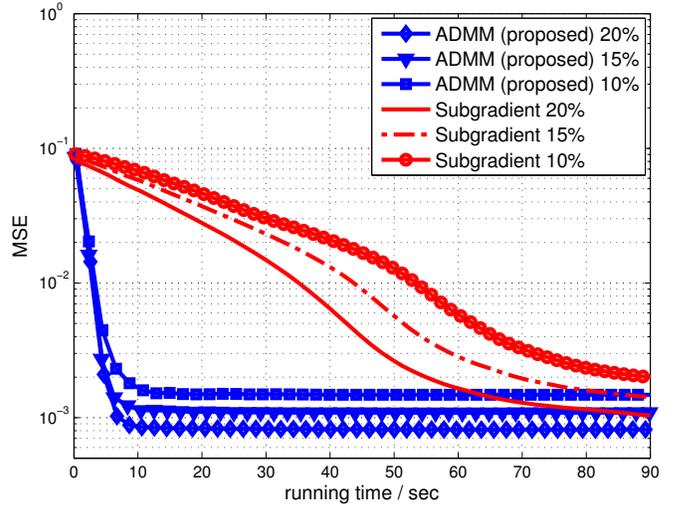}
\caption{Comparison of the rate of convergence between ADMM (proposed) and subgradient algorithms \cite{Hawe_Kleinsteuber_Diepold_2011} for single wavelet dictionary. We used ``Aloe'' as a test image. The ADMM algorithm requires approximately 10 seconds to reach steady state. The subgradient algorithm requires more than $9\times$ running time than the ADMM algorithm to reach steady state.}
\label{Fig:Converge_Curves}
\end{figure}

\fref{Fig:Converge_Curves} shows the convergence results of our proposed algorithm and the subgradient algorithm. It is evident from the figure that the ADMM algorithm converges at a significantly faster rate than the subgradient algorithm. In particular, we see that the ADMM algorithm reaches a steady state in around 10 seconds, whereas the subgradient algorithm requires more than 90 seconds.

\subsection{Multiscale ADMM}
The ADMM algorithm shown in Algorithm~\ref{alg:ADMM_algorithm} can be modified to incorporate a multiscale warm start. The idea works as follows.

First, given the observed data $\vb$, we construct a multiscale pyramid $\{ \vb_{q} \;|\; q = 0,\ldots,Q-1 \}$ of $Q$ levels, with a scale factor of 2 across adjacent levels. Mathematically, by assuming without loss of generality that $N$ is a power of 2, we define a downsampling matrix $\mA_q$ at the $q$th level as
\begin{align*}
\mA_q = [\ve_1, \vzero, \ve_2, \vzero, \ldots, \vzero, \ve_{N/{2^q}}],
\end{align*}
where $\ve_k$ is the $k$th standard basis. Then, we define $\vb_q$ as
\begin{align}
\vb_{q} = \mA_q \vb_{q-1},
\end{align}
for $q = 1,\ldots,Q-1$, and $\vb_0 = \vb$. Correspondingly, we define a pyramid of sampling matrices $\{\mS_q \;|\; q = 0,\ldots,Q-1\}$, where
\begin{align}
\mS_q = \mA_q \mS_{q-1},
\end{align}
with the initial sampling matrix $\mS_0 = \mS$.

The above downsampling operation allows us to solve \eref{eq:problem P4} at different resolution levels. That is, for each $q = 0,\ldots,Q-1$, we solve the problem
\begin{equation}
\vx_q = \argmin{\vx} \; \frac{1}{2}\| \mS_q \vx - \vb_q \|_2^2 + \lambda_{\ell} \|\mW_{\ell}\mPhi_{\ell}^T\vx\|_1 + \beta\|\vx\|_{TV},
\end{equation}
where $\mPhi_{\ell}$ and $\mW_{\ell}$ are understood to have appropriate dimensions.

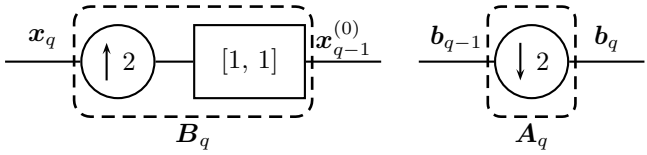
\begin{figure}[h]
\centering
\begin{pspicture}(0,-1.15)(8,0.65)
\psline{-}(0,0)(1,0)
\rput(1.5,0){\upsampling{$2$}}
\psline{-}(2,0)(2.5,0)
\rput(3.25,0){\shortfilter{$[1,\,1]$}}
\psline{-}(4,0)(5,0)
\rput(0.5,0.3){$\vx_q$}
\rput(4.5,0.3){$\vx_{q-1}^{(0)}$}
\psframe[linewidth=1pt,framearc=0.3,linestyle=dashed](0.9,-0.75)(4.1,0.75)
\rput(2.5,-1){$\mB_q$}
\psline{-}(5.5,0)(6.5,0)
\rput(7,0){\downsampling{$2$}}
\psline{-}(7.5,0)(8.5,0)
\rput(6,0.3){$\vb_{q-1}$}
\rput(8,0.3){$\vb_{q}$}
\psframe[linewidth=1pt,framearc=0.3,linestyle=dashed](6.4,-0.75)(7.6,0.75)
\rput(7,-1){$\mA_q$}
\end{pspicture}
\caption{Schematic diagram showing the operations of $\mA_q$ and $\mB_q$: $\mA_q$ downsamples the observed data $\vb_q$ by a factor of 2; $\mB_q$ upsamples the solution $\vx_q$ by a factor of 2, followed by a two-tap filter of impulse response $[1,\,1]$.}
\label{fig:schematic up down sampling}
\end{figure}

Once $\vx_q$ is computed, we feed an upsampled version of $\vx_q$ as the initial point to the $(q-1)$th level's optimization. More specifically, we define an upsampling and averaging operation:
\begin{equation}
\mB_q =  \left[\ve_1^T;\, \ve_1^T;\, \ve_2^T;\, \ve_2^T;\, \ldots;\, \ve_{N/2^q}^T;\, \ve_{N/2^q}^T\right],
\end{equation}
and we feed $\vx_q$, the solution at the $q$th level, as the initial guess to the problem at the $(q-1)$th level:
\begin{equation}
\vx_{q-1}^{(0)} = \mB_q\vx_q.
\end{equation}
A pictorial illustration of the operations of $\mA_q$ and $\mB_q$ is shown in \fref{fig:schematic up down sampling}. The algorithm is shown in Algorithm~\ref{alg:Multi_Scale_ADMM}.

\begin{algorithm}
  \caption{Multiscale ADMM Algorithm}
  \label{alg:Multi_Scale_ADMM}
  {\fontsize{9.5}{9.5}\selectfont
  \begin{algorithmic}[1]
  \REQUIRE {$\mS_{0},\ldots,\mS_{Q-1}$ and $\vb_{0},\ldots,\vb_{Q-1}$}
  \FOR{$q = Q-1$ \TO 0 }
    \STATE $\vx_{q}$ = ADMM($\vb_q,\mS_q$) with initial guess $\vx_{q}^{(0)}$
    \STATE Let $\vx_{q-1}^{(0)} = \mB_q\vx_q$, if $q\ge1$.
  \ENDFOR
  \STATE Output $\vx = \vx_{0}$.
  \end{algorithmic}}
\end{algorithm}

\begin{figure}[ht]
\centering
\includegraphics[width=1\linewidth]{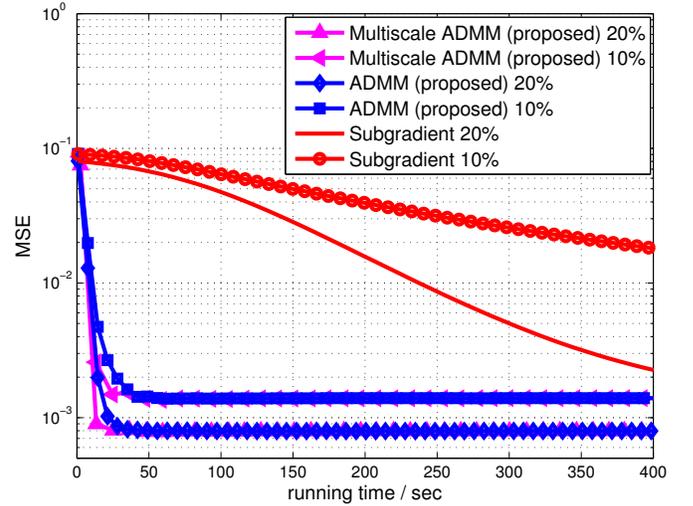}
\caption{Runtime comparison of original ADMM algorithm, multiscale ADMM algorithm and subgradient algorithm. All algorithms use the combined wavelet-contourlet dictionary. The testing image is ``Aloe'' and two sampling ratios (10\% and 20\%) are tested. $Q=3$ multiscale levels are implemented in this experiment.}
\label{fig:ADMM_multiADMM_comp}	
\end{figure}

To validate the effectiveness of the proposed multiscale warm start, we compare the convergence rate against the original ADMM algorithm for a combined dictionary case. In \fref{fig:ADMM_multiADMM_comp}, we observe that the multiscale ADMM converges at a significantly faster rate than the original ADMM algorithm. More specifically, at a sampling ratio of $20\%$, the multiscale ADMM algorithm converges in 20 seconds whereas the original ADMM algorithm converges in 50 seconds which corresponds to a factor of 2.5 in runtime reduction. For fairness, both algorithms are tested under the same platform of MATLAB 2012b / 64-bit Windows 7 / Intel Core i7 / CPU 3.2GHz (single thread) / 12 GB RAM.

\begin{remark}
When propagating the $q$th solution, $\vx_q$, to the $(q-1)$th level, we should also propagate the corresponding auxiliary variables {$\vu_{\ell}$}, $\vr$, $\vv$ and the Lagrange multipliers {$\vy_{\ell}$}, $\vw$ and $\vz$. The auxiliary variables can be updated according to $\vx_{q-1}^{(0)}$ as $\vu_{\ell,q-1}^{(0)} = \mPhi_{\ell}\vx_{q-1}^{(0)}$, $\vr_{q-1}^{(0)} = \vx_{q-1}^{(0)}$, and $\vv_{q-1}^{(0)} = \mD\vx_{q-1}^{(0)}$. For the Lagrange multipliers, we let $\vy_{\ell,q-1}^{(0)} = \mB_q\vy_{\ell,q}$, $\vw_{q-1}^{(0)} = \mB_q\vw_{q}$, and $\vz_{q-1}^{(0)} = \mB_q\vz_{q}$.
\end{remark}

\vspace{0.1cm}

\begin{remark}
The choice of the up/down sampling factor is not important. In our experiment, we choose a factor of 2 for simplicity in implementation. Other sampling factors such as $\sqrt{2}$ are equally applicable. Furthermore, the two-tap average filter $[1,1]$ in \fref{fig:schematic up down sampling} can be replaced by any valid averaging filter. However, experimentally we find that other choices of filters do not make a significant difference comparing to $[1,1]$.
\end{remark}

\section{Sampling Scheme}
In the above sections, we assume that the sampling matrix $\mS$ is given and is fixed. However, we have not yet discussed the design of the sampling probability $\{p_j\}_{j=1}^N$. The purpose of this section is to present an efficient design procedure.

\subsection{Motivating Example}
\begin{figure}[ht!]
\centering
\def\iw{0.27\linewidth}
\begin{tabular}{ccc}
\includegraphics[width=\iw]{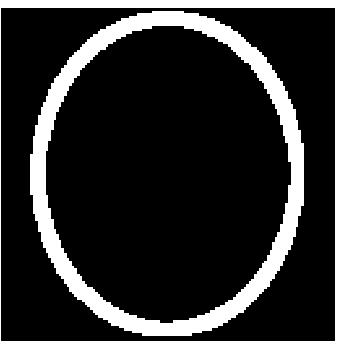}&
\includegraphics[width=\iw]{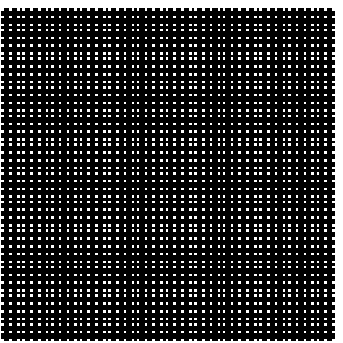}&
\includegraphics[width=\iw]{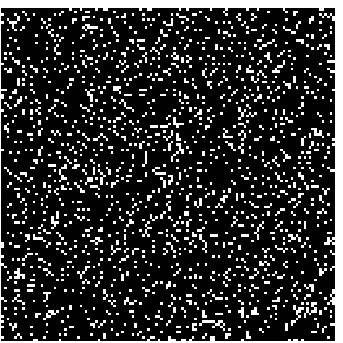}\\
$\xi = 0.1314$ & $\xi = 0.1348$ & $\xi = 0.1332$\\
\includegraphics[width=\iw]{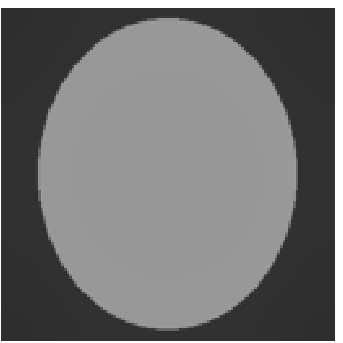}&
\includegraphics[width=\iw]{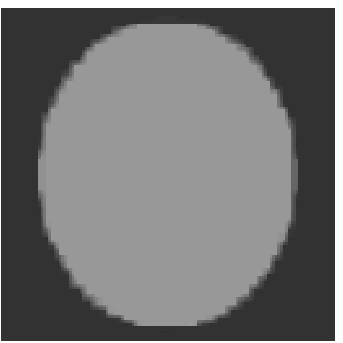}&
\includegraphics[width=\iw]{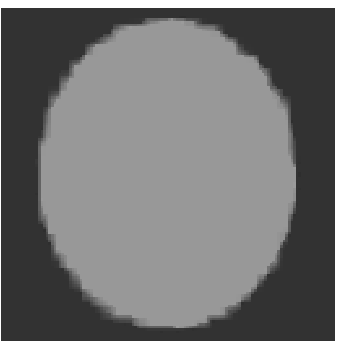}\\
(a) 45.527dB & (b) 29.488dB & (c) 30.857dB\\
\end{tabular}
\caption{Three sampling patterns and the corresponding reconstruction results using the proposed ADMM algorithm. Here, $\xi$ denotes the actual sampling ratio. (a) Sampling along the gradient; (b) Sampling from a grid; (c) Sampling from an uniformly random pattern.}
\label{fig: Sampling Example}
\end{figure}

Before our discussion, perhaps we should first ask about what kind of sampling matrix $\mS$ would work (or would not work). To answer this question, we consider an example shown in \fref{fig: Sampling Example}. In \fref{fig: Sampling Example} we try to recover a simple disparity map consisting of an ellipse of constant intensity and a plain background. We consider three sampling patterns of approximately equal sampling ratios $\xi$: (a) a sampling pattern defined according to the magnitude of the disparity gradient; (b) an uniform grid with specified sampling ratio $\sqrt{\xi}$ along both directions; (c) a random sampling pattern drawn from an uniform distribution with probability $\xi$. The three sampling patterns correspondingly generate three sampled disparity maps. For each sampled disparity map, we run the proposed ADMM algorithm and record the reconstructed disparity map. In all experiments, we use a wavelet dictionary for demonstration. \\

\fref{fig: Sampling Example} suggests a strong message: For a fixed sampling budget $\xi$, one should pick samples along gradients. However, the pitfall is that this approach is not practical for two reasons. First, the gradient of the disparity map is not available prior to reconstructing the disparity. Therefore, all gradient information can only be inferred from the color image. Second, the gradients of a color image could be very different from the gradients of the corresponding disparity map. Thus, inferring the disparity gradient from the color image gradient is a challenging task. In the followings, we present a randomized sampling scheme to address these two issues.

\subsection{Oracle Random Sampling Scheme}

We first consider an oracle situation where the gradients are assumed \emph{known}. The goal is to see how much improvement one should expect to see.

Let $\va = [a_1,\ldots,a_N]^T$ be a vector denoting the magnitude of the ground truth disparity map's gradient. Given this oracle information about the disparity gradients, we consider a soft decision rule where a pixel is sampled with probability defined according to some function of $\{a_j\}_{j=1}^N$. Such a function is chosen based on the intuition that the sampled subset of gradients should carry the maximum amount of information compared to the full set of gradients. One way to capture this intuition is to require that the average gradient computed from \emph{all} $N$ samples is similar to the average gradient computed from a \emph{subset} of $\xi N$ samples. 

To be more precise, we define the average gradient computed from all $N$ samples as
\begin{equation}
\mu \bydef \frac{1}{N} \sum_{j=1}^N a_j.
\label{eq:mu def}
\end{equation}
Similarly, we define the average gradient computed from a random subset of $\xi N$ samples as
\begin{equation}
Y \bydef \frac{1}{N} \sum_{j=1}^N \frac{a_j}{p_j} I_j,
\label{eq:Y def}
\end{equation}
where $\{I_j\}_{j=1}^N$ is a sequence of Bernoulli random variables with probability $\Pr[I_j = 1] = p_j$. Here, the division of $a_j$ by $p_j$ is to ensure that $Y$ is unbiased, \emph{i.e.}, $\E[Y] = \mu$. 

From \eref{eq:mu def} and \eref{eq:Y def}, minimizing the difference between $Y$ and $\mu$ can be achieved by minimizing the variance $\E[(Y-\mu)^2]$. Moreover, we observe that
\begin{align*}
\E\left[(Y-\mu)^2\right]= \frac{1}{N}\sum_{j=1}^{N}\frac{a_j^2}{p_j^2}\Var\left[I_j\right] = \frac{1}{N}\sum_{j=1}^{N}a_j^2 \left(\frac{1-p_j}{p_j}\right),
\end{align*}
where the last equality holds because $\Var[I_j] = p_j(1-p_j)$. Therefore, the optimal sampling probability $\{p_j\}_{j=1}^N$ can be found by solving the optimization problem
\begin{align*}
(P): \quad\quad \minimize{p_1,\ldots,p_N} &\quad\quad \frac{1}{N}\sum\limits_{j=1}^N \frac{a_j^2}{p_j}\\
\subjectto     &\quad\quad \frac{1}{N}\sum_{j=1}^N p_j = \xi, \;\mbox{and}\; 0 \le p_j \le 1,
\end{align*}
of which the solution is given by \cite[Lemma 2]{Chan_Zickler_Lu_2013}
\begin{equation}
p_j = \min( \tau a_j, 1),
\label{eq:optimal p}
\end{equation}
where $\tau$ is the root of the equation
\begin{equation}
g(\tau) \bydef \sum_{j=1}^N \min( \tau a_j, 1) - \xi N.
\label{eq:g tau}
\end{equation}

\begin{figure}[t]
\centering
\def\iw{0.48\linewidth}
\begin{tabular}{cc}
\includegraphics[width=\iw]{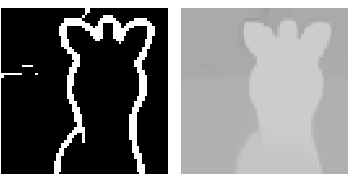}&
\includegraphics[width=\iw]{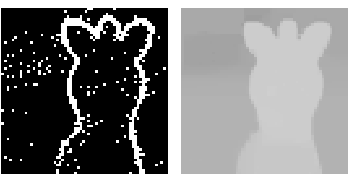}\\
(a) Greedy sampling  & (b) Random sampling \\
35.5201 dB, $\xi = 0.1157$ & 39.8976 dB, $\xi = 0.1167$
\end{tabular}
\caption{Comparison between a deterministic sampling pattern by selecting samples greedily according to the magnitude of $\{a_j\}$, and a randomized sampling pattern using the proposed scheme.}
\label{fig:compare sampling_deterministic vs random}
\end{figure}

It is interesting to compare this new random sampling scheme versus a greedy sampling scheme by picking the $\xi N$ pixels with the largest gradients. \fref{fig:compare sampling_deterministic vs random} shows the result. For the greedy sampling scheme, we first compute the gradient of the disparity map $\nabla \vx \bydef \sqrt{ (\mD_x \vx)^2 + (\mD_y \vx)^2}$ and threshold it to obtain a set of samples $\Omega \bydef \{ j \;|\; [\nabla \vx]_j > \alpha \|\nabla \vx\|_{\infty}\}$, where $\alpha = 0.1$ is the threshold. The actual sampling ratio is then $|\Omega|/N$. For the randomized scheme, we let $\va = \nabla \vx$ and we compute $p_j$ according to \eref{eq:optimal p}. In this particular example, we observe that the randomized sampling scheme achieves a PSNR improvement of more than 4 dB.

\subsection{Practical Random Sampling Scheme}
We now present a practically implementable sampling scheme. The challenge that we have to overcome is that the gradient information of the disparity is not available. Therefore, we propose the following two-stage sampling process.

Our proposed sampling scheme consists of two stages - a pilot stage to obtain a rough estimate of the disparity, and a refinement stage to improve the disparity estimate. In the first step pilot stage, we pick $\xi N/2$ samples (\emph{i.e.}, half of the desired number of samples) using an uniformly random sampling pattern. This gives a sampling pattern $\{I_j^{(1)}\}_{j=1}^N$, where the superscript denotes the first stage. Correspondingly, we have a sampling matrix $\mS^{(1)}$ and the sampled data $\vb^{(1)}$. Given $\mS^{(1)}$ and $\vb^{(1)}$, we apply the ADMM algorithm to obtain a pilot estimate $\vx^{(1)}$.

In the second stage, we use the pilot estimate $\vx^{(1)}$ as a guide to compute the gradient $\nabla \vx^{(1)}$. By \eref{eq:optimal p}, this suggests that the optimal sampling probability is $p_j = \min( \tau [\nabla \vx^{(1)}]_j, 1 )$. However, in order to ensure that the $\xi N/2$ samples picked at the second stage \emph{do not overlap} with those picked in the first stage, instead of letting $p_j = \min( \tau [\nabla \vx^{(1)}]_j, 1 )$, we let $p_j = \min( \tau a_j, 1 )$, where
\begin{equation}
a_j =
\begin{cases}
[\nabla \vx^{(1)}]_j, &\quad \mbox{if} \quad I_j^{(1)} = 0,\\
0,                    &\quad \mbox{if} \quad I_j^{(1)} = 1.
\end{cases}
\label{eq:aj stage 2}
\end{equation}
In words, $a_j$ defined by \eref{eq:aj stage 2} forces $p_j = 0$ when the $j$th pixel is picked in the first step. Thus, the non-zero entries of $\{I_j^{(1)}\}$ and $\{I_j^{(2)}\}$ are mutually exclusive, and hence we can now apply the ADMM algorithm to recover $\vx^{(2)}$ from $\mS_1+\mS_2$ and $\vb_1+\vb_2$. The overall method is summarized in Algorithm~\ref{alg:Two_stage_algorithm}.

\begin{algorithm}
\scriptsize
{\fontsize{9.5pt}{9.5pt}\selectfont
  \caption{\fontsize{9pt}{9pt}{Two-Stage Algorithm} }
  \label{alg:Two_stage_algorithm}
  \begin{algorithmic}[1]
   \STATE Input: {$N$, $\xi$, $\vb$}
   \STATE Output: $\vx^{(2)}$ \vspace{0.5cm}
   \STATE \textbf{Stage 1:}
   \STATE \ \ Let $I_j^{(1)} = 1$ with probability $\xi/2$, for $j=1,\ldots,N$.
   \STATE \ \ Define $\mS^{(1)}$ and $\vb^{(1)}$ according to $\{I_j^{(1)}\}$.
   \STATE \ \ Compute $\vx^{(1)} = $ ADMM $(\mS^{(1)}, \vb^{(1)})$. \vspace{0.5cm}
   \STATE \textbf{Stage 2:}
   \STATE \ \ Compute $\nabla \vx^{(1)}$.
   \STATE \ \ For $j=1,\ldots,N$, define $a_j = \begin{cases}
                                [\nabla \vx^{(1)}]_j, &\; \mbox{if} \; I_j^{(1)} = 0,\\
                                0,                    &\; \mbox{if} \; I_j^{(1)} = 1.
                            \end{cases}$.
   \STATE \ \ Compute $\tau$ such that $\sum_{j=1}^N \min\{\tau a_j,\,1\} = N\xi/2$.
   \STATE \ \ Let $p_j = \min\{\tau a_j, \, 1\}$, for $j = 1,\ldots,N$.
   \STATE \ \ Let $I_j^{(2)} = 1$ with probability $p_j$, for $j=1,\ldots,N$.
   \STATE \ \ Define $\mS^{(2)}$ and $\vb^{(2)}$ according to $\{I_j^{(2)}\}$.
   \STATE \ \ Compute $\vx^{(2)} = $ ADMM $(\mS^{(1)}+\mS^{(2)}, \vb^{(1)}+\vb^{(2)})$.
  \end{algorithmic}
  }
\end{algorithm}

\subsection{Further Improvement by PCA}
The two-stage sampling procedure can be further improved by utilizing the prior information of the color image. The intuition is that since both color image and disparity map are captured from the same scene, strong gradients in the disparity map should align with those in the color image. However, since a color image typically contains complex gradients which are irrelevant to the disparity reconstruction, it is important to filter out these unwanted gradients while preserving the important ones. To this end, we consider the following patch-based principal component analysis.

Given a color image $\vy \in \R^N$, we construct a collection of patches $\{\vy_j\}_{j=1}^N$ where $\vy_j \in \R^d$ denotes a vectorization of the $j$th patch of size $\sqrt{d} \times \sqrt{d}$ centered at pixel $j$ of the image. For patches centered at the corners or boundaries of the image, we apply a symmetrical padding to make sure that their sizes are $\sqrt{d} \times \sqrt{d}$. This will give us a total of $N$ patches.

\begin{figure}[t]
\centering
\includegraphics[width=0.85\linewidth]{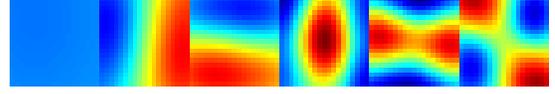}
\caption{The first 6 eigenvectors of the data matrix $\mY\mY^T$, where $\mY$ is obtained from the color image corresponding to \fref{fig:compare sampling_deterministic vs random}. In this example we set the patch size as $19 \times 19$ so that $d = 361$. The range of the color index of this figure is $[-0.1,\,0.1]$.}
\label{fig:PCA basis}
\end{figure}

Next, we form a data matrix $\mY \bydef [\vy_1,\vy_2,\ldots,\vy_N]$. This data matrix leads to a principal component decomposition as
\begin{equation}
\mY\mY^T = \mU \mLambda \mU^T,
\label{eq:pca decomposition}
\end{equation}
where $\mU$ is the eigenvector matrix, and $\mLambda$ is the eigenvalue matrix. Geometrically, the projection of any patch $\vy_j$ onto the subspace spanned by any eigenvector $\vu_i$ is equivalent to applying a finite impulse response filter to the patch, \emph{i.e.}, $\vu_i^T\vy_j$. In many cases, except for the first eigenvector $\vu_1$, all remaining eigenvectors $\vu_2,\ldots,\vu_d$ are in the form of differential operators (of different orders and orientations, see examples in \fref{fig:PCA basis}). More interestingly, these filters are typically \emph{bandpass} filters, which suggest that both low frequency components (\emph{e.g.}, smooth regions) and high frequency components (\emph{e.g.}, complex textures) of the color image can be filtered by applying the projections. Consequently, we consider the following filtered signal
\begin{equation}
a_j = \sum_{i=2}^{d'} |\inprod{\vu_i,\vy_j}|, \quad j = 1,\ldots, N,
\label{eq:aj pca}
\end{equation}
where $d' < d$ is a tunable parameter (which was set to $d' = 16$ for $d = 49$ in this paper). Here, the absolute value in \eref{eq:aj pca} is used to get the magnitude of $\inprod{\vu_i,\vy_j}$, as $a_j$ must be a non-negative number.

To see how this PCA concept can be incorporated into our two-stage sampling scheme, we make the following observations. First, the uniform sampling in Stage-1 can well be replaced by the PCA approach. In particular, instead of setting $I_j^{(1)} = 1$ with probability $\xi/2$, we can define $a_j$ according to \eref{eq:aj pca}, and let $p_j = \min(\tau a_j, 1)$ for $\tau$ being the root of \eref{eq:g tau}. Consequently, we let $I_j^{(1)} = 1$ with probability $p_j$.

In Stage-2, since we have already had a pilot estimate of the disparity map, it is now possible to replace $\mY$ in \eref{eq:pca decomposition} by a data matrix $\mX = [\vx_1^{(1)},\ldots,\vx_N^{(1)}]$, where each $\vx_j^{(1)}$ is a $d$-dimensional patch centered at the $j$th pixel of $\vx^{(1)}$.  Thus, instead of setting $a_j = [\nabla \vx^{(1)}]_j$ in \eref{eq:aj stage 2}, we can set $a_j = \sum_{i=2}^{d'} |\inprod{\vu_i,\vx_j^{(1)}}|$ using \eref{eq:aj pca}. The advantage of this new $a_j$ is that it softens the sampling probability at the object boundaries to a neighborhood surrounding the boundary. This reduces the risk of selecting irrelevant samples because of a bad pilot estimate.
\begin{figure*}[ht!]
\centering
\def\iw{0.22\linewidth}
\begin{tabular}{cccc}
\includegraphics[width=\iw]{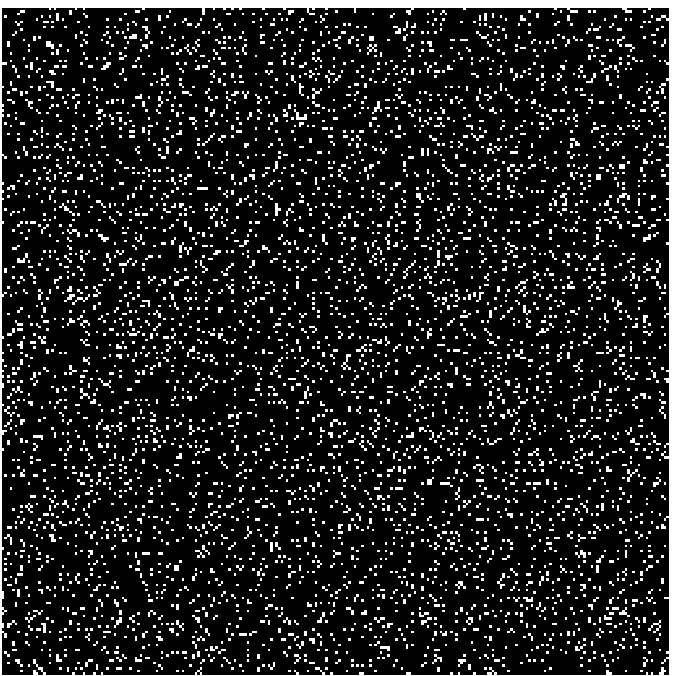}&
\includegraphics[width=\iw]{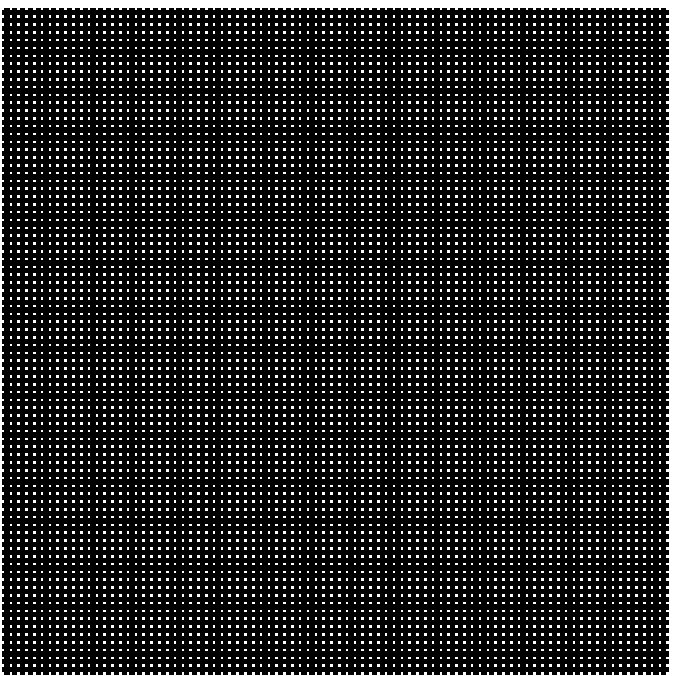}&
\includegraphics[width=\iw]{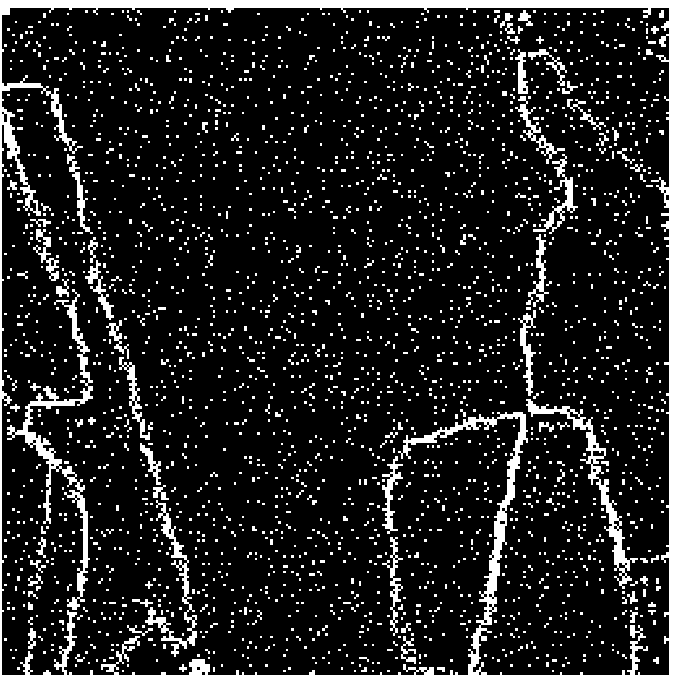}&
\includegraphics[width=\iw]{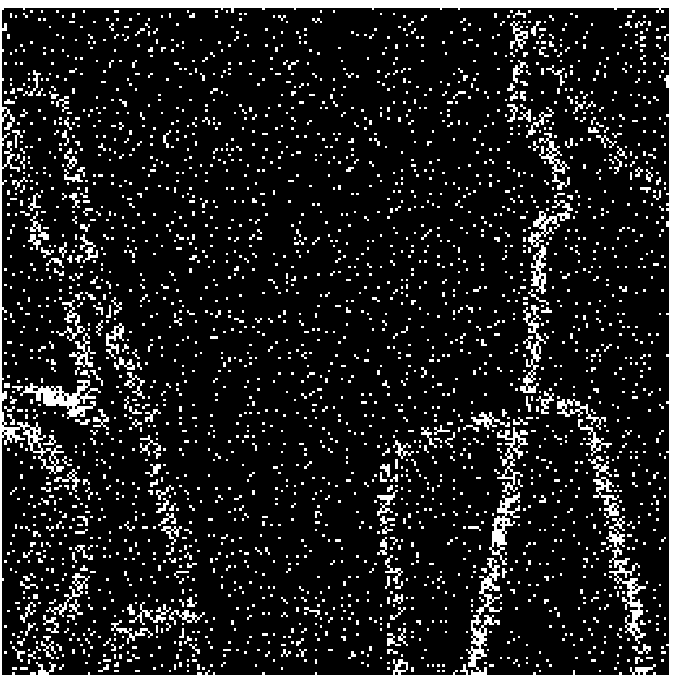}\\
\includegraphics[width=\iw]{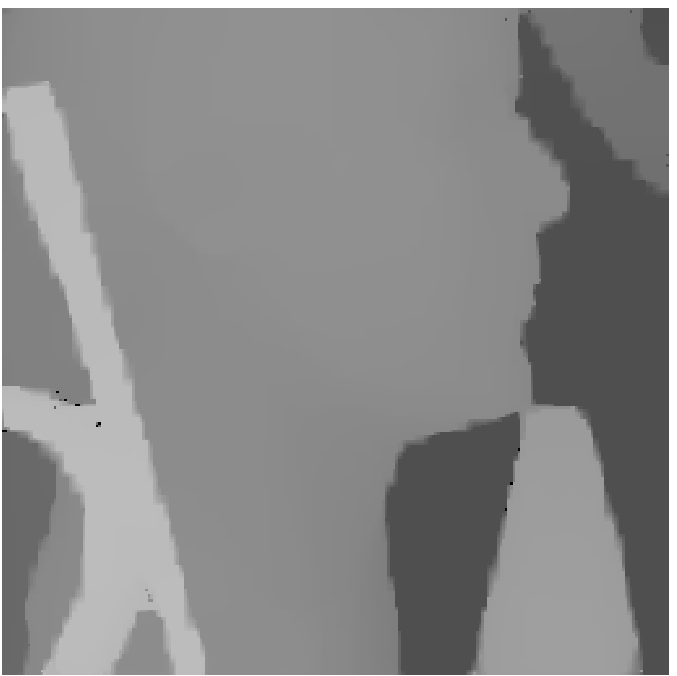}&
\includegraphics[width=\iw]{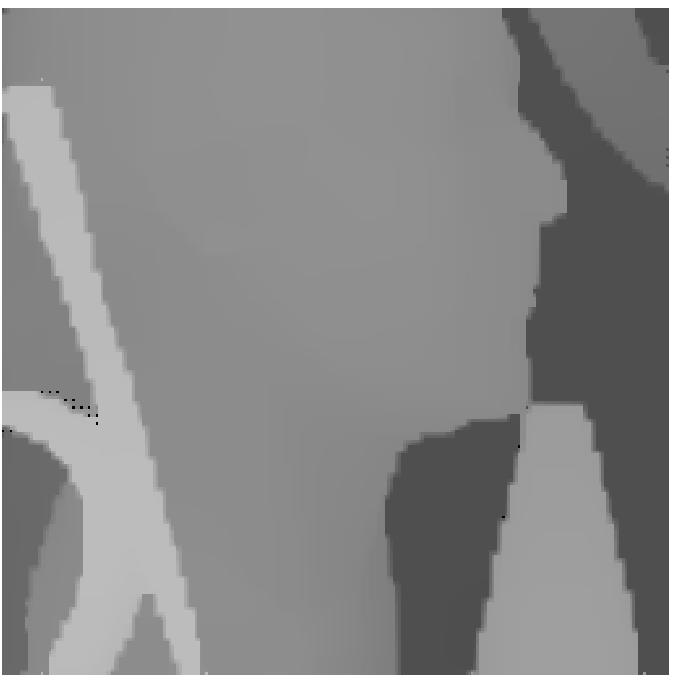}&
\includegraphics[width=\iw]{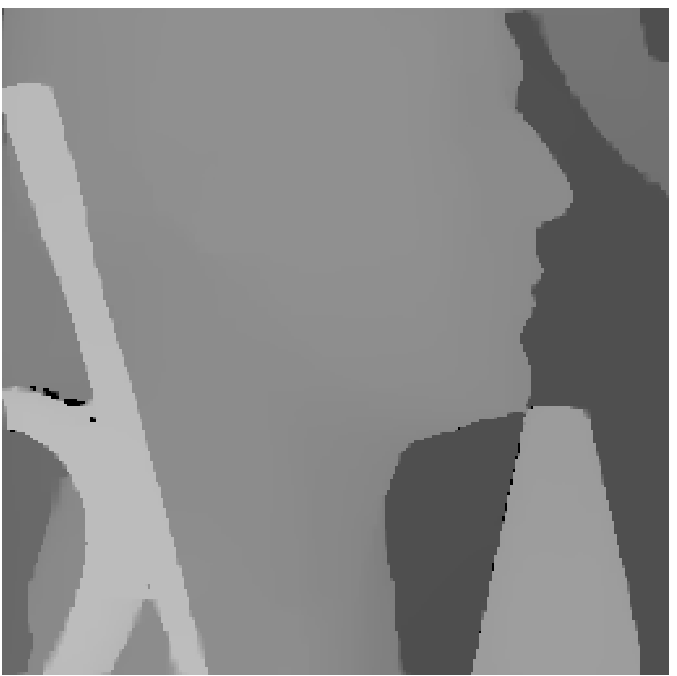}&
\includegraphics[width=\iw]{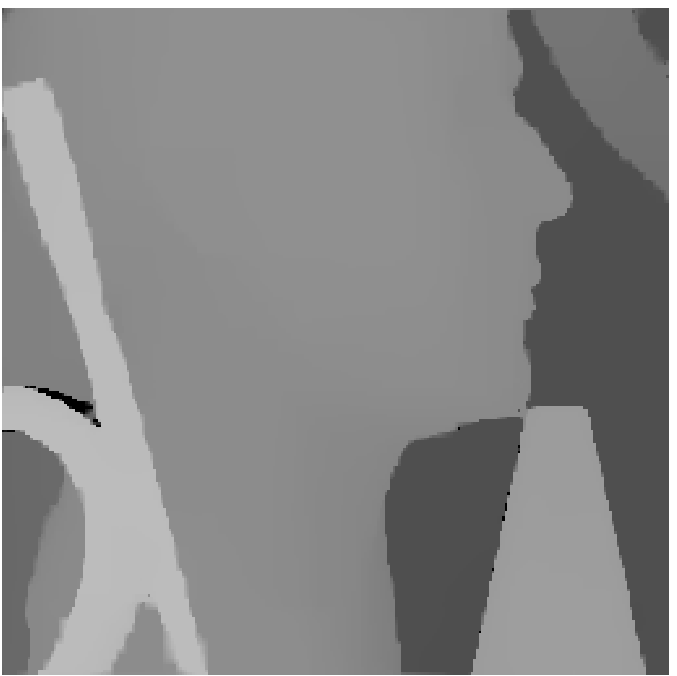}\\
(a) Uniform random & (b) Uniform grid & (c) Proposed w/o PCA & (d) Proposed w/ PCA\\
\end{tabular}
\begin{center}
\begin{tabular}{cccc}
\hline
 Method & Actual Sampling Ratio & Average PSNR / dB & Standard Deviation \\
 \hline
Uniform random & 0.1001 & 29.7495 & 0.3768\\
Uniform grid & 0.1128 & 30.2726 & 0.0000\\
Proposed w/o PCA & 0.1000 & 32.4532 & 0.8962\\
Proposed w/ PCA & 0.1002 &$\mathbf{33.7707}$ & 1.0435\\
\hline
\end{tabular}
\end{center}
\caption{Comparison between four sampling patterns. (a) Uniformly random sampling pattern; (b) Uniform grid; (c) Proposed two-stage sampling without PCA improvement; (d) Proposed two-stage sampling with PCA improvement. For the two-stage sampling in (c)-(d), we pick $\xi N/2$ uniformly random samples in stage 1, and pick the remaining $\xi N/2$ samples according to the pilot estimate from Stage 1. We conduct a Monte-Carlo simulation with 32 independent trials. The averages of PSNRs are presented in the Table.}
\label{fig:sampling compare pca}
\end{figure*}

\subsection{Comparisons}
As a comparison between sampling patterns, we consider a disparity map shown in \fref{fig:sampling compare pca}. Setting $\xi = 0.1$ (\emph{i.e.}, 10\%), we study four sampling patterns including two versions of our proposed two-stage method. We conduct a Monte-Carlo simulation by repeating 32 independent trials, and average the PSNRs. The results shown in \fref{fig:sampling compare pca}(c) are generated using the original two-stage sampling scheme without PCA improvement, whereas the results shown in \fref{fig:sampling compare pca}(d) are generated using an improved two-stage sampling scheme where the first stage is uniform and the second stage is PCA. These results indicate that for the same sampling ratio $\xi$, the choice of the sampling pattern has some strong influence to the reconstruction quality. For example, as compared to both uniform random sampling and grid sampling, the original two-stage sampling has about 2.44 dB improvement, and can be further improved by almost 3.76 dB using the PCA idea.


\section{Experimental Results}
In this section we present additional results to illustrate the performance of the proposed method.

\subsection{Synthetic Data}
We first compare the proposed algorithm with existing methods on the Middlebury dataset\footnote{http://vision.middlebury.edu/stereo/data/} where ground truth disparities are available. We consider two versions of the proposed algorithm: ``Proposed WT+CT Grid'' and ``Proposed WT+CT 2-Stage''. ``Proposed WT+CT Grid'' is the ADMM algorithm presented in Section IV using both wavelet and contourlet bases. Here, ``Grid'' refers to using a deterministic uniform grid sampling pattern and ``2-stage'' refers to using the 2-stage randomized sampling scheme presented in Section V. We use Daubechies wavelet ``db2'' with 2 decomposition levels for wavelet dictionary, and we set ``bior9-7'' wavelet function with [5 6] directional decompositions for contourlet dictionary.

We also compare our method with \cite{Hawe_Kleinsteuber_Diepold_2011}, which has three differences from ours: (1) \cite{Hawe_Kleinsteuber_Diepold_2011} uses a subgradient descent algorithm whereas we use an ADMM algorithm; (2) \cite{Hawe_Kleinsteuber_Diepold_2011} considers only a wavelet basis whereas we consider a combined wavelet-contourlet basis; (3) \cite{Hawe_Kleinsteuber_Diepold_2011} uses a combination of canny edges and uniformly random samples whereas we use a principled design process to determine samples.

In this experiment we do not compare with depth super resolution algorithms, \emph{e.g.}, \cite{Yang_Yang_Davis_Nister_2007, Li_Yu_Chai_2008, Park_Kim_Tai_Brown_Kweon_2011}. These methods require a color image to guide the actual reconstruction process, which is different from what is presented here because we only use the color image for designing the sampling pattern. As a reference of these methods, we show the results of a bicubic interpolation using uniform grid sampling pattern.

\tref{tab:PSNR_P_Bad_Noiseless} shows the PSNR values of various methods at different sampling ratios and sampling methods. It is clear that ``Proposed WT+CT 2-Stage'' outperforms the other methods by a significant margin. Moreover, as the sampling ratio increases, the PSNR gain of ``Proposed WT+CT 2-Stage'' is more prominent than that of other methods. For example, increasing from 5\% to 25\% for ``Art'', ``Proposed WT+CT 2-Stage'' demonstrates an 18 dB PSNR improvement whereas bicubic only demonstrates 3 dB improvement.

It is also instructive to compare the percentage of bad pixels (\% Bad Pixel), which is a popular metric to measure the quality of disparity estimates \cite{Scharstein_Szeliski_IJCV2002}. Given a threshold $\tau > 0$, the percentage of bad pixels is defined as
\begin{equation}
\mbox{\% Bad Pixel} \bydef \frac{1}{N}\sum_{j=1}^{N}\left(|\widehat{x}_j - x_j^*|>\tau \right),
\end{equation}
where $\widehat{\vx}$ is the reconstructed disparity and $\vx^*$ is the ground truth disparity. Percentage of bad pixels can be considered as an absolute difference metric as compared to the mean squared metric of PSNR.

\tref{tab:P_Bad_Noiseless} shows the percentage of bad pixels of various methods at different sampling ratios and sampling methods. The results indicate that ``Proposed WT+CT 2-Stage'' has a relatively higher \% Bad Pixel at $\tau = 1$ than other methods, but has a lower \% Bad Pixel at $\tau = 2$ and $\tau = 3$. This result suggests that most of the errors of ``Proposed WT+CT 2-Stage'' are \emph{small} and there are very few outliers. In contrast, bicubic grid (for example) has a low \% Bad Pixel at $\tau = 1$ but high \% Bad Pixel at $\tau = 2$ and $\tau = 3$. This implies that a significant portion of the bicubic results has large error. Intuitively, the results suggest that in the bicubic case, some strong edges and corners are completely missed, whereas these information are kept in ``Proposed WT+CT 2-Stage''.

Finally, we show the performance of the proposed algorithm towards additive i.i.d. Gaussian noise. The purpose of this experiment is to demonstrate the sensitivity and robustness of the algorithm in the presence of noise. While in reality the noise in disparity estimates is not i.i.d. Gaussian, the result presented here serves as a reference for the algorithm's performance. A more realistic experiment on real data will be illustrated in the next subsection.

\begin{figure}[h!]
\centering
    \includegraphics[width=0.85\linewidth]{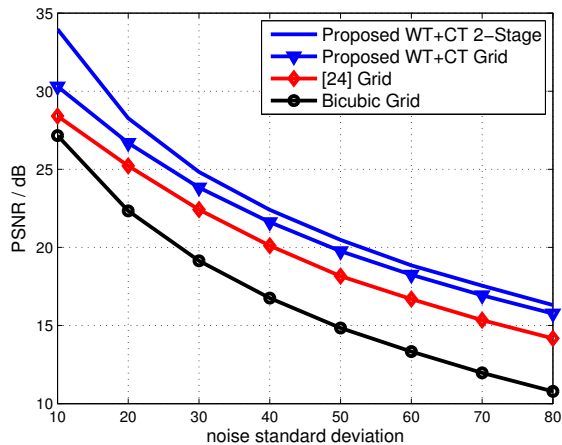}
    \caption{Comparison of reconstruction performance with noisy samples. We use ``Art'' disparity map as a test image, and set $\xi = 0.2$.}
    \label{fig:PSNR_noise_comp}
    \vspace{-0.5cm}
\end{figure}

The results are shown in Figure \ref{fig:PSNR_noise_comp}. Using ``Bicubic Grid'' as the baseline, we observe that ``Proposed WT+CT 2-Stage'' on average has 5.79 dB improvement, ``Proposed WT+CT Grid'' has 3.60 dB improvement, whereas ``\cite{Hawe_Kleinsteuber_Diepold_2011} Grid'' has only 3.02 dB improvement. This provides a good indicator of the robustness of the proposed methods.

 \begin{table*}[ht!]
\caption{Comparisons of reconstruction algorithms in terms of PSNR. We put N/A when the algorithm does not converge in 1000 iterations.}
\label{tab:PSNR_P_Bad_Noiseless}
\begin{center}
\scriptsize
\renewcommand{\arraystretch}{1.2}
\begin{tabular}{c|ccccccc}
    \hline
  Disparity  & Method  &   \multicolumn{5}{c}{Percentage of Samples / PSNR (dB)} &\\
 Name & Algorithm / Sampling Strategy & 5$\%$ & 10$\%$ & 15$\%$ & 20$\%$ & 25$\%$ &\\
    \hline\hline
   \hspace{-0.4cm}\multirow{4}{*}{Aloe}&\hspace{-0.4cm}Proposed WT+CT 2-Stage &\hspace{-0.1cm} 27.5998
                                       &\hspace{-0.1cm} \textbf{31.3877}  &\hspace{-0.1cm} \textbf{33.3693}
                                       &\hspace{-0.1cm} \textbf{36.4102}  &\hspace{-0.1cm} \textbf{38.6265} &\\
                                       &\hspace{-0.4cm}Proposed WT+CT Grid &\hspace{-0.1cm} 25.3236 &\hspace{-0.1cm} 28.9052
                                       &\hspace{-0.1cm} 30.0940  &\hspace{-0.1cm} 31.2956  &\hspace{-0.1cm} 32.3548 &\\
                                       &\hspace{-0.4cm}\cite{Hawe_Kleinsteuber_Diepold_2011} Grid &\hspace{-0.1cm} 25.1248
                                       &\hspace{-0.1cm} 27.8941  &\hspace{-0.1cm} 28.9504
                                       &\hspace{-0.1cm} 30.2371  &\hspace{-0.1cm} 31.6646 &\\
                                       &\hspace{-0.4cm}Bicubic Grid &\hspace{-0.1cm} \textbf{27.8899}  &\hspace{-0.1cm} 29.3532
                                       &\hspace{-0.1cm} 30.1019  &\hspace{-0.1cm} 31.0031  &\hspace{-0.1cm} 31.8908  &\\
    \hline
    \hspace{-0.4cm}\multirow{4}{*}{Art}&\hspace{-0.4cm}Proposed WT+CT 2-Stage &\hspace{-0.1cm} \textbf{30.8669}
                                       &\hspace{-0.1cm} \textbf{34.1495}  &\hspace{-0.1cm} \textbf{37.2801}
                                       &\hspace{-0.1cm} \textbf{42.9706}  &\hspace{-0.1cm} \textbf{48.0002} &\\
                                       &\hspace{-0.4cm}Proposed WT+CT Grid &\hspace{-0.1cm} 27.5176 &\hspace{-0.1cm} 28.9528
                                       &\hspace{-0.1cm} 30.8371  &\hspace{-0.1cm} 32.5150  &\hspace{-0.1cm} 33.7126 &\\
                                       &\hspace{-0.4cm}\cite{Hawe_Kleinsteuber_Diepold_2011} Grid &\hspace{-0.1cm} 27.0300 &\hspace{-0.1cm} N/A
                                       &\hspace{-0.1cm} N/A  &\hspace{-0.1cm} N/A  &\hspace{-0.1cm} N/A &\\
                                       &\hspace{-0.4cm}Bicubic Grid &\hspace{-0.1cm} 29.1550  &\hspace{-0.1cm} 30.3536
                                       &\hspace{-0.1cm} 31.1098  &\hspace{-0.1cm} 31.9473  &\hspace{-0.1cm} 32.8366 &\\
     \hline
    \hspace{-0.4cm}\multirow{4}{*}{Baby}&\hspace{-0.4cm}Proposed WT+CT 2-Stage &\hspace{-0.1cm} \textbf{39.6978}
                                        &\hspace{-0.1cm} \textbf{44.8958}  &\hspace{-0.1cm} \textbf{48.6631}
                                        &\hspace{-0.1cm} \textbf{52.5000}  &\hspace{-0.1cm} \textbf{52.0031} &\\
                                        &\hspace{-0.4cm}Proposed WT+CT Grid &\hspace{-0.1cm} 34.4421  &\hspace{-0.1cm} 36.7965
                                        &\hspace{-0.1cm} 37.6708  &\hspace{-0.1cm} 39.0504  &\hspace{-0.1cm} 40.0689 &\\
                                        &\hspace{-0.4cm}\cite{Hawe_Kleinsteuber_Diepold_2011} Grid &\hspace{-0.1cm} 33.6627
                                        &\hspace{-0.1cm} 35.3166  &\hspace{-0.1cm} 36.2522
                                        &\hspace{-0.1cm} 37.4513  &\hspace{-0.1cm} 38.7670 &\\
                                        &\hspace{-0.4cm}Bicubic Grid &\hspace{-0.1cm} 34.8368 &\hspace{-0.1cm} 36.2385
                                        &\hspace{-0.1cm} 37.1749  &\hspace{-0.1cm} 37.5973  &\hspace{-0.1cm} 38.3961  &\\
    \hline
   \hspace{-0.4cm}\multirow{4}{*}{Dolls}&\hspace{-0.4cm}Proposed WT+CT 2-Stage &\hspace{-0.1cm} \textbf{29.5087}  &\hspace{-0.1cm} \textbf{32.5336}
                                        &\hspace{-0.1cm} \textbf{33.9974}  &\hspace{-0.1cm} \textbf{36.2741}
                                        &\hspace{-0.1cm} \textbf{37.6527} &\\
                                        &\hspace{-0.4cm}Proposed WT+CT Grid&\hspace{-0.1cm} 28.4858  &\hspace{-0.1cm} 29.0453
                                        &\hspace{-0.1cm} 30.0949  &\hspace{-0.1cm} 30.8123  &\hspace{-0.1cm} 31.6725  &\\
                                        &\hspace{-0.4cm}\cite{Hawe_Kleinsteuber_Diepold_2011} Grid &\hspace{-0.1cm} 28.4959  &\hspace{-0.1cm} N/A
                                        &\hspace{-0.1cm} N/A &\hspace{-0.1cm} N/A &\hspace{-0.1cm} 32.0521 &\\
                                        &\hspace{-0.4cm}Bicubic Grid &\hspace{-0.1cm} 29.0612  &\hspace{-0.1cm} 30.0475
                                        &\hspace{-0.1cm} 30.4374  &\hspace{-0.1cm} 31.0053  &\hspace{-0.1cm} 31.8800 &\\
    \hline
 \hspace{-0.4cm}\multirow{4}{*}{Moebius}&\hspace{-0.4cm}Proposed WT+CT 2-Stage &\hspace{-0.1cm} \textbf{31.0663} &\hspace{-0.1cm} \textbf{35.1060}
                                        &\hspace{-0.1cm} \textbf{37.7626} &\hspace{-0.1cm} \textbf{39.9225}
                                        &\hspace{-0.1cm} \textbf{41.8933} &\\
                                        &\hspace{-0.4cm}Proposed WT+CT Grid &\hspace{-0.1cm} 27.6882  &\hspace{-0.1cm} 28.7245
                                        &\hspace{-0.1cm} 29.8527  &\hspace{-0.1cm} 31.1663  &\hspace{-0.1cm} 32.2399  &\\
                                        &\hspace{-0.4cm}\cite{Hawe_Kleinsteuber_Diepold_2011} Grid &\hspace{-0.1cm} 27.6851  &\hspace{-0.1cm} 28.7973
                                        &\hspace{-0.1cm} N/A &\hspace{-0.1cm} N/A &\hspace{-0.1cm} 32.0990 &\\
                                        &\hspace{-0.4cm}Bicubic Grid &\hspace{-0.1cm} 28.3987 &\hspace{-0.1cm} 29.9338
                                        &\hspace{-0.1cm} 30.6607&\hspace{-0.1cm} 30.9427  &\hspace{-0.1cm} 32.0143 &\\
    \hline
 \hspace{-0.4cm}\multirow{4}{*}{Rocks}&\hspace{-0.4cm}Proposed WT+CT 2-Stage &\hspace{-0.1cm} \textbf{30.7662}
                                      &\hspace{-0.1cm} \textbf{35.3975} &\hspace{-0.1cm} \textbf{37.5056}
                                      &\hspace{-0.1cm} \textbf{40.4494} &\hspace{-0.1cm} \textbf{42.5089} &\\
                                      &\hspace{-0.4cm}Proposed WT+CT Grid&\hspace{-0.1cm} 25.5924 &\hspace{-0.1cm} 29.0848
                                      &\hspace{-0.1cm} 30.4766  &\hspace{-0.1cm} 31.2311  &\hspace{-0.1cm} 32.9218 &\\
                                      &\hspace{-0.4cm}\cite{Hawe_Kleinsteuber_Diepold_2011} Grid &\hspace{-0.1cm} 25.4444
                                      &\hspace{-0.1cm} 28.7973  &\hspace{-0.1cm} 29.5364
                                      &\hspace{-0.1cm} 30.2058  &\hspace{-0.1cm} 32.1672  &\\
                                      &\hspace{-0.4cm}Bicubic Grid &\hspace{-0.1cm} 28.7241 &\hspace{-0.1cm} 30.4212
                                      &\hspace{-0.1cm} 30.7552 &\hspace{-0.1cm} 31.6722 &\hspace{-0.1cm} 32.6706 &\\
  \hline
\end{tabular}
\end{center}
\end{table*}

\begin{table*}[ht!]
\caption{Comparisons of reconstruction algorithms in terms of \% Bad Pixel.}
\label{tab:P_Bad_Noiseless}
\scriptsize
\renewcommand{\arraystretch}{1.2}
\begin{center}
\begin{tabular}{cc|cccccc|cccccc|ccccc}
    \hline
   & Method  &   \multicolumn{5}{c}{ $\%$ of Bad Pixels [$\tau=1$]}
                            &\hspace{-0.6cm} &   \multicolumn{5}{c}{ $\%$ of Bad Pixels [$\tau=2$]}
                            &\hspace{-0.6cm} &   \multicolumn{5}{c}{ $\%$ of Bad Pixels [$\tau=3$]}\\
    \hline\hline
       Disparity & Algorithm  & \multicolumn{5}{c}{Percentage of Samples} &\hspace{-0.4cm} & \multicolumn{5}{c}{Percentage of Samples} &\hspace{-0.4cm} & \multicolumn{5}{c}{Percentage of Samples}
           \\
       Name & Sampling Strategy
       &\hspace{-0.0cm} 5$\%$
       &\hspace{-0.3cm} 10$\%$
       &\hspace{-0.3cm} 15$\%$
       &\hspace{-0.3cm} 20$\%$
       &\hspace{-0.3cm} 25$\%$
       &\hspace{-0.6cm}
       &\hspace{-0.0cm} 5$\%$
       &\hspace{-0.3cm} 10$\%$
       &\hspace{-0.3cm} 15$\%$
       &\hspace{-0.3cm} 20$\%$
       &\hspace{-0.3cm} 25$\%$
       &\hspace{-0.6cm}
       &\hspace{-0.0cm} 5$\%$
       &\hspace{-0.3cm} 10$\%$
       &\hspace{-0.3cm} 15$\%$
       &\hspace{-0.3cm} 20$\%$
       &\hspace{-0.3cm} 25$\%$
           \\
       \hline
       \hspace{-0.4cm}\multirow{4}{*}{Aloe}
           & Prop. WT+CT 2-Stage
           &\hspace{-0.3cm} 41.47
           &\hspace{-0.3cm} 21.37
           &\hspace{-0.3cm} 14.00
           &\hspace{-0.3cm} 8.85
           &\hspace{-0.3cm} 5.81
           &\hspace{-0.6cm}
           &\hspace{-0.3cm} \textbf{20.03}
           &\hspace{-0.3cm} \textbf{7.15}
           &\hspace{-0.3cm} \textbf{3.70}
           &\hspace{-0.3cm} \textbf{1.99}
           &\hspace{-0.3cm} \textbf{1.11}
           &\hspace{-0.6cm}
           &\hspace{-0.3cm} \textbf{13.42}
           &\hspace{-0.3cm} \textbf{4.80}
           &\hspace{-0.3cm} \textbf{2.52}
           &\hspace{-0.3cm} \textbf{1.43}
           &\hspace{-0.3cm} \textbf{0.79}
           \\
            & Prop. WT+CT Grid &\hspace{-0.3cm} 36.88 &\hspace{-0.3cm} 22.96 &\hspace{-0.3cm} 15.61 &\hspace{-0.3cm} 11.62 &\hspace{-0.3cm} 8.69
           &\hspace{-0.6cm}
                               &\hspace{-0.3cm} 21.16 &\hspace{-0.3cm} 10.11 &\hspace{-0.3cm} 6.87 &\hspace{-0.3cm} 5.17 &\hspace{-0.3cm} 3.92
           &\hspace{-0.6cm}
                               &\hspace{-0.3cm} 15.80 &\hspace{-0.3cm} 7.62 &\hspace{-0.3cm} 5.55 &\hspace{-0.3cm} 4.25 &\hspace{-0.3cm} 3.27
           \\
            & \cite{Hawe_Kleinsteuber_Diepold_2011} Grid &\hspace{-0.3cm} 31.44 &\hspace{-0.3cm} \textbf{17.65} &\hspace{-0.3cm} \textbf{11.58} &\hspace{-0.3cm} \textbf{8.39} &\hspace{-0.3cm} \textbf{5.79}
           &\hspace{-0.6cm}
                                                         &\hspace{-0.3cm} 20.12 &\hspace{-0.3cm} 8.87 &\hspace{-0.3cm} 6.08 &\hspace{-0.3cm} 4.73 &\hspace{-0.3cm} 3.56
           &\hspace{-0.6cm}
                                                         &\hspace{-0.3cm} 14.73 &\hspace{-0.3cm} 6.97 &\hspace{-0.3cm} 5.03 &\hspace{-0.3cm} 4.01 &\hspace{-0.3cm} 3.09
           \\
             & Bicubic Grid &\hspace{-0.3cm} \textbf{31.23} &\hspace{-0.3cm} 23.37 &\hspace{-0.3cm} 18.62 &\hspace{-0.3cm} 15.88 &\hspace{-0.3cm} 13.39
           &\hspace{-0.6cm}
                               &\hspace{-0.3cm} 23.51 &\hspace{-0.3cm} 17.49 &\hspace{-0.3cm} 13.78 &\hspace{-0.3cm} 11.96 &\hspace{-0.3cm} 10.04
           &\hspace{-0.6cm}
                               &\hspace{-0.3cm} 19.40 &\hspace{-0.3cm} 14.47 &\hspace{-0.3cm} 11.51 &\hspace{-0.3cm} 9.96 &\hspace{-0.3cm} 8.30
           \\
   \hline
        \hspace{-0.4cm}\multirow{4}{*}{Baby}
           & Prop. WT+CT 2-Stage &\hspace{-0.3cm} 28.00 &\hspace{-0.3cm} 12.37 &\hspace{-0.3cm} 5.72 &\hspace{-0.3cm} \textbf{2.67} &\hspace{-0.3cm} \textbf{1.31}
           &\hspace{-0.6cm}
                               &\hspace{-0.3cm} 9.95 &\hspace{-0.3cm} \textbf{2.31} &\hspace{-0.3cm}\textbf{0.39} &\hspace{-0.3cm} \textbf{0.13} &\hspace{-0.3cm} \textbf{0.07}
           &\hspace{-0.6cm}
                               &\hspace{-0.3cm} \textbf{3.69} &\hspace{-0.3cm} \textbf{0.64} &\hspace{-0.3cm} \textbf{0.16} &\hspace{-0.3cm} \textbf{0.03} &\hspace{-0.3cm} \textbf{0.01}
           \\
            & Prop. WT+CT Grid &\hspace{-0.3cm} 15.80 &\hspace{-0.3cm} 8.27 &\hspace{-0.3cm} 5.80 &\hspace{-0.3cm} 4.12 &\hspace{-0.3cm} 3.14
           &\hspace{-0.6cm}
                               &\hspace{-0.3cm} \textbf{6.31} &\hspace{-0.3cm} 3.01 &\hspace{-0.3cm} 2.22 &\hspace{-0.3cm} 1.58 &\hspace{-0.3cm} 1.22
           &\hspace{-0.6cm}
                               &\hspace{-0.3cm} 4.25 &\hspace{-0.3cm} 2.30 &\hspace{-0.3cm} 1.77 &\hspace{-0.3cm} 1.31 &\hspace{-0.3cm} 1.05
           \\
            & \cite{Hawe_Kleinsteuber_Diepold_2011} Grid &\hspace{-0.3cm} 12.31 &\hspace{-0.3cm} \textbf{6.02} &\hspace{-0.3cm} \textbf{3.93} &\hspace{-0.3cm} 2.71 &\hspace{-0.3cm} 1.86
           &\hspace{-0.6cm}
                                                         &\hspace{-0.3cm} 6.44 &\hspace{-0.3cm} 2.73 &\hspace{-0.3cm} 1.94 &\hspace{-0.3cm} 1.47 &\hspace{-0.3cm} 1.10
           &\hspace{-0.6cm}
                                                         &\hspace{-0.3cm} 4.21 &\hspace{-0.3cm} 2.09 &\hspace{-0.3cm} 1.55 &\hspace{-0.3cm} 1.21 &\hspace{-0.3cm} 0.93
           \\
             & Bicubic Grid &\hspace{-0.3cm} \textbf{12.22} &\hspace{-0.3cm} 8.53 &\hspace{-0.3cm} 6.54 &\hspace{-0.3cm} 5.59 &\hspace{-0.3cm} 4.58
           &\hspace{-0.6cm}
                               &\hspace{-0.3cm} 7.89 &\hspace{-0.3cm} 5.63 &\hspace{-0.3cm} 4.34 &\hspace{-0.3cm} 3.73 &\hspace{-0.3cm} 3.10
           &\hspace{-0.6cm}
                               &\hspace{-0.3cm} 6.24 &\hspace{-0.3cm} 4.42 &\hspace{-0.3cm} 3.51 &\hspace{-0.3cm} 3.00 &\hspace{-0.3cm} 2.41
           \\
   \hline
        \hspace{-0.4cm}\multirow{4}{*}{Rocks}
           & Prop. WT+CT 2-Stage &\hspace{-0.3cm} 25.90 &\hspace{-0.3cm} 10.67 &\hspace{-0.3cm} 6.27 &\hspace{-0.3cm} \textbf{3.55} &\hspace{-0.3cm} \textbf{2.19}
           &\hspace{-0.6cm}
                               &\hspace{-0.3cm} 8.26 &\hspace{-0.3cm} \textbf{2.26} &\hspace{-0.3cm} \textbf{0.93} &\hspace{-0.3cm} \textbf{0.41} &\hspace{-0.3cm} \textbf{0.22}
           &\hspace{-0.6cm}
                               &\hspace{-0.3cm} \textbf{4.75} &\hspace{-0.3cm} \textbf{1.22} &\hspace{-0.3cm} \textbf{0.51} &\hspace{-0.3cm} \textbf{0.21} &\hspace{-0.3cm} \textbf{0.10}
           \\
            & Prop. WT+CT Grid &\hspace{-0.3cm} 20.67 &\hspace{-0.3cm} 11.74 &\hspace{-0.3cm} 8.03 &\hspace{-0.3cm} 5.79 &\hspace{-0.3cm} 4.44
           &\hspace{-0.6cm}
                               &\hspace{-0.3cm} \textbf{7.64} &\hspace{-0.3cm} 4.12 &\hspace{-0.3cm} 2.93 &\hspace{-0.3cm} 2.34 &\hspace{-0.3cm} 1.72
           &\hspace{-0.6cm}
                               &\hspace{-0.3cm} 5.16 &\hspace{-0.3cm} 3.01 &\hspace{-0.3cm} 2.27 &\hspace{-0.3cm} 1.88 &\hspace{-0.3cm} 1.43
           \\
            & \cite{Hawe_Kleinsteuber_Diepold_2011} Grid &\hspace{-0.3cm} 16.36 &\hspace{-0.3cm} \textbf{9.09} &\hspace{-0.3cm} \textbf{6.10} &\hspace{-0.3cm} 4.38 &\hspace{-0.3cm} 3.00
           &\hspace{-0.6cm}
                                                         &\hspace{-0.3cm} 8.33 &\hspace{-0.3cm} 4.02 &\hspace{-0.3cm} 2.86 &\hspace{-0.3cm} 2.24 &\hspace{-0.3cm} 1.62
           &\hspace{-0.6cm}
                                                         &\hspace{-0.3cm} 5.52 &\hspace{-0.3cm} 2.93 &\hspace{-0.3cm} 2.19 &\hspace{-0.3cm} 1.76 &\hspace{-0.3cm} 1.26
           \\
             & Bicubic Grid &\hspace{-0.3cm} \textbf{15.32} &\hspace{-0.3cm} 11.51 &\hspace{-0.3cm} 9.36 &\hspace{-0.3cm} 7.88 &\hspace{-0.3cm} 6.59
           &\hspace{-0.6cm}
                               &\hspace{-0.3cm} 10.20 &\hspace{-0.3cm} 7.95 &\hspace{-0.3cm} 6.46 &\hspace{-0.3cm} 5.26 &\hspace{-0.3cm} 4.61
           &\hspace{-0.6cm}
                               &\hspace{-0.3cm} 8.28 &\hspace{-0.3cm} 6.51 &\hspace{-0.3cm} 5.24 &\hspace{-0.3cm} 4.42 &\hspace{-0.3cm} 3.76
           \\
   \hline
\end{tabular}
\end{center}
\end{table*}

\subsection{Real Data}
\begin{figure*}[ht!]
\small
  \centering
  \begin{tabular}{cccc}
    \includegraphics[width=0.19\linewidth]{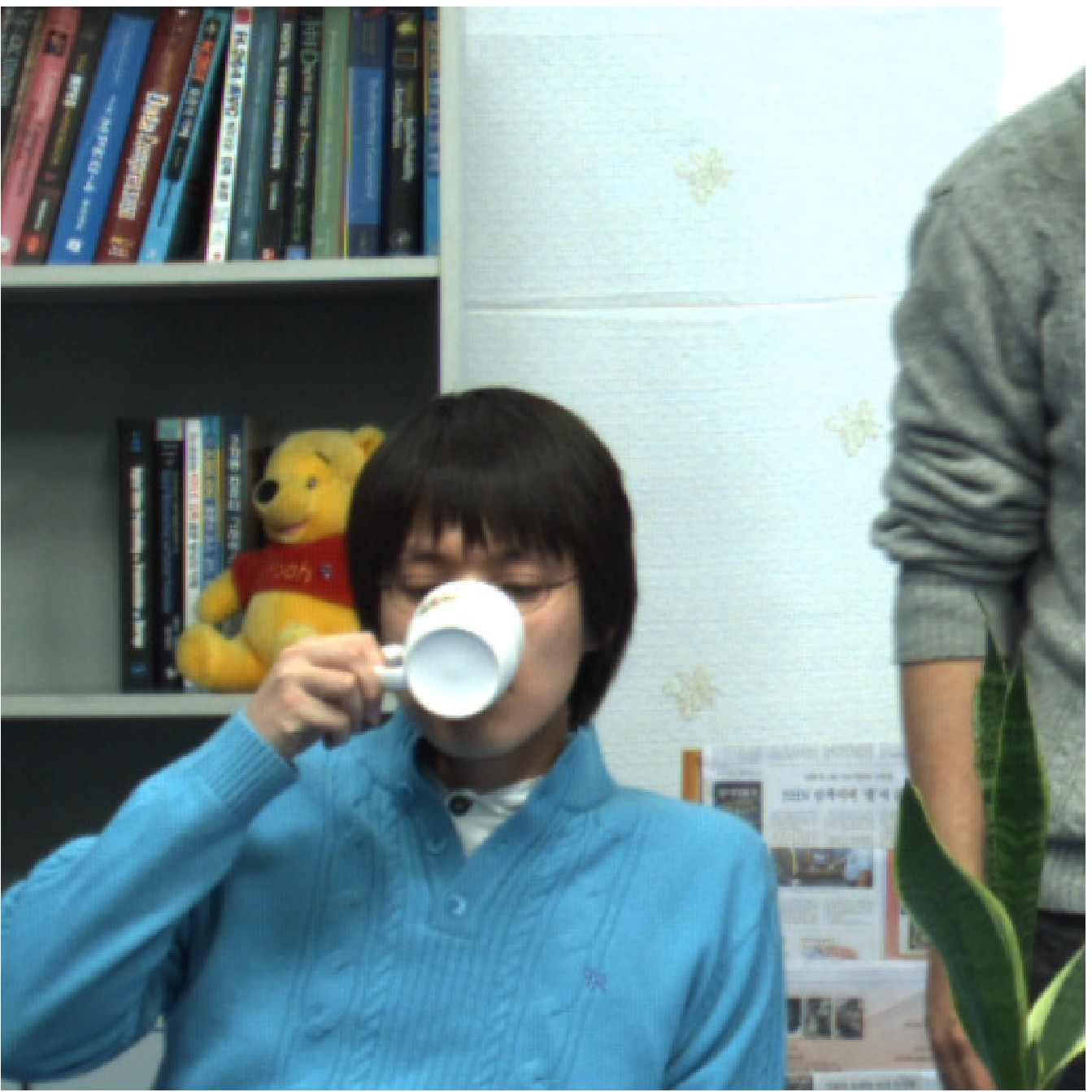}
    &\hspace{-0.38cm}\includegraphics[width=0.19\linewidth]{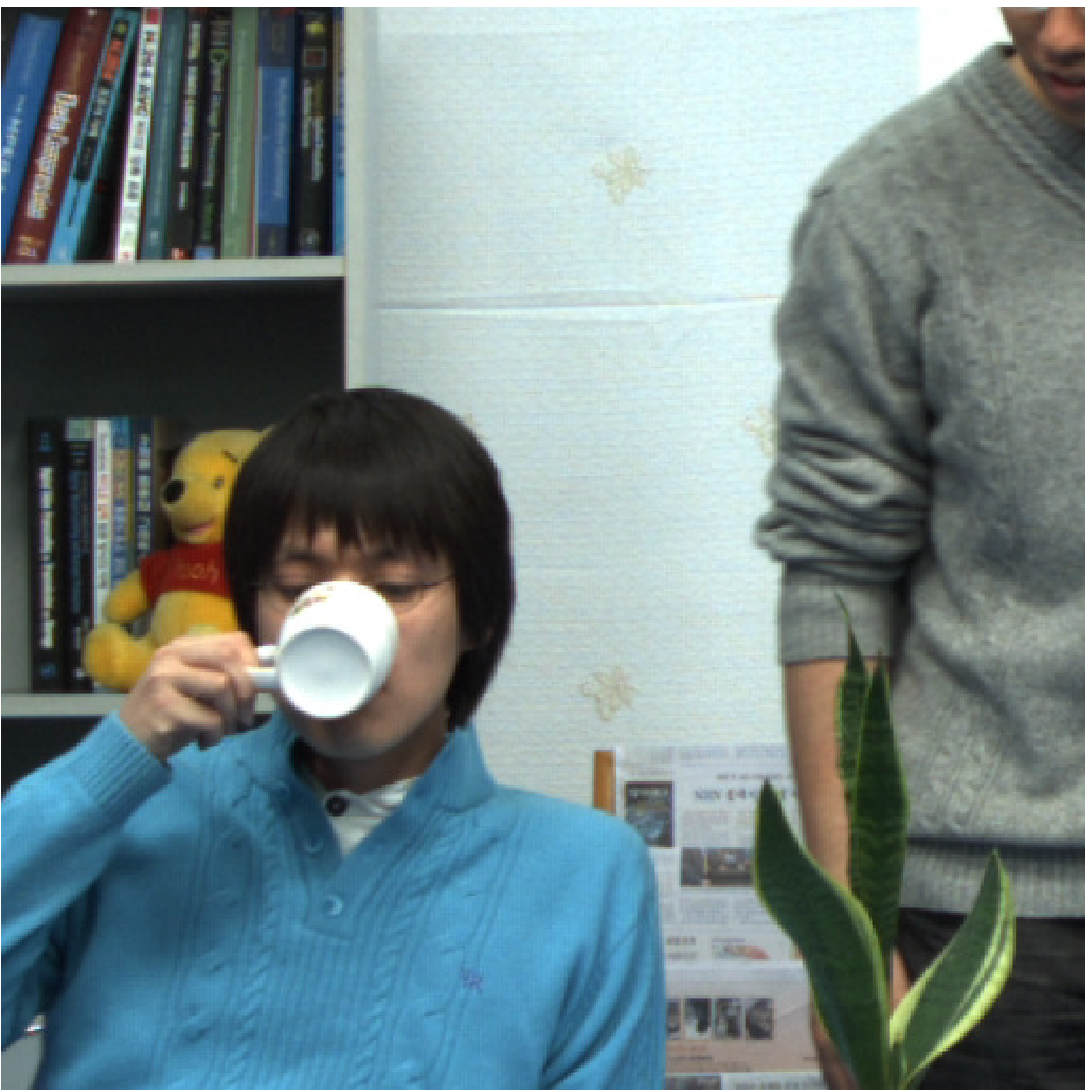}
    &\hspace{-0.4cm}\includegraphics[width=0.255\linewidth]{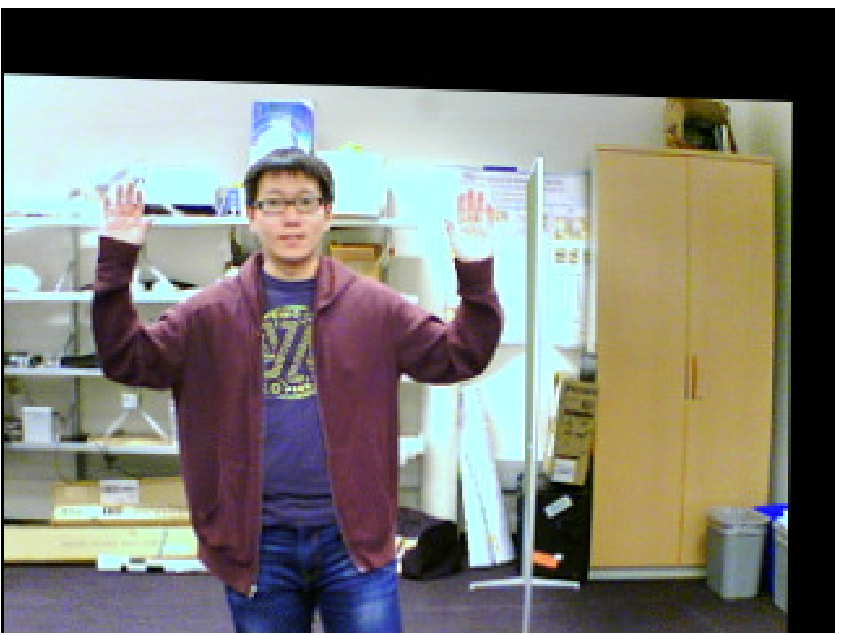}
    &\hspace{-0.42cm}\includegraphics[width=0.255\linewidth]{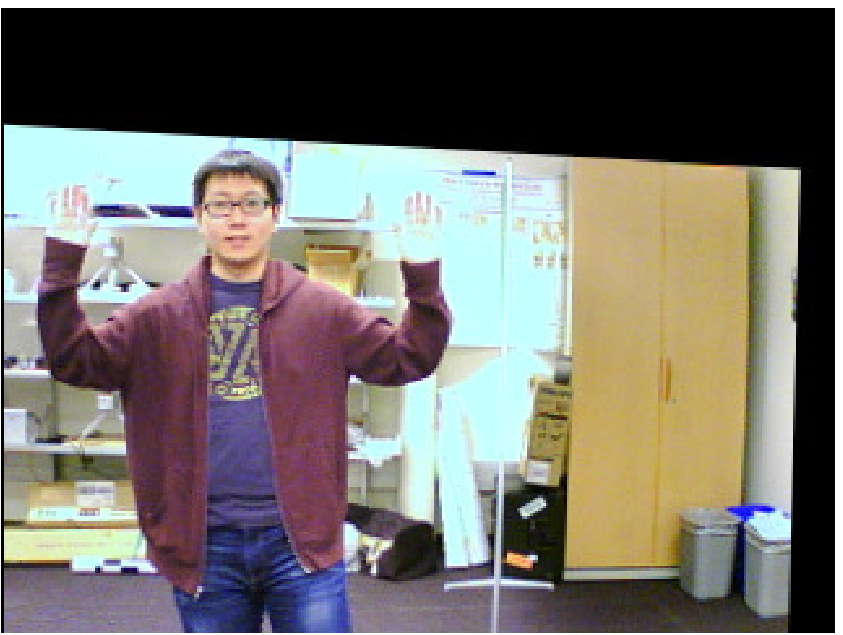}\\
     Left View  & Right View  & Left View & Right View \\
   \end{tabular}
   \begin{tabular}{ccccc}
    \hspace{-0.1cm}\includegraphics[width=0.195\linewidth]{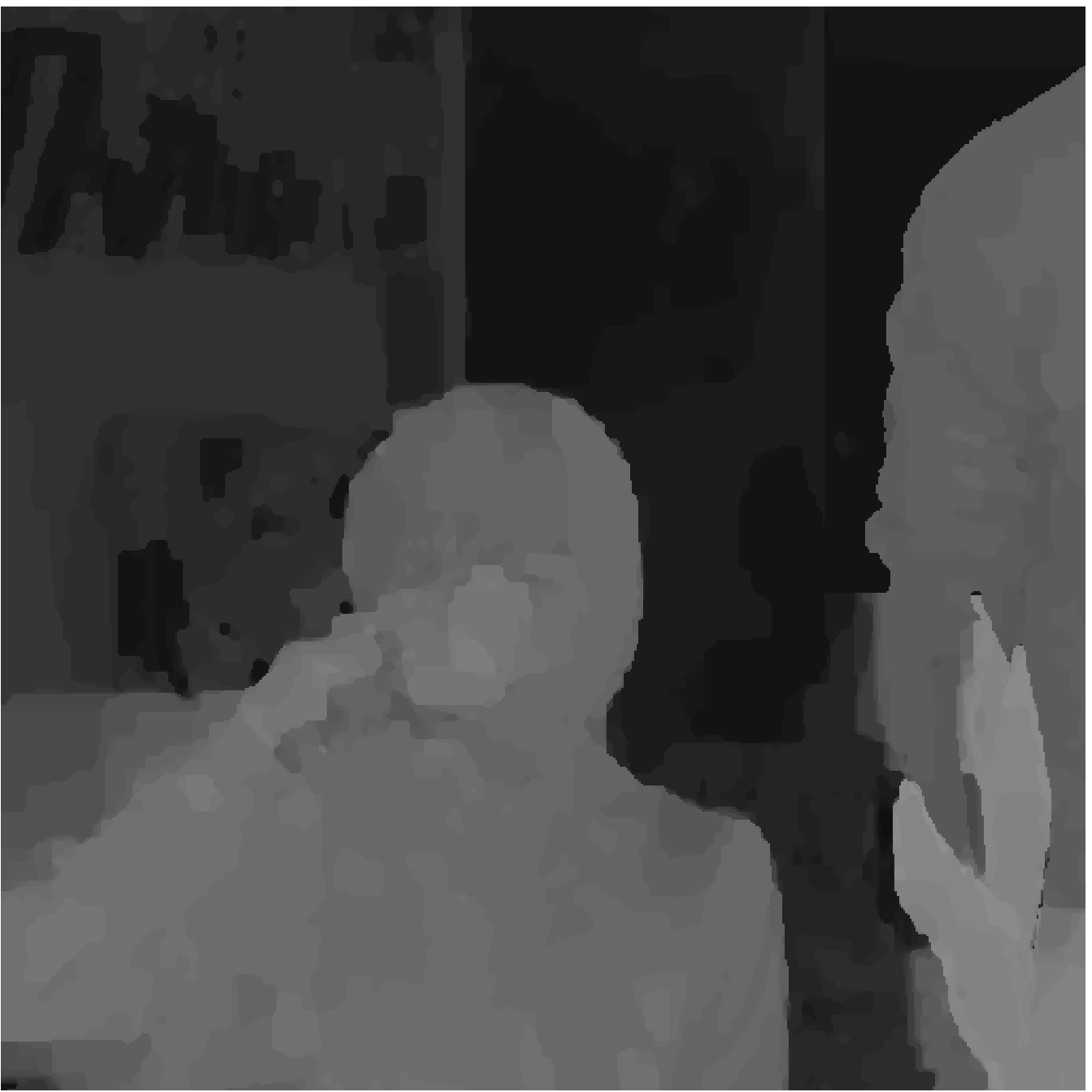}
    &\hspace{-0.4cm}\includegraphics[width=0.195\linewidth]{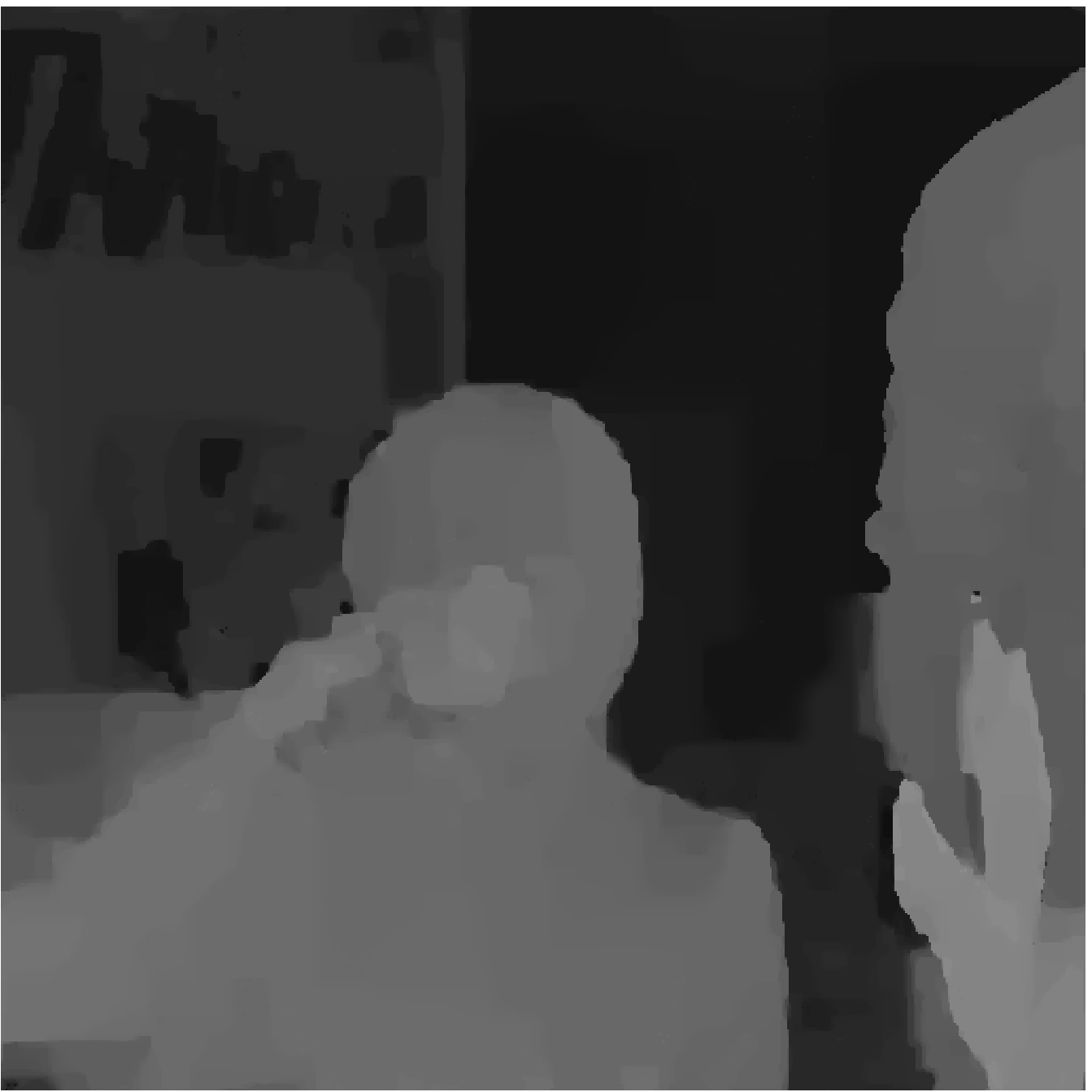}
    &\hspace{-0.4cm}\includegraphics[width=0.195\linewidth]{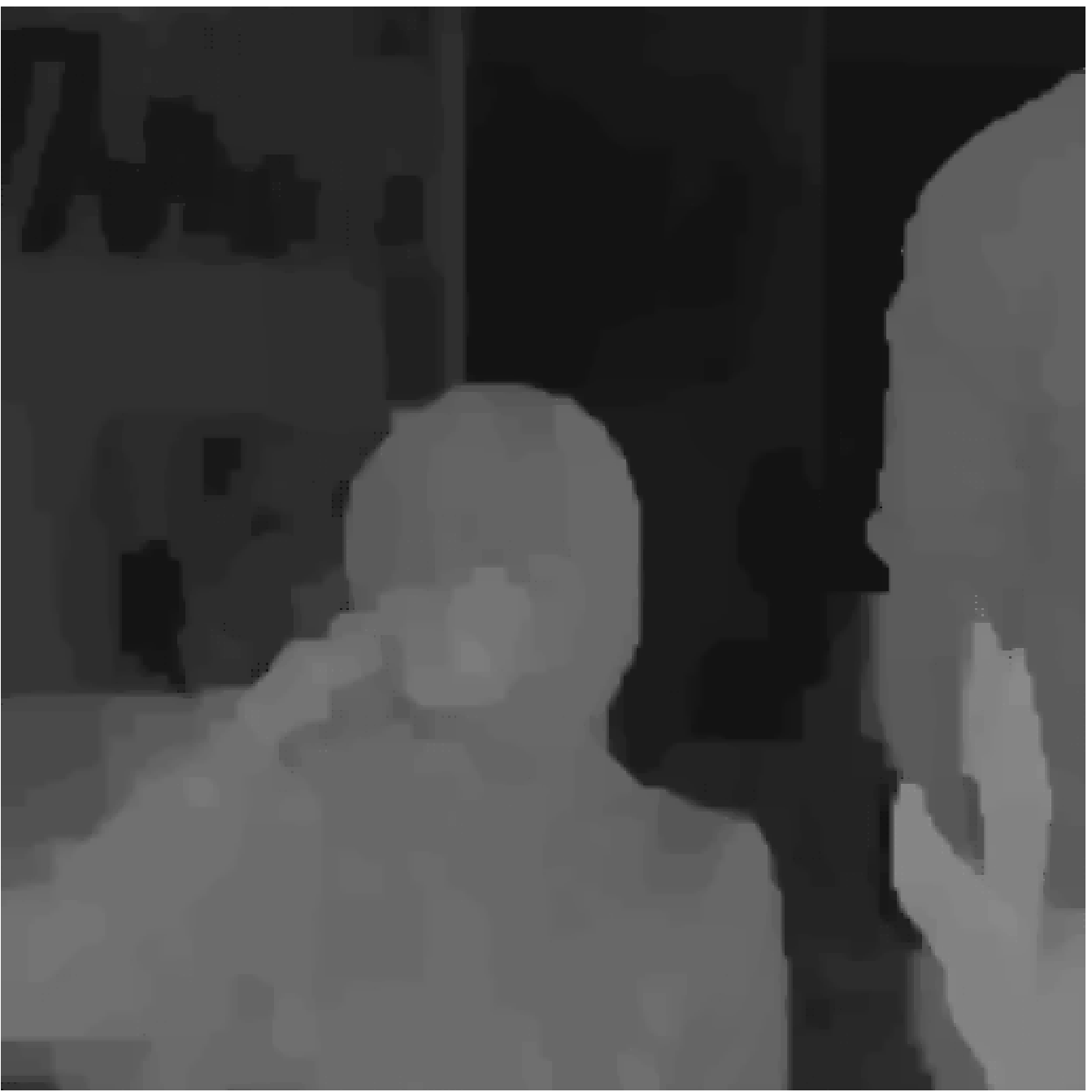}
    &\hspace{-0.4cm}\includegraphics[width=0.195\linewidth]{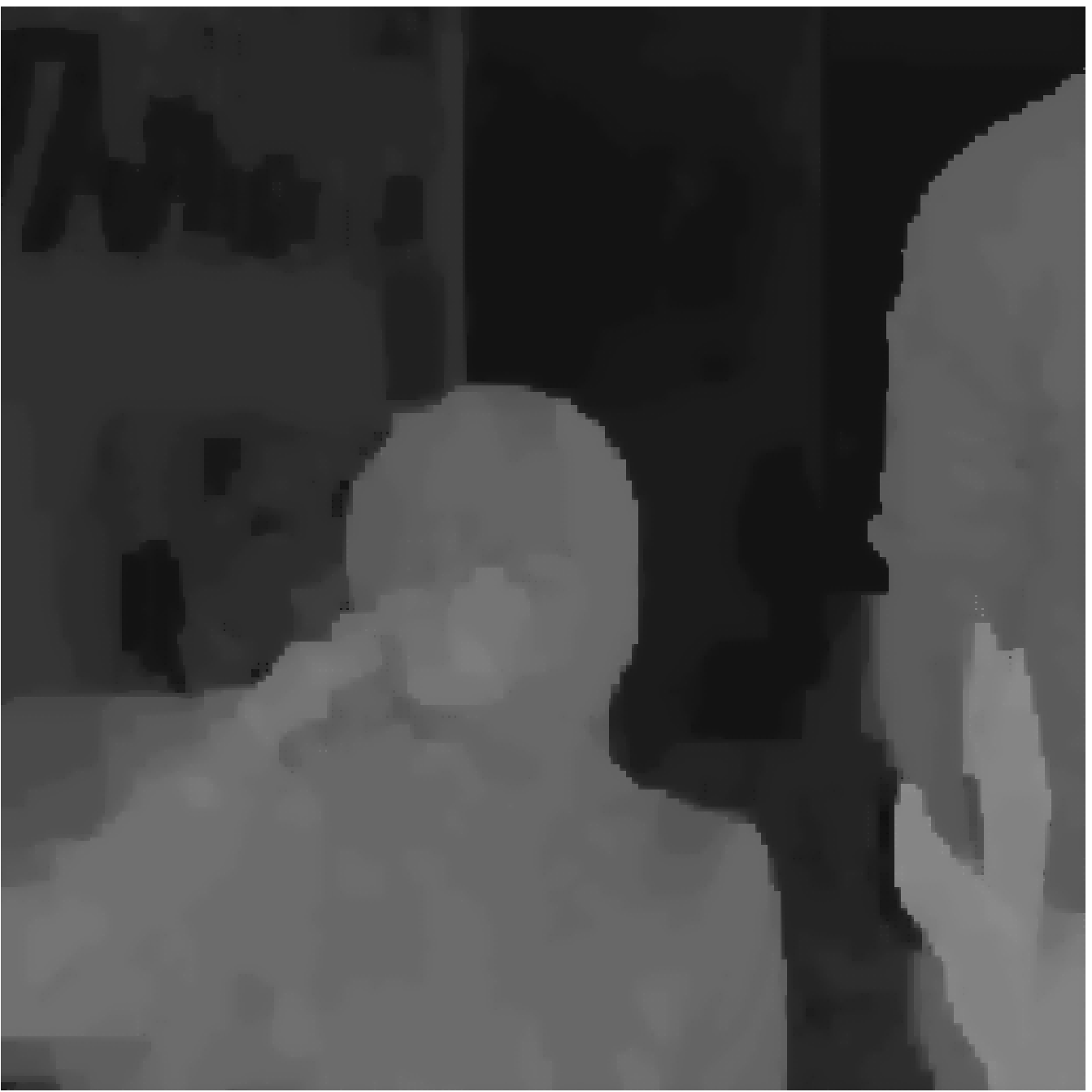}
    &\hspace{-0.4cm}\includegraphics[width=0.195\linewidth]{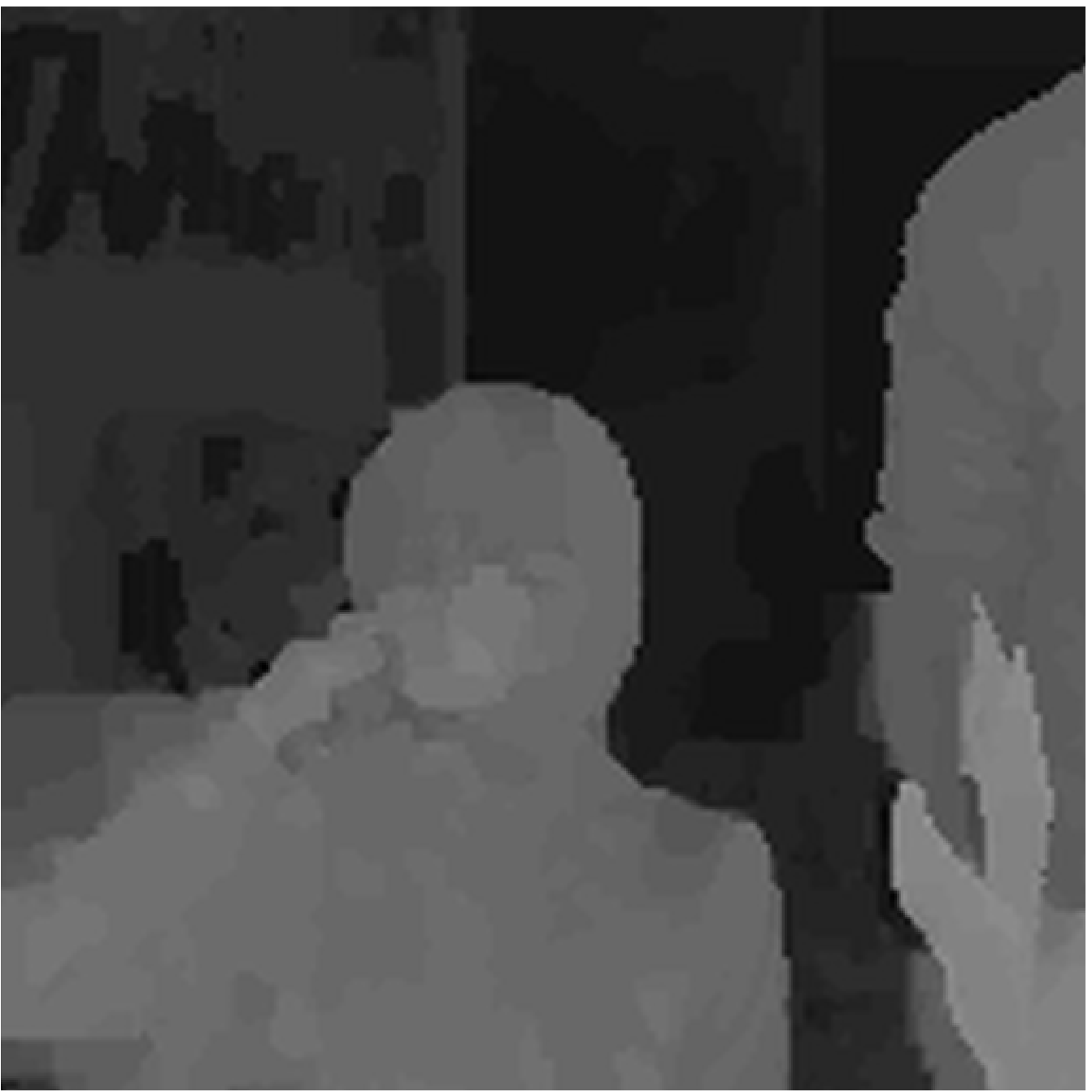}\\
    \hspace{-0.1cm}\includegraphics[width=0.197\linewidth]{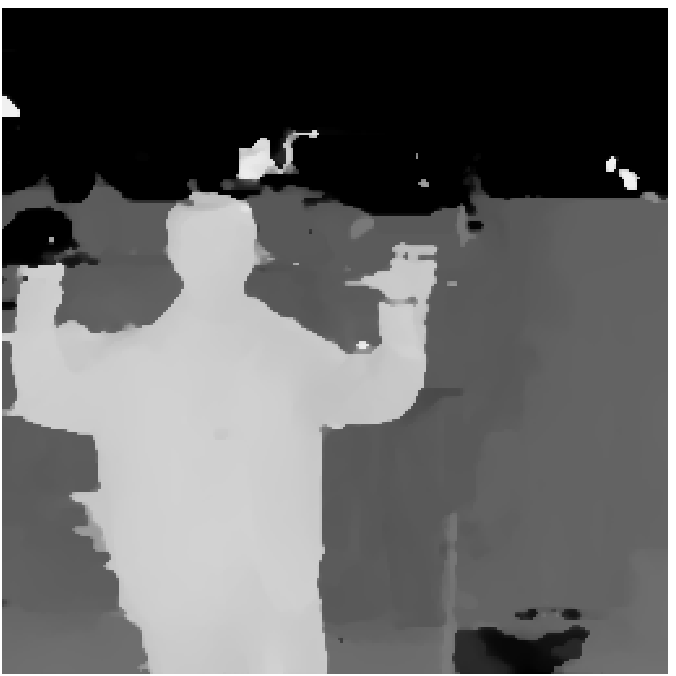}
    &\hspace{-0.4cm}\includegraphics[width=0.197\linewidth]{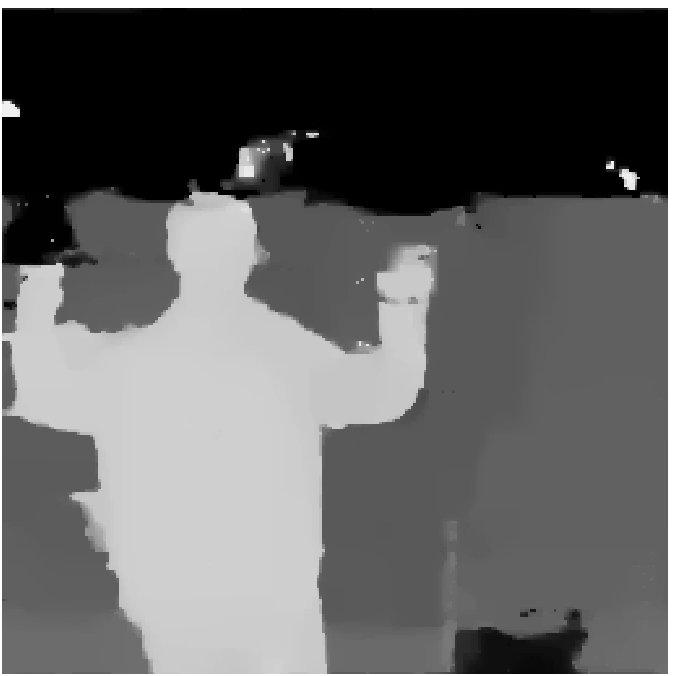}
    &\hspace{-0.4cm}\includegraphics[width=0.197\linewidth]{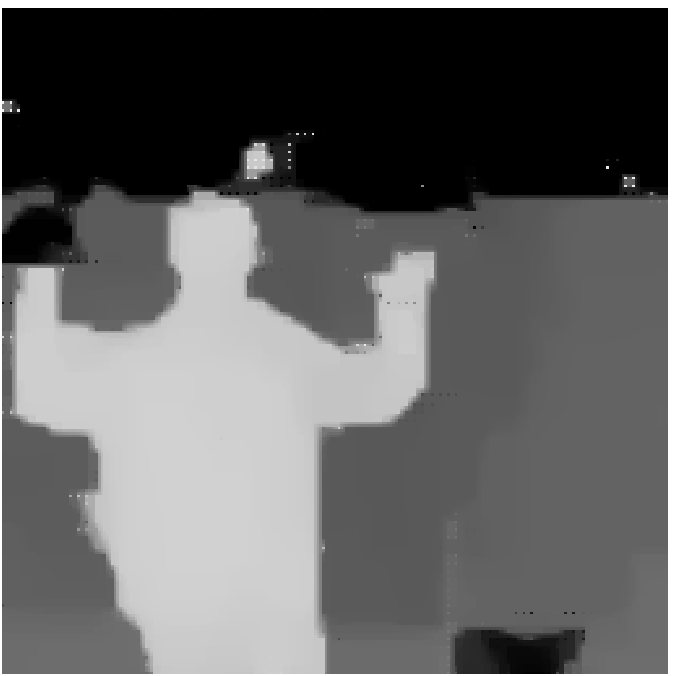}
    &\hspace{-0.4cm}\includegraphics[width=0.197\linewidth]{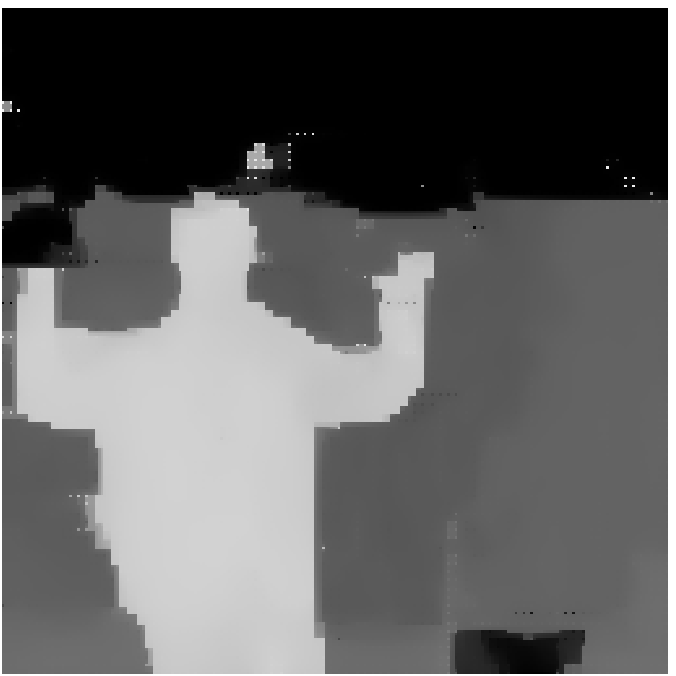}
    &\hspace{-0.4cm}\includegraphics[width=0.197\linewidth]{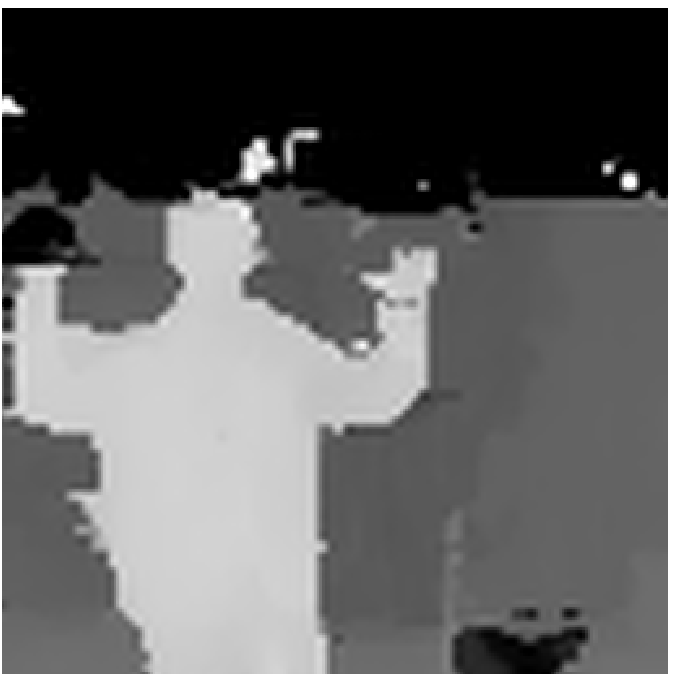}\\
    \hspace{-0.1cm}  Dense Estimation \cite{Lee_Juang_Nguyen_2013} & \hspace{-0.45cm} Proposed WT+CT 2-Stage & \hspace{-0.4cm} Proposed WT+CT Grid & \hspace{-0.45cm} \cite{Hawe_Kleinsteuber_Diepold_2011} Grid & \hspace{-0.4cm} Bicubic Grid\\
   \end{tabular}
    \caption{Examples of disparity map reconstruction from 10\% measured samples using real data. [Top] Left and right view images of the ``Newspaper'' dataset, and a sequence captured by a stereo system we developed. [Middle] The reconstructed disparity maps of ``Newspaper''. [Bottom] The reconstructed disparity maps of our sequence. For the reconstructed disparity maps, we show the zoom-in results of size $256 \times 256$ for better visualization. Methods under comparisons include: a dense disparity estimation \cite{Lee_Juang_Nguyen_2013} to acquire initial estimate; ``Proposed WT+CT 2-Stage'' which applies the 2-Stage randomized scheme to determine sampling locations; ``Proposed WT+CT Grid'' which picks samples from a uniform grid; ``\cite{Hawe_Kleinsteuber_Diepold_2011} Grid'' which applies a subgradient algorithm to samples picked from a uniform grid; ``Bicubic Grid'' which applies bicubic interpolation to samples picked from a uniform grid.}
    \label{fig:Real_Data_Experiment1}
\end{figure*}

In this experiment we study the performance of the proposed algorithm for real data. The top left part of Figure \ref{fig:Real_Data_Experiment1} shows a snapshot of a stereo video (with resolution $320\times 240$, 30 fps). For this video sequence, we apply the block matching algorithm by Lee et al. \cite{Lee_Juang_Nguyen_2013} to obtain the initial disparity estimates. However, instead of computing the \textit{full disparity map}, we only compute 10\% of the disparity pixel values and use the proposed reconstruction algorithm to recover the dense disparity map. The 10\% samples are selected according to the two stages of ``Proposed WT+CT 2-Stage''. In the first stage, we select the locations of the 5\% samples using our oracle random sampling scheme with PCA improvement applied to the color image. A pilot estimate of the disparity is thus computed and the remaining 5\% samples can be located according to the second stage of ``Proposed WT+CT 2-Stage''. The results shown in the middle row of \fref{fig:Real_Data_Experiment1} illustrate that the ``Proposed WT+CT 2-Stage'' generates the closest disparity maps compared to an ideal dense estimate.


In addition to real video sequences, we also test the proposed algorithm on a stereo system we developed. The system consists of a low cost stereo camera with customized block matching algorithms. The bottom row of \fref{fig:Real_Data_Experiment1} shows the results of the reconstructed disparity maps. Referring to the results of ``\cite{Hawe_Kleinsteuber_Diepold_2011} Grid'' and ``Bicubic Grid'', we note that there are serious stair-like artifacts located at object boundaries. In contrast, the two proposed methods in general produce much smoother object boundaries, thanks to the superior modeling and the optimized sampling scheme. More interestingly, we observe that ``Proposed WT+CT 2-Stage'' indeed removes some undesirable noisy estimates in the recovered disparity maps. This shows that the proposed method could potentially further developed as a depth enhancement method.

\section{Conclusion}
A framework for dense depth reconstruction from sparse samples is presented in this paper. Three contributions are made. First, we provide empirical evidence that depth data can be more sparsely encoded by a combination of wavelet and contourlet dictionaries. This provides a better understanding of the structures of depth data. Second, we propose a general optimization formulation. An alternating direction method of multipliers (ADMM) with a multi-scale warm start is proposed to achieve fast reconstruction. The ADMM algorithm achieves faster rate of convergence than the existing subgradient descent algorithms. Third, we propose an efficient method to select samples by a randomized sampling scheme. The proposed sampling scheme achieves high quality reconstruction results at a given sampling budget. The new tools developed in this paper are applicable to many depth data processing tasks, with applications in acquisition, compression, and enhancement. Future work shall focus on extending the methods to space-time data volume to further improve consistency of the estimates. 
\bibliographystyle{IEEEbib}
\bibliography{ref}
\end{document}


\maketitle
\begin{abstract}
The purpose of this supplementary report is to present experimental results on the selecting parameters for the proposed depth reconstruction algorithm.
\end{abstract}

\section{Depth Reconstruction Algorithm}
For completeness we first recall the reconstruction algorithm. Referring to the main article, the optimization problem that we would like to solve is
\begin{equation}
\minimize{\vx} \;\; \frac{1}{2}\| \mS \vx - \vb \|_2^2 + \sum_{\ell=1}^L \lambda_{\ell} \|\mW_{\ell}\mPhi_{\ell}^T\vx\|_1 + \beta\|\vx\|_{TV},
\label{eq:problem_P1}
\end{equation}
where $\mPhi_{\ell}$ is a dictionary, $\mW_{\ell}$ is the corresponding weighting matrix, $\mS$ is the sampling pattern, and the vector $\vb$ is the observation. We consider $L=2$ dictionaries, where $\mPhi_1$ and $\mPhi_2$ are the wavelet and contourlet dictionaries, respectively. 

To solve \eref{eq:problem_P1}, we apply the alternating direction method of multipliers (ADMM). The ADMM algorithm tries to find the stationary point of the augmented Lagrangian function, defined as
\begin{align}
\calL\left(\vx,\vu_{1},\vu_{2},\vr,\vv,\vw,\vy_{1},\vy_{2},\vz\right)
&=\frac{1}{2}\|\vb - \mS\vr \|^2 + \lambda_{1}\|\mW_{1}\vu_{1}\|_1 + \lambda_{2}\|\mW_{2}\vu_{2}\|_1 + \beta\|\vv\|_1 \label{eq:ADMM_Lagrangian} \\
&\quad-\vw^{T}\left(\vr - \vx \right) - \vy_{1}^{T}\left(\vu_{1} - \mPhi_{1}^T\vx \right) - \vy_{2}^{T}\left(\vu_{2} - \mPhi_{2}^T\vx \right) - \vz^{T}\left(\vv - \mD\vx \right) \notag\\
&\quad+ \frac{\mu}{2}\|\vr - \vx \|^2 + \frac{\rho_{1}}{2}\|\vu_{1} - \mPhi_{1}^T\vx \|^2 + \frac{\rho_{2}}{2}\|\vu_{2} - \mPhi_{2}^T\vx \|^2 + \frac{\gamma}{2}\|\vv - \mD\vx \|^2. \notag
\end{align}
Note that in \eref{eq:ADMM_Lagrangian}, we need to specify the regularization parameters ($\lambda_1, \lambda_2, \beta$) and internal parameters ($\mu, \rho_1, \rho_2, \gamma$). The purpose of this supplementary report is to discuss how the parameters are chosen.

\section{Parameter Selection and Experiments}
We now present how to choose the regularization parameters ($\lambda_1, \lambda_2, \beta$), and the internal parameters ($\mu,\rho_1, \rho_2, \gamma$).

\subsection{Experimental Configurations}
\begin{figure}[ht!]
\centering
\begin{tabular}{cccccc}
\includegraphics[width=0.15\textwidth]{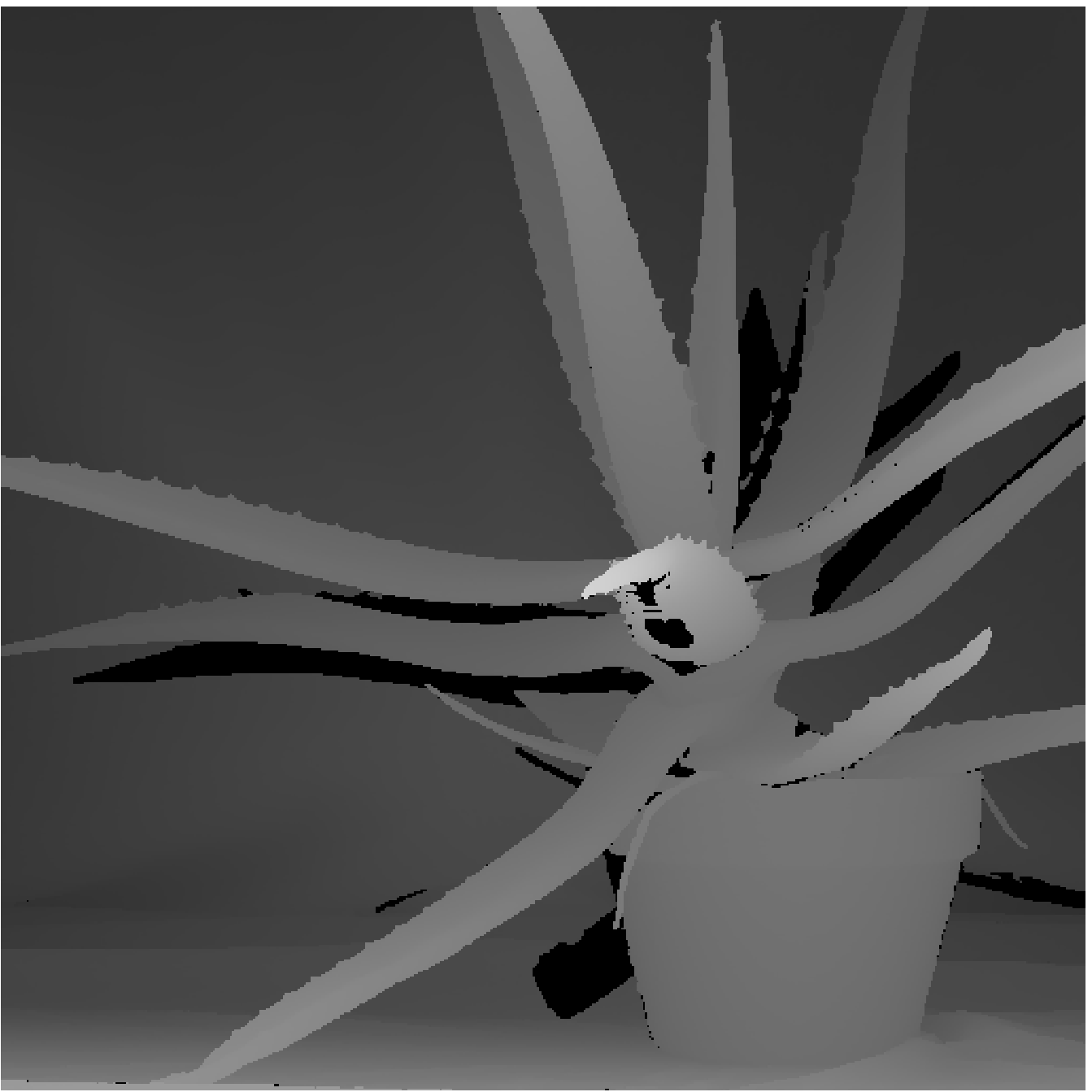} \hspace{-10pt}
&\includegraphics[width=0.15\textwidth]{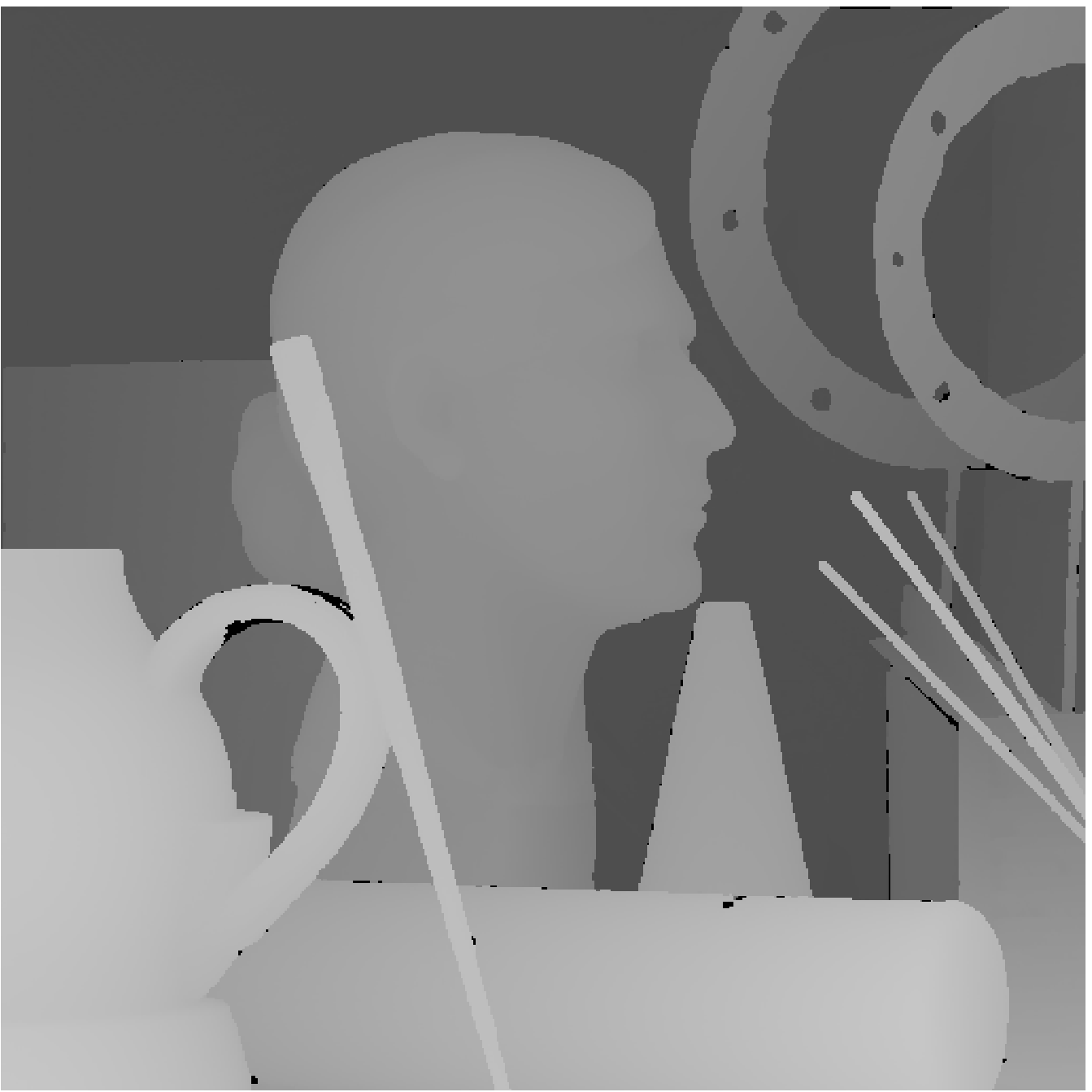} \hspace{-10pt}
&\includegraphics[width=0.15\textwidth]{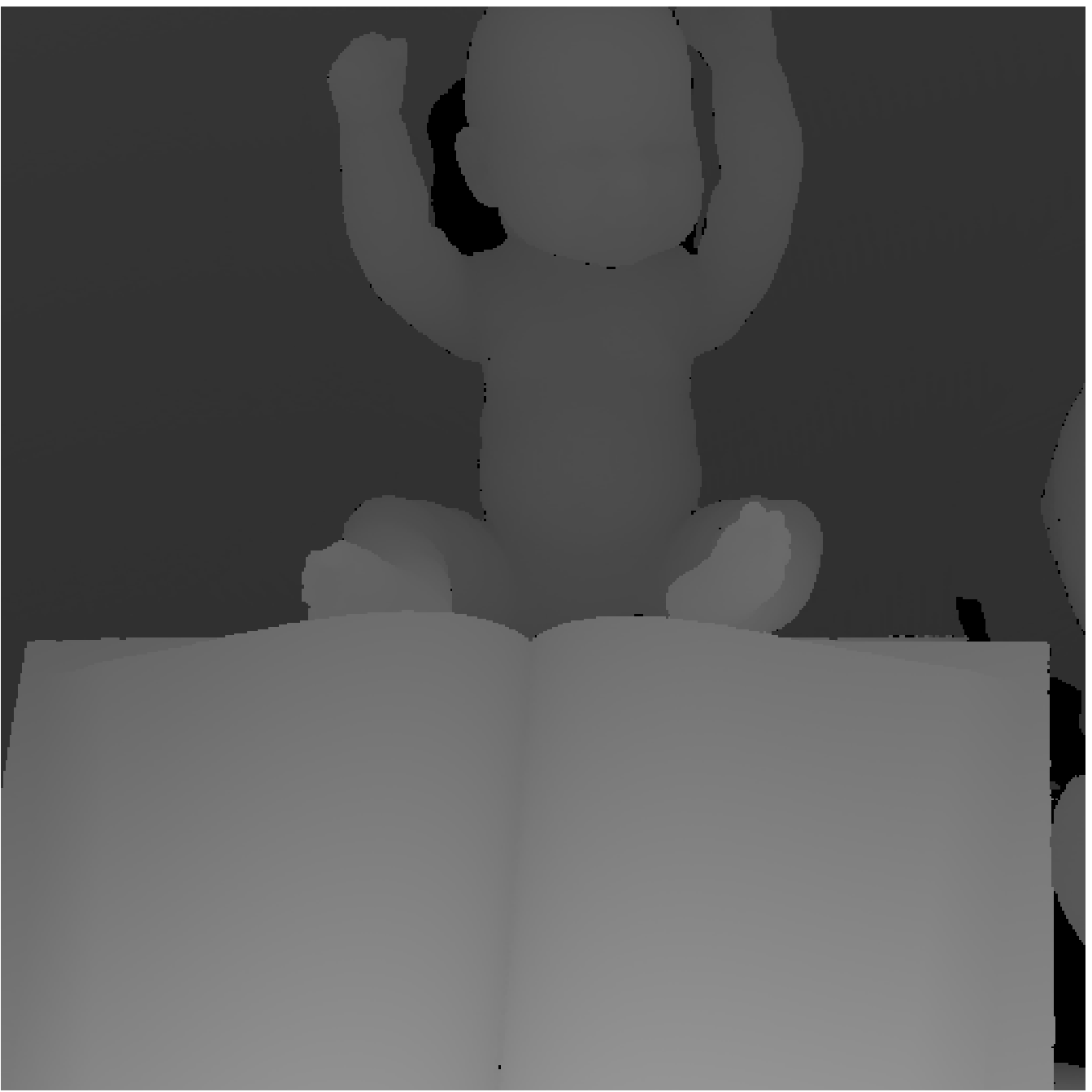} \hspace{-10pt}
&\includegraphics[width=0.15\textwidth]{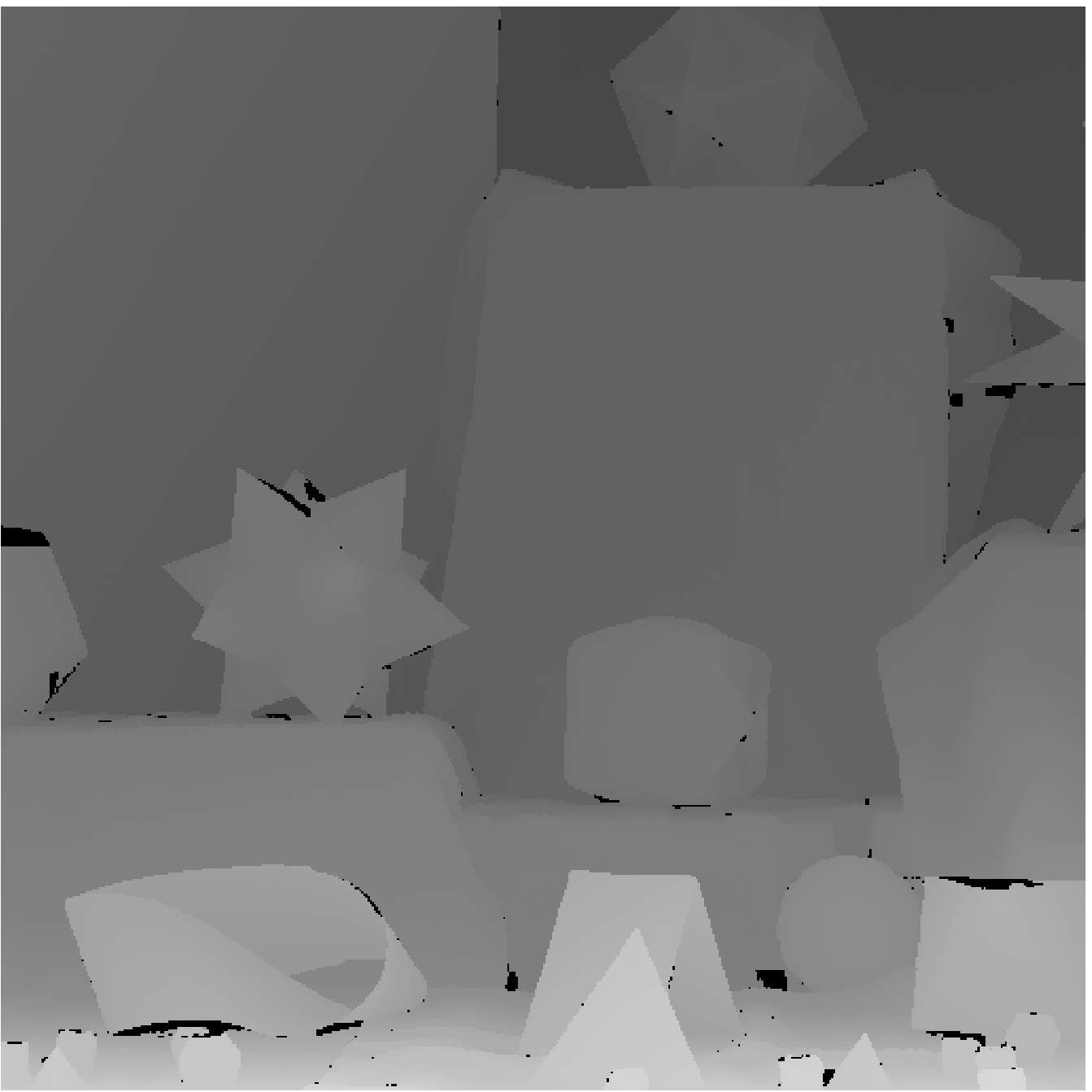} \hspace{-10pt}
&\includegraphics[width=0.15\textwidth]{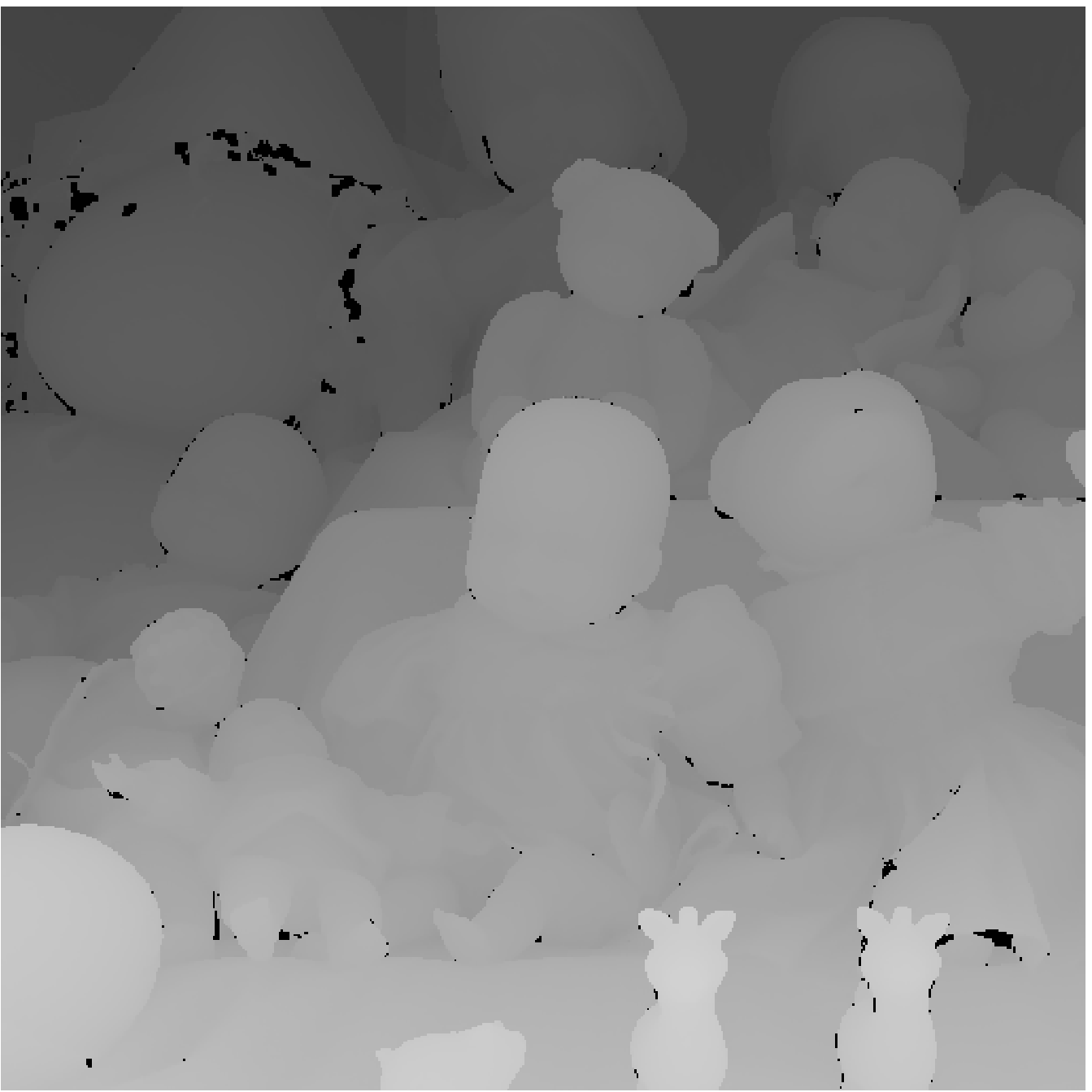} \hspace{-10pt}
&\includegraphics[width=0.15\textwidth]{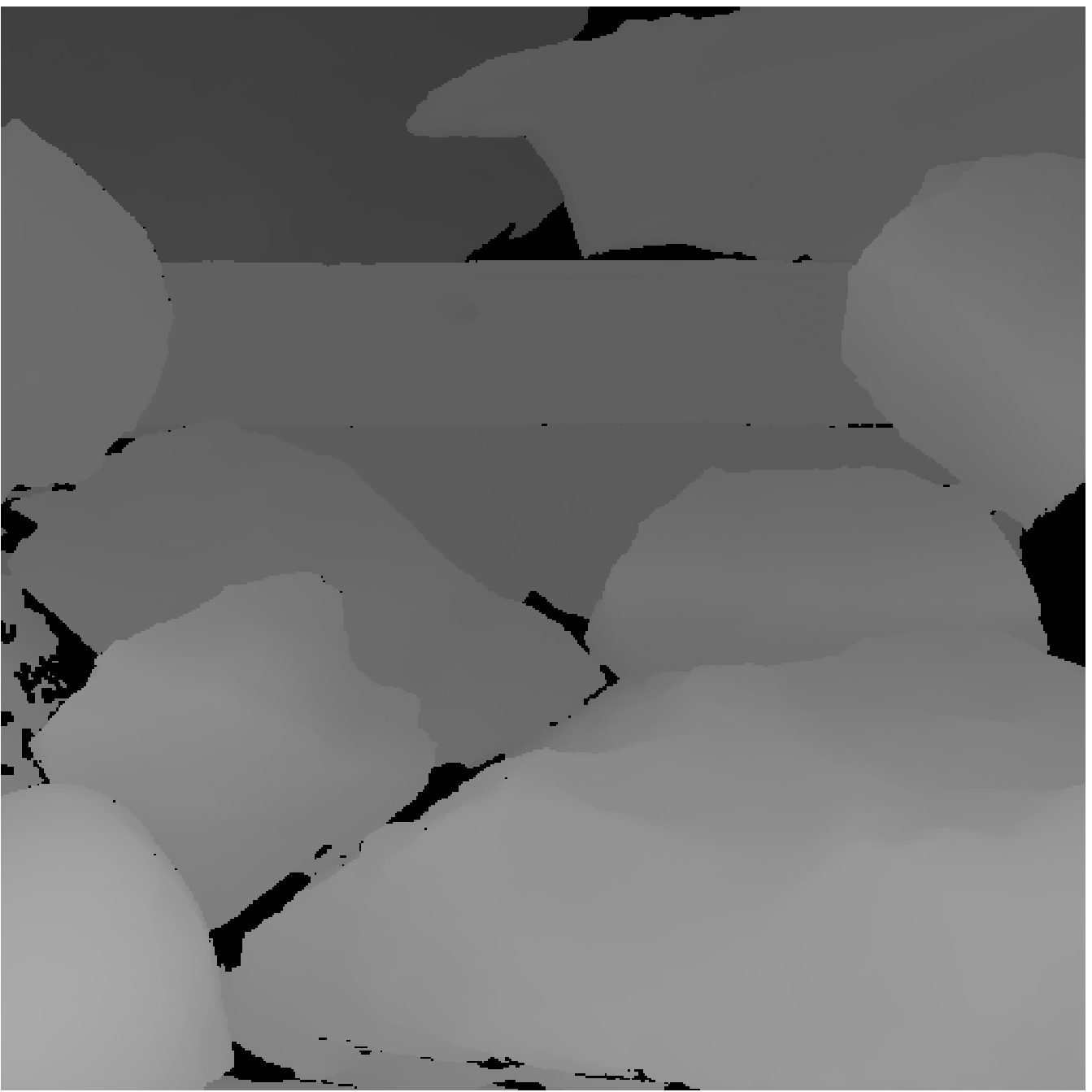}\\
Aloe \hspace{-10pt}& Art \hspace{-10pt}& Baby2 \hspace{-10pt} & Moebius \hspace{-10pt} & Dolls\hspace{-10pt} & Rocks1 \\
\end{tabular}
\caption{Example disparity maps from Middlebury dataset.}
\label{fig:data_example}
\end{figure}

Before presenting results, we first describe our experimental configurations. Testing disparity maps are chosen from Middlebury datasets \footnote{http://vision.middlebury.edu/stereo/data/}. All disparity values are normalized to the range $\left[0,\,1\right]$. \fref{fig:data_example} shows some examples of disparity maps. For the sampling patterns, we choose the uniformly random samples to minimize any bias towards the sampling. For wavelet dictionary, we use ``db2'' wavelet function with decomposition level 2, and for contourlet dictionary, we set frequency partition ``5, 6''. These settings are fixed throughout the experiment.

\subsection{Regularization Parameters ($\lambda_1, \lambda_2, \beta$)}
We empirically evaluate the mean square error (MSE) by sweeping the parameters $(\lambda_1,\lambda_2,\beta)$ from $10^{-6}$ to  $10^{0}$, with a fixed sampling rate of $\xi = 0.2$. The optimal values of the parameters are chosen to minimize the average MSE.

\begin{figure}[h!]
\centering
\begin{tabular}{ccc}
\includegraphics[width=0.31\textwidth]{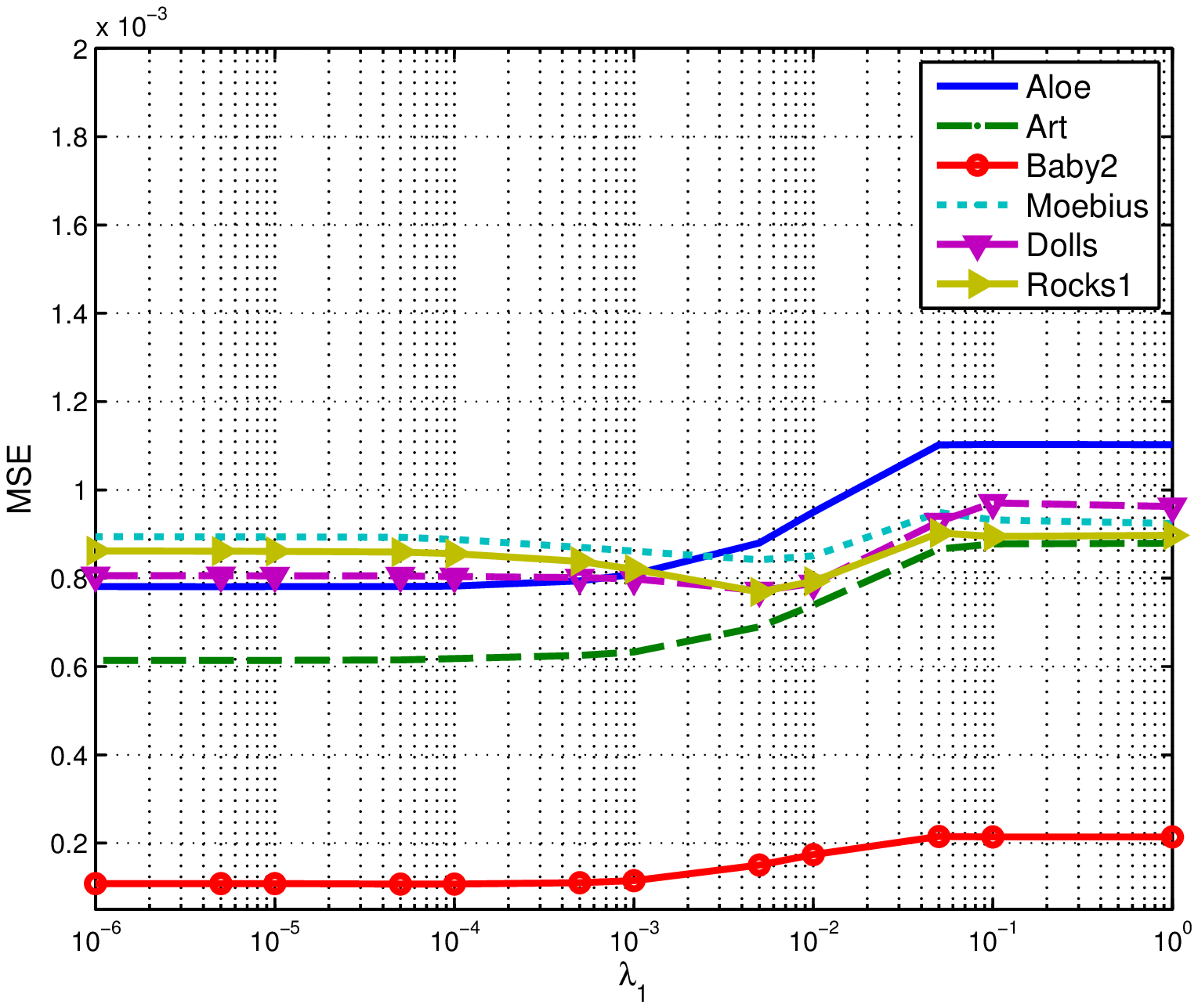} \hspace{-10pt}
&\includegraphics[width=0.31\textwidth]{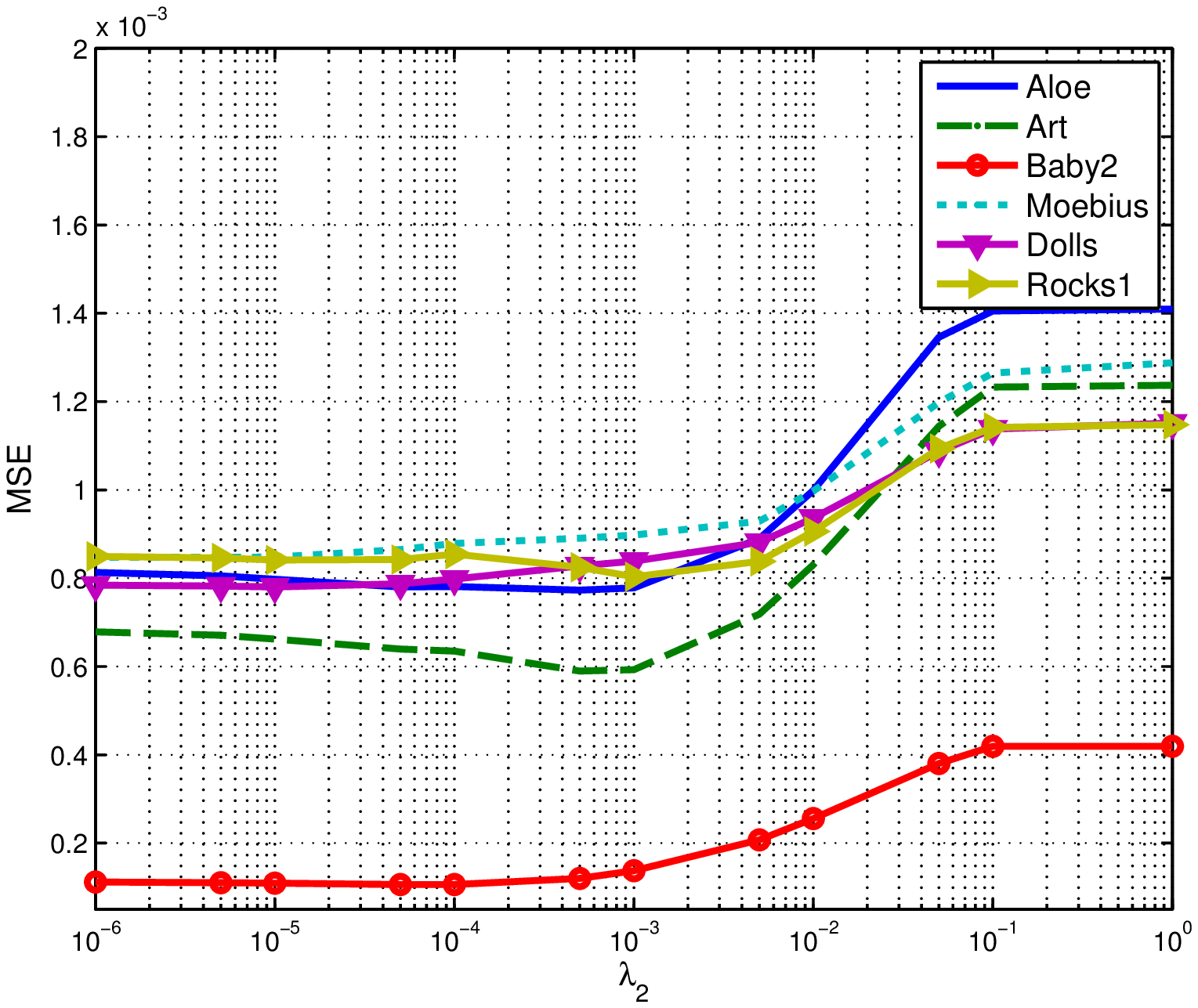} \hspace{-10pt}
&\includegraphics[width=0.31\textwidth]{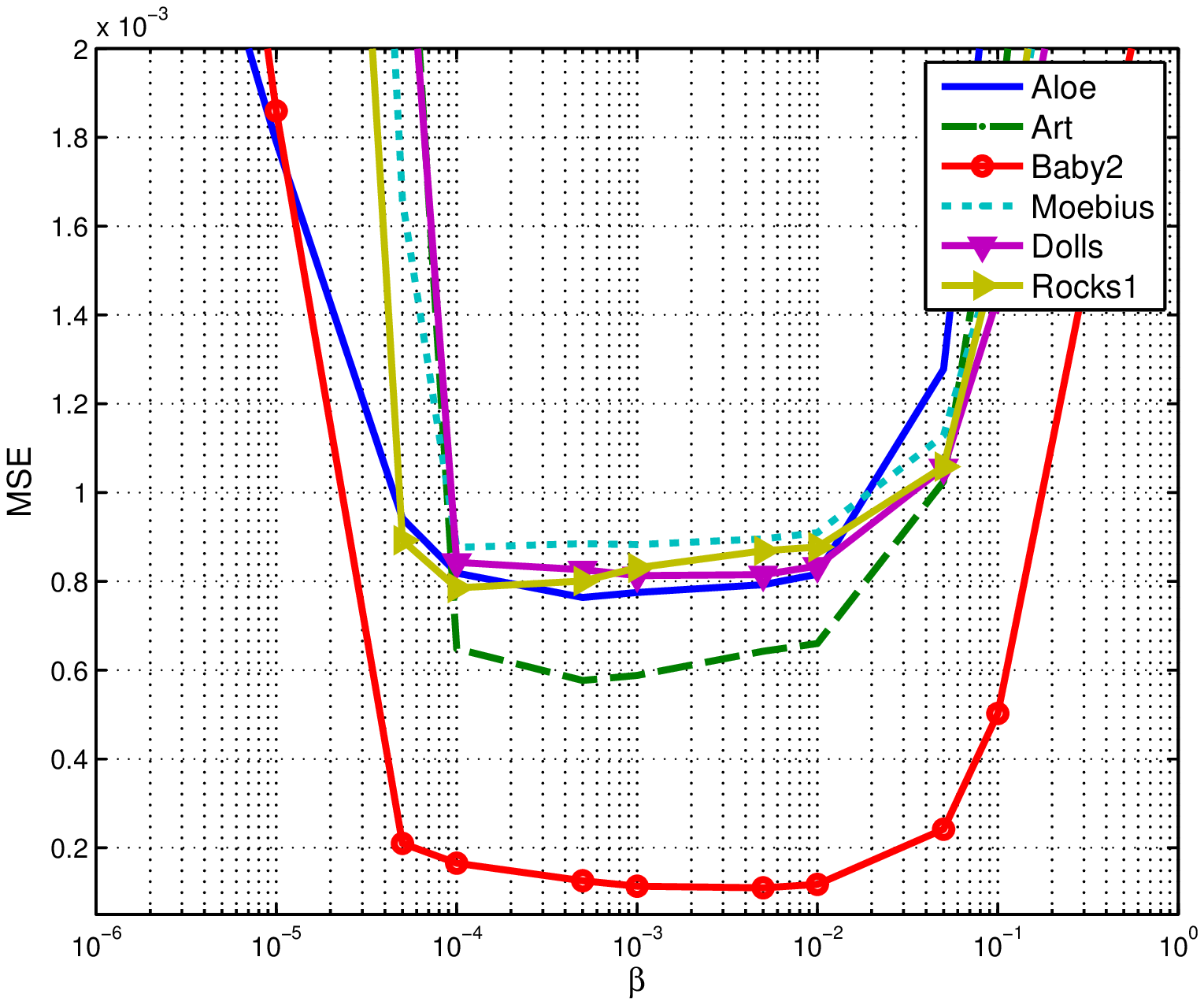} \\
$\lambda_1,\xi=0.2$ \hspace{-10pt}& $\lambda_2,\xi=0.2$ \hspace{-10pt}& $\beta,\xi=0.2$ \\
\includegraphics[width=0.31\textwidth]{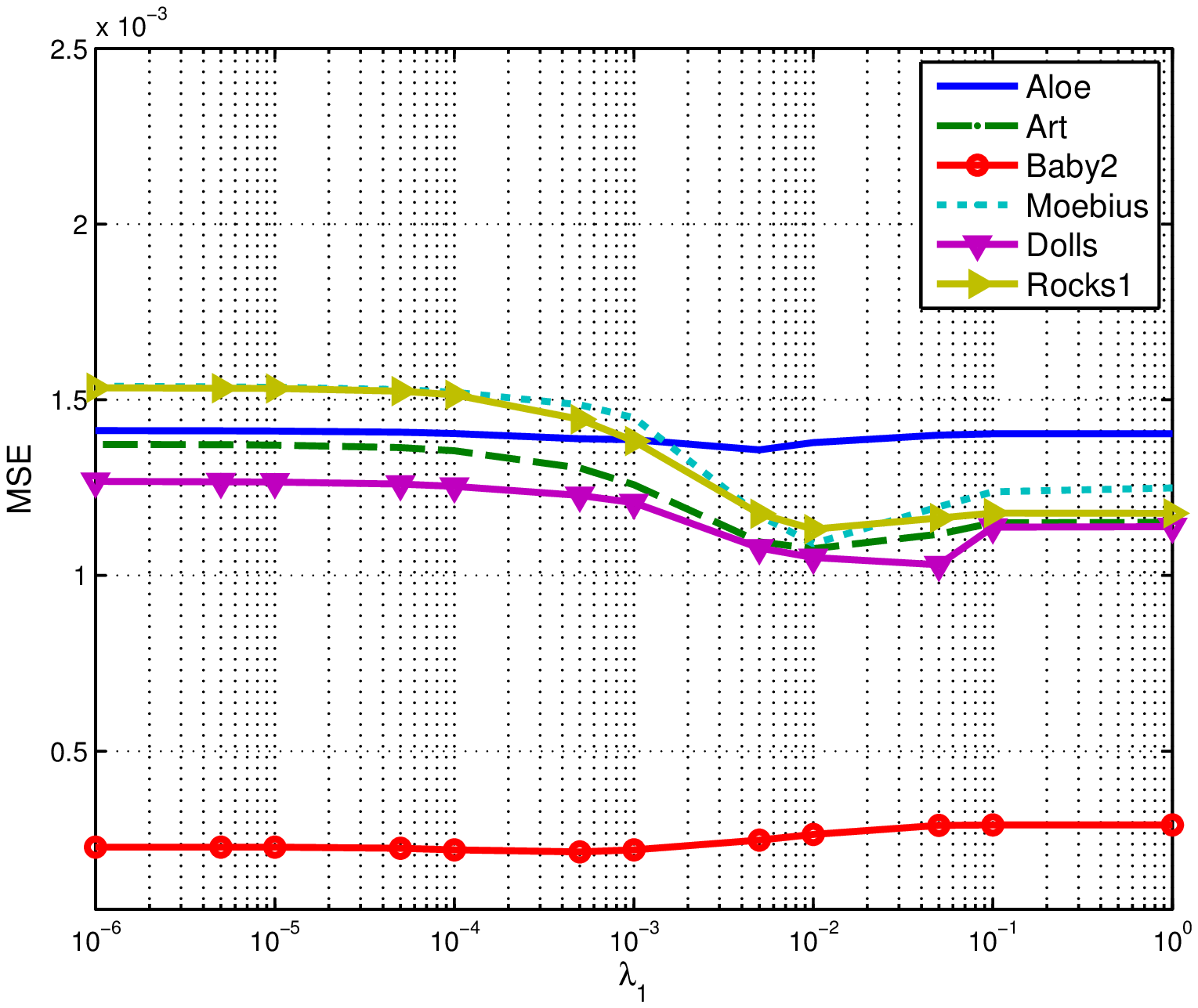} \hspace{-10pt}
&\includegraphics[width=0.31\textwidth]{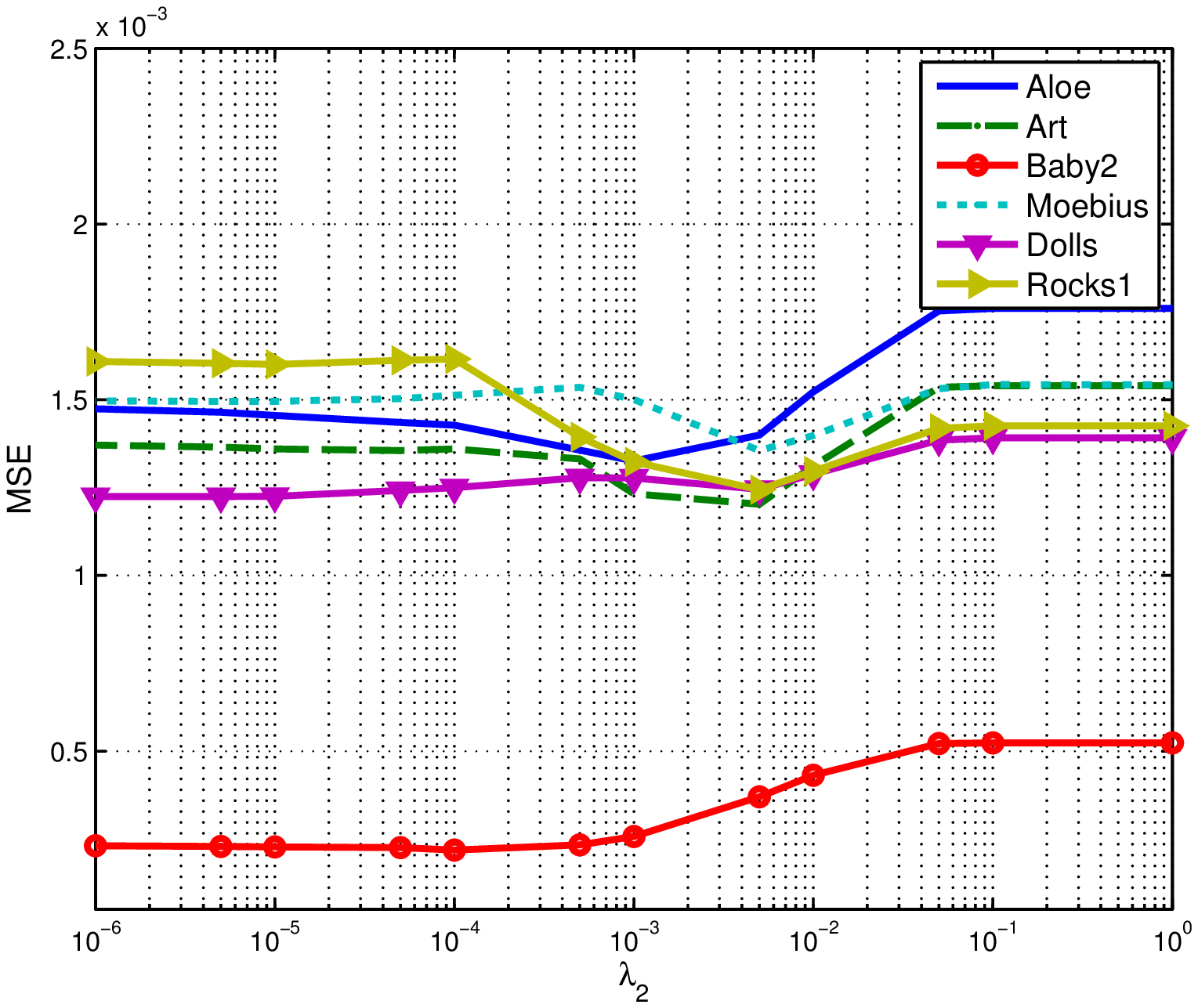} \hspace{-10pt}
&\includegraphics[width=0.31\textwidth]{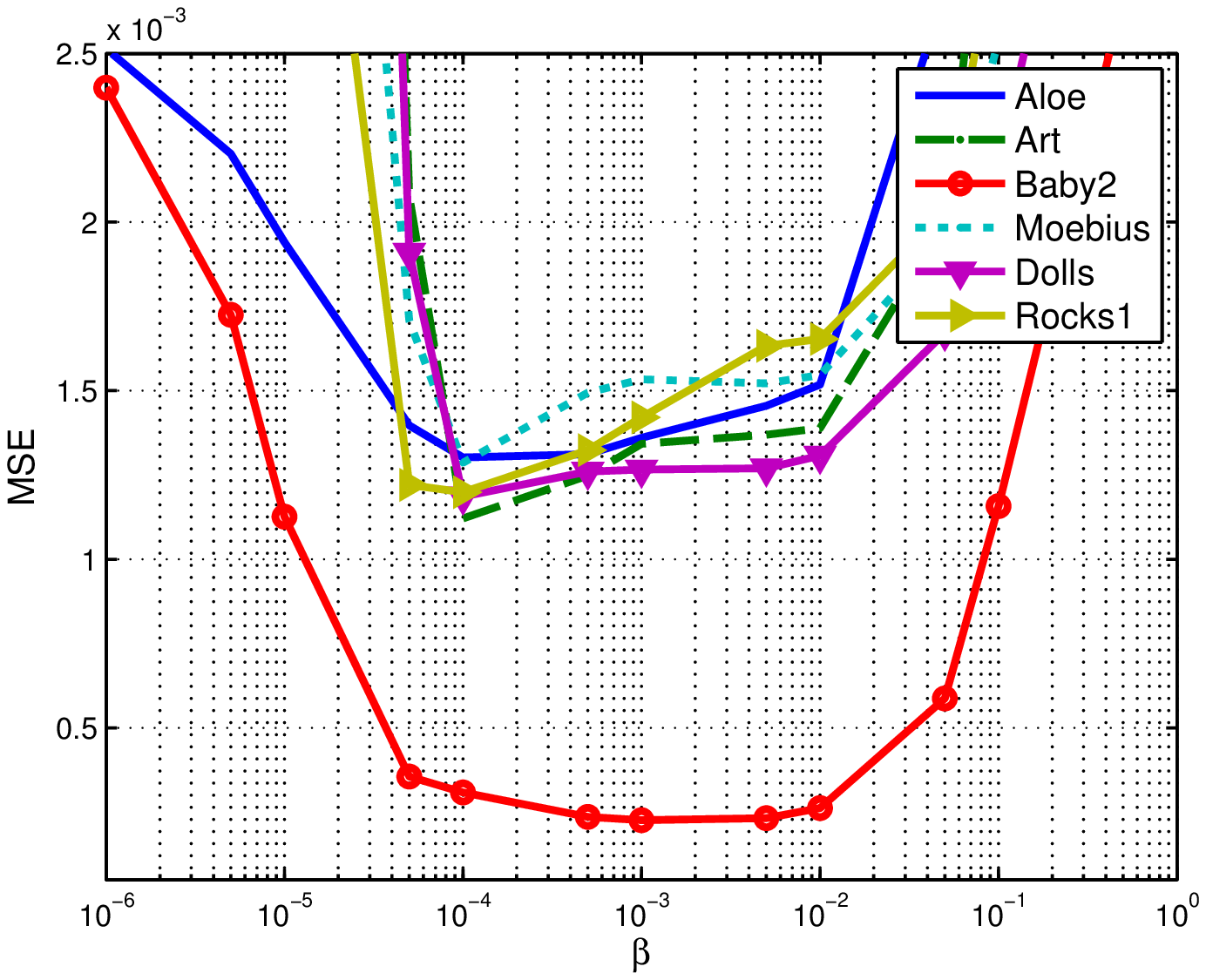} \\
$\lambda_1,\xi=0.1$ \hspace{-10pt}& $\lambda_2,\xi=0.1$ \hspace{-10pt}& $\beta,\xi=0.1$ \\
\end{tabular}
\caption{Comparison of reconstruction performance with varying regularization parameters and depth images. For each plot, we sweep a parameter from $10^{-6}$ to $10^{0}$ while fixing others to be our typical values. We set the sampling rate to be 20$\%$ (1st row) and 10$\%$ (2nd row). Typical values of regularization parameters are $\lambda_1=4\times 10^{-5}$, $\lambda_2=2\times 10^{-4}$ and $\beta=2\times 10^{-3}$. }
\label{fig:regularization_parameter_sweep}
\end{figure}

\fref{fig:regularization_parameter_sweep} shows the MSE curves for various images. For each plot, the MSE is computed by sweeping one parameter while fixing the other parameters. Observing the top row of \eref{fig:regularization_parameter_sweep}, we see that the optimal $\lambda_1$ across all images is approximately located in the range of $10^{-6}\leq \lambda_1 \leq 10^{-3}$. Therefore, we select $\lambda_1 = 4\times 10^{-5}$. Similarly, we can determine $\lambda_2 = 2\times 10^{-4}$ and $\beta=2\times 10^{-3}$.

We repeat the above analysis for $\xi=0.1$. The results are shown in the bottom row of \fref{fig:regularization_parameter_sweep}. The result indicates that while there are some difference in the MSE as compared to the top row, the optimal value does not change. Therefore, we keep the parameters using the above settings.

\newpage
\subsection{Internal Parameters ($\mu,\rho_1, \rho_2, \gamma$)}
\begin{figure}[ht!]
\centering
\begin{tabular}{cccc}
Aloe & & & \\
\includegraphics[width=0.24\textwidth]{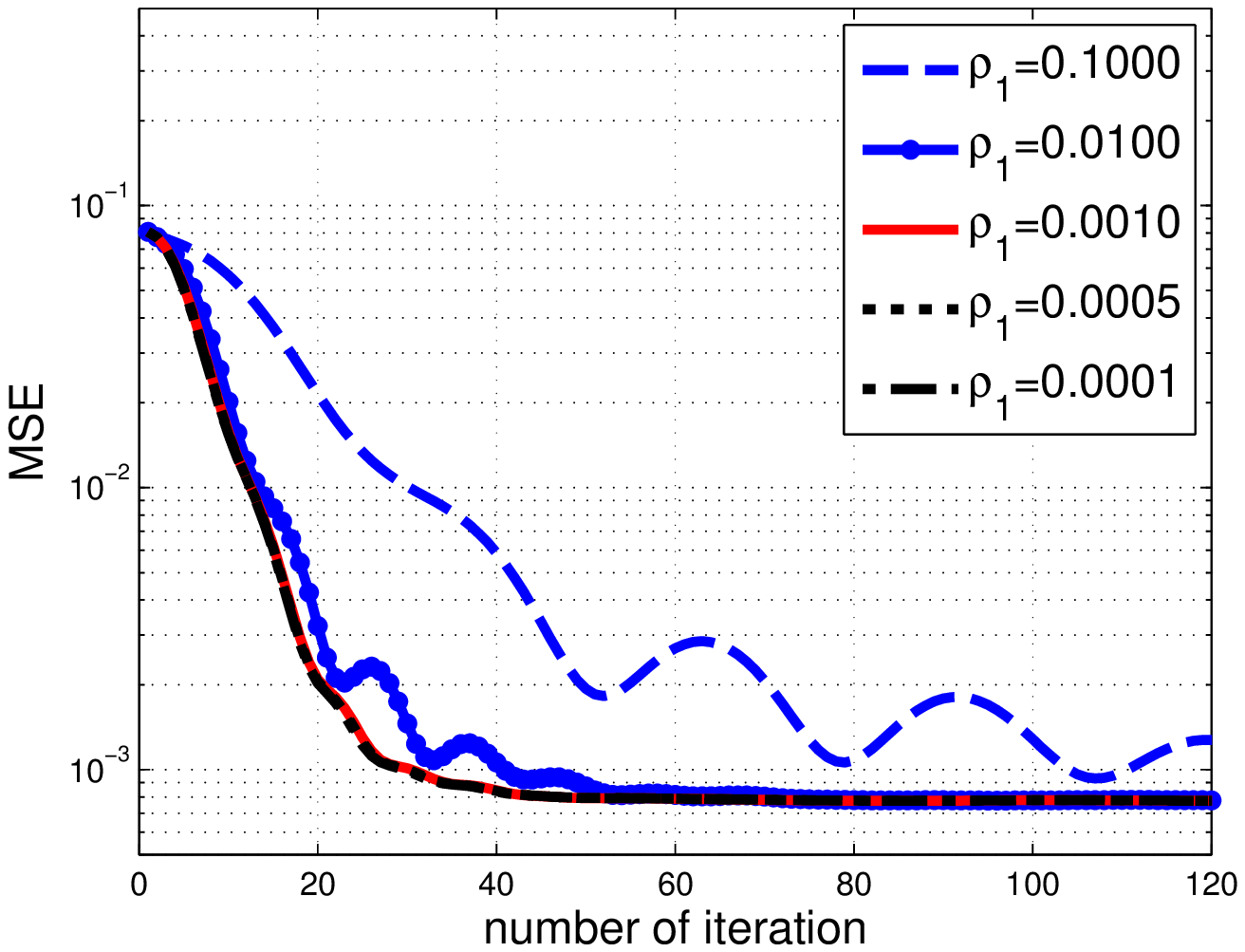} \hspace{-15pt}
&\includegraphics[width=0.24\textwidth]{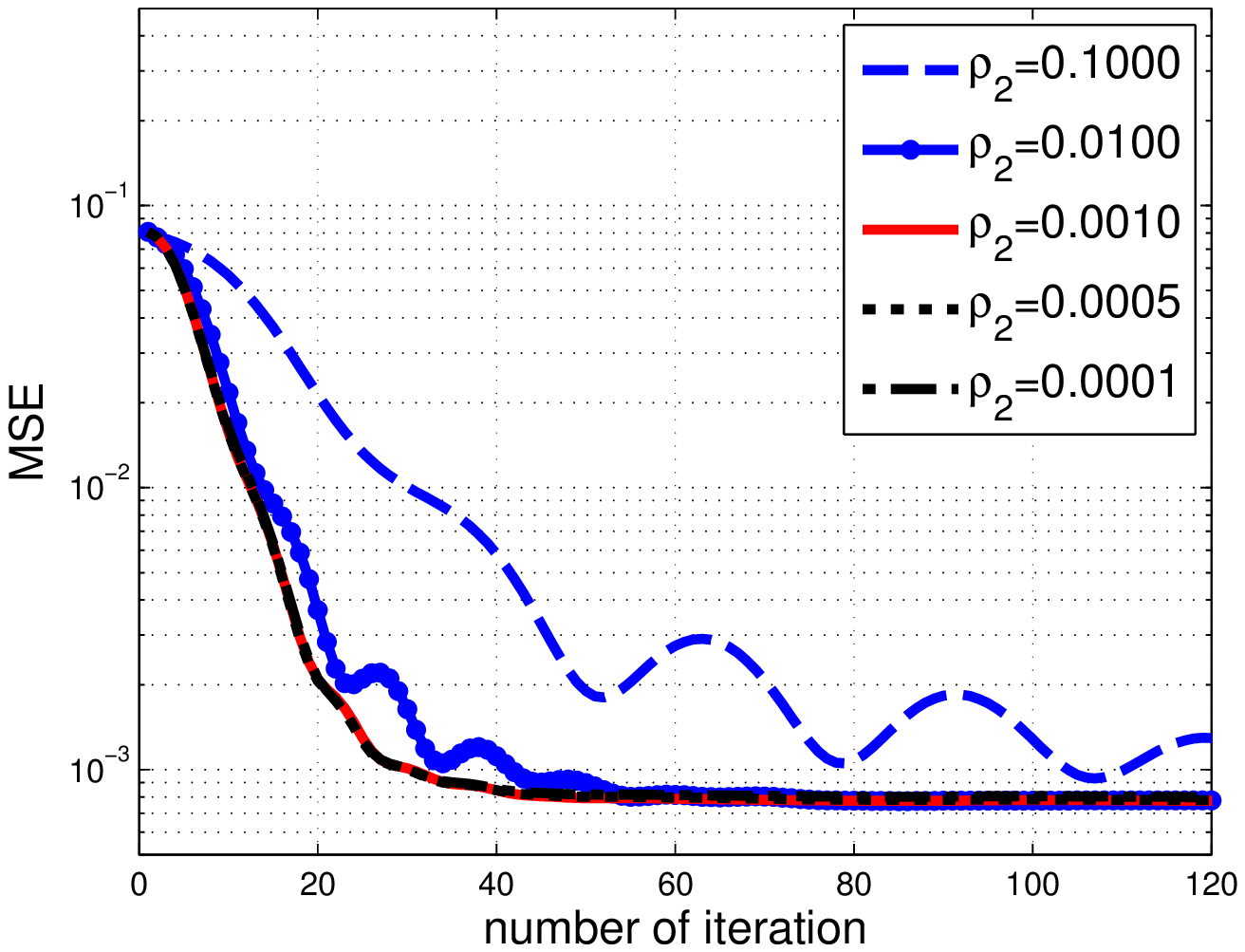} \hspace{-15pt}
&\includegraphics[width=0.24\textwidth]{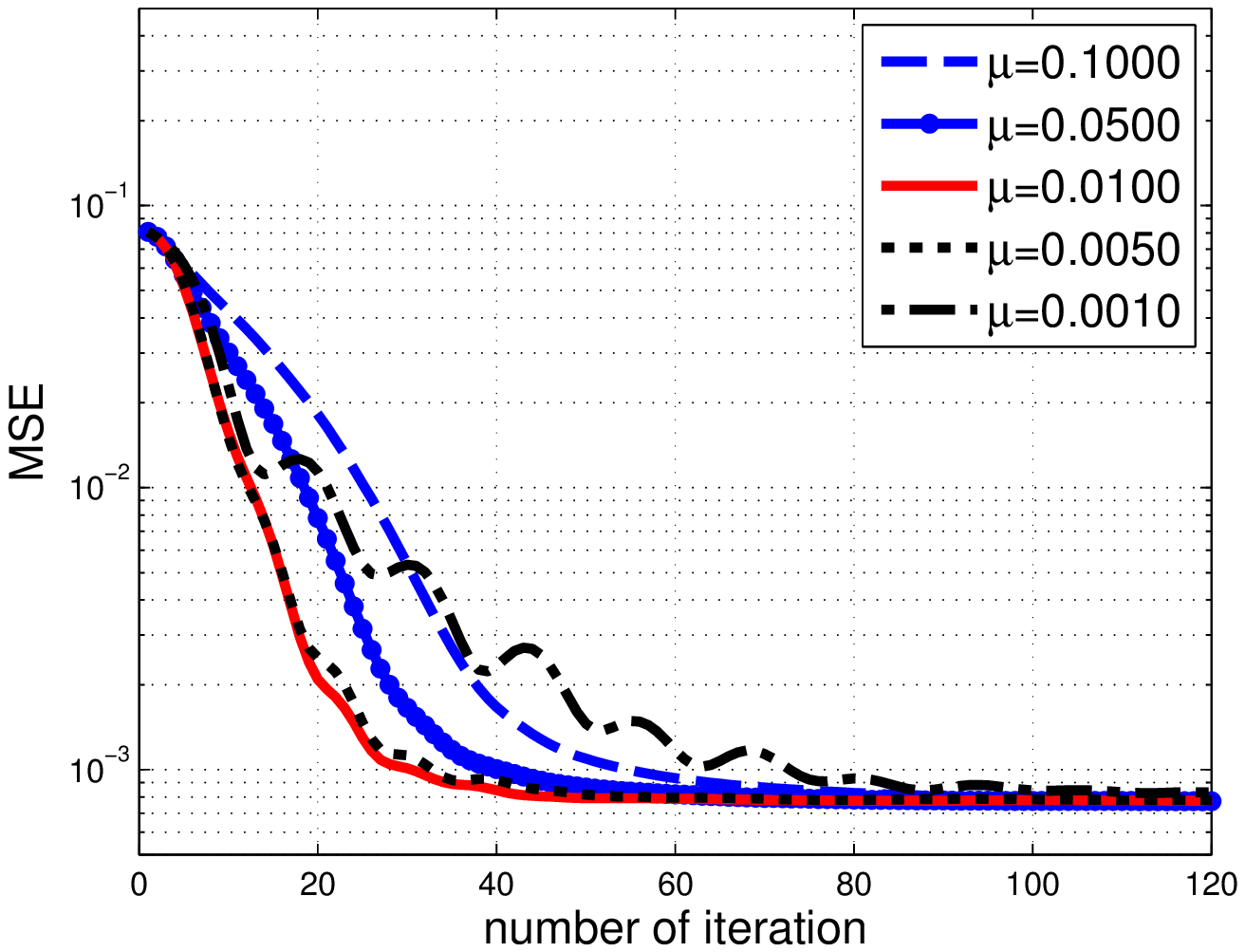} \hspace{-15pt}
&\includegraphics[width=0.24\textwidth]{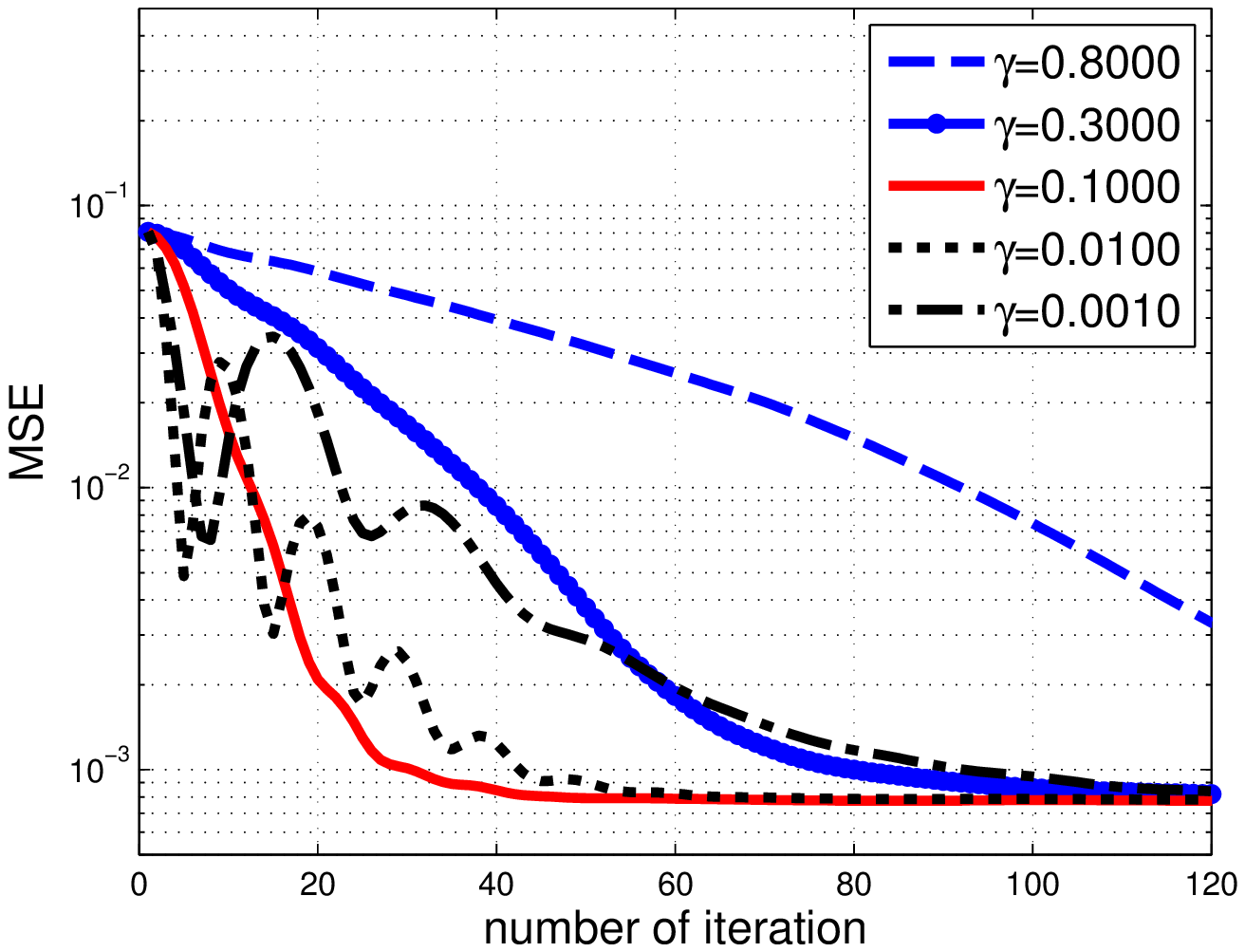}\\
$\rho_1$ \hspace{-15pt}& $\rho_2$\hspace{-15pt} & $\mu$\hspace{-15pt} & $\gamma$\\
Art & & & \\
\includegraphics[width=0.24\textwidth]{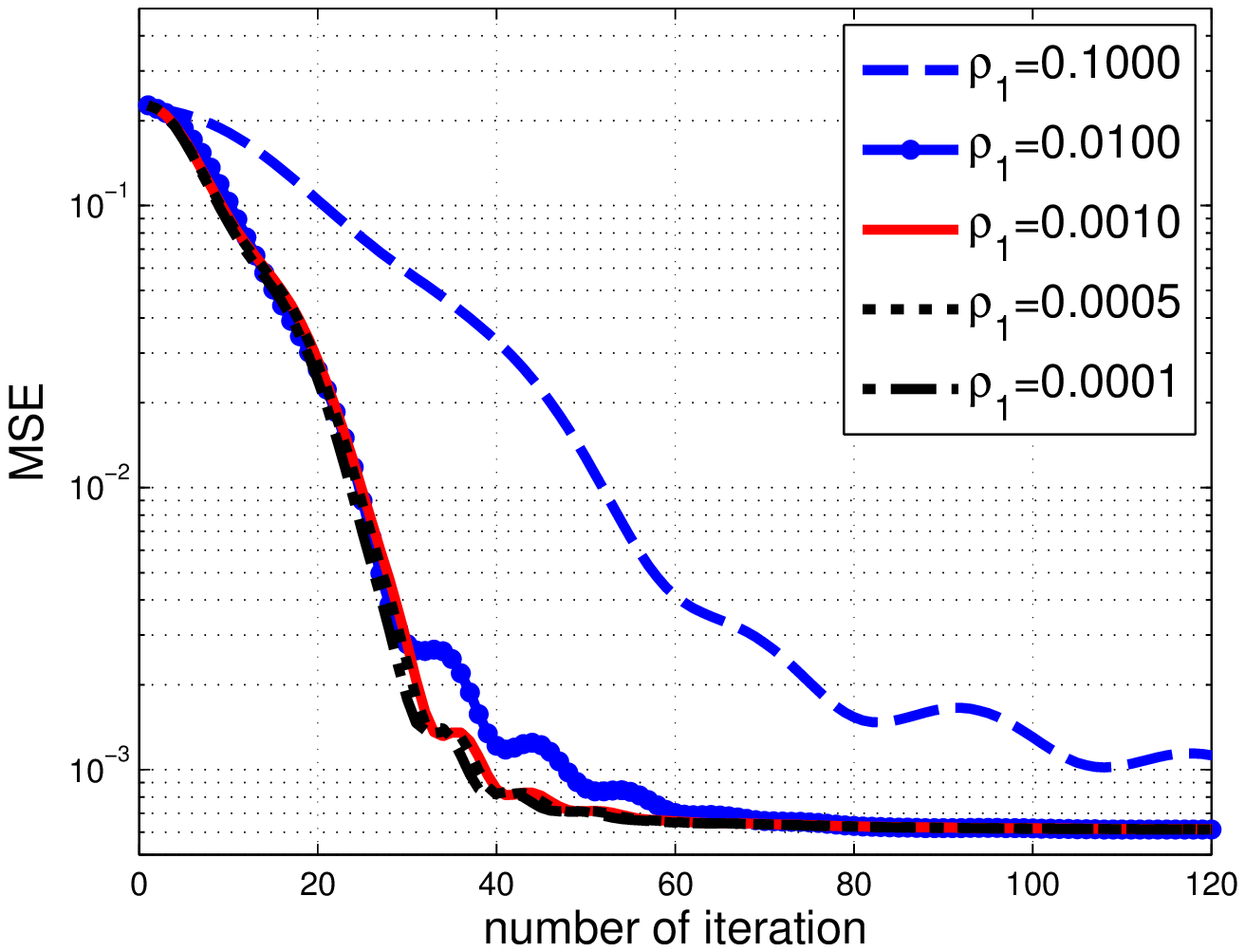} \hspace{-15pt}
&\includegraphics[width=0.24\textwidth]{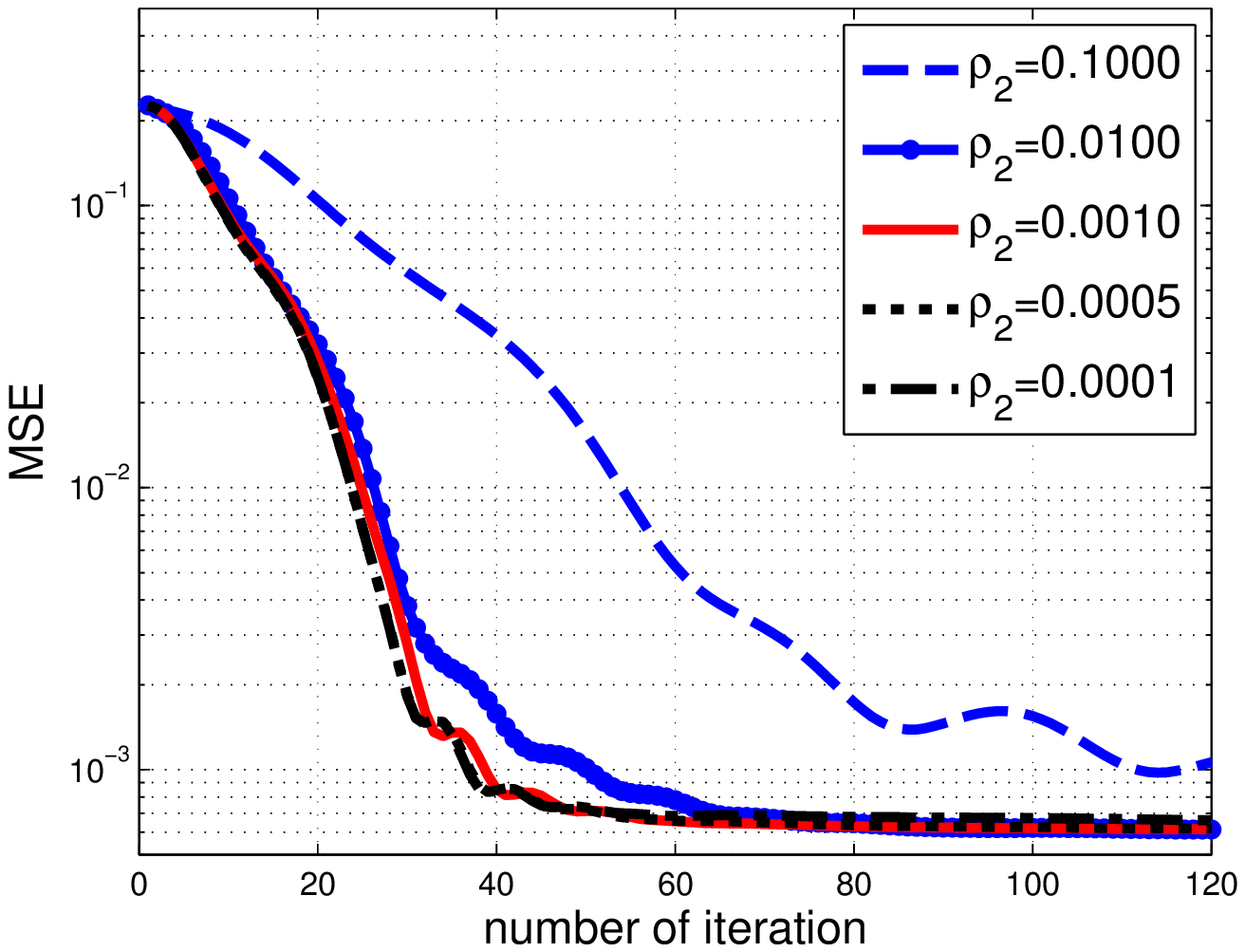} \hspace{-15pt}
&\includegraphics[width=0.24\textwidth]{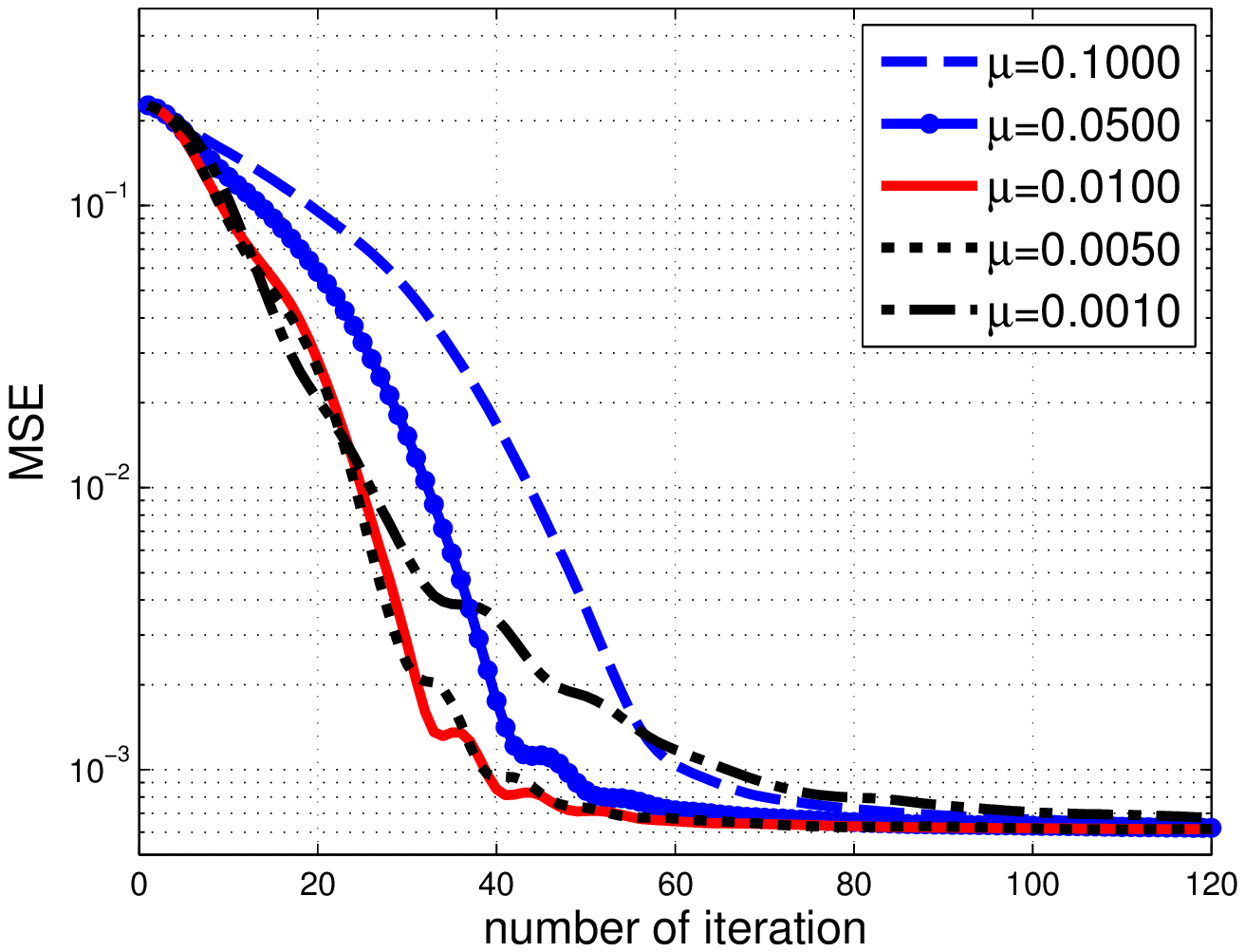} \hspace{-15pt}
&\includegraphics[width=0.24\textwidth]{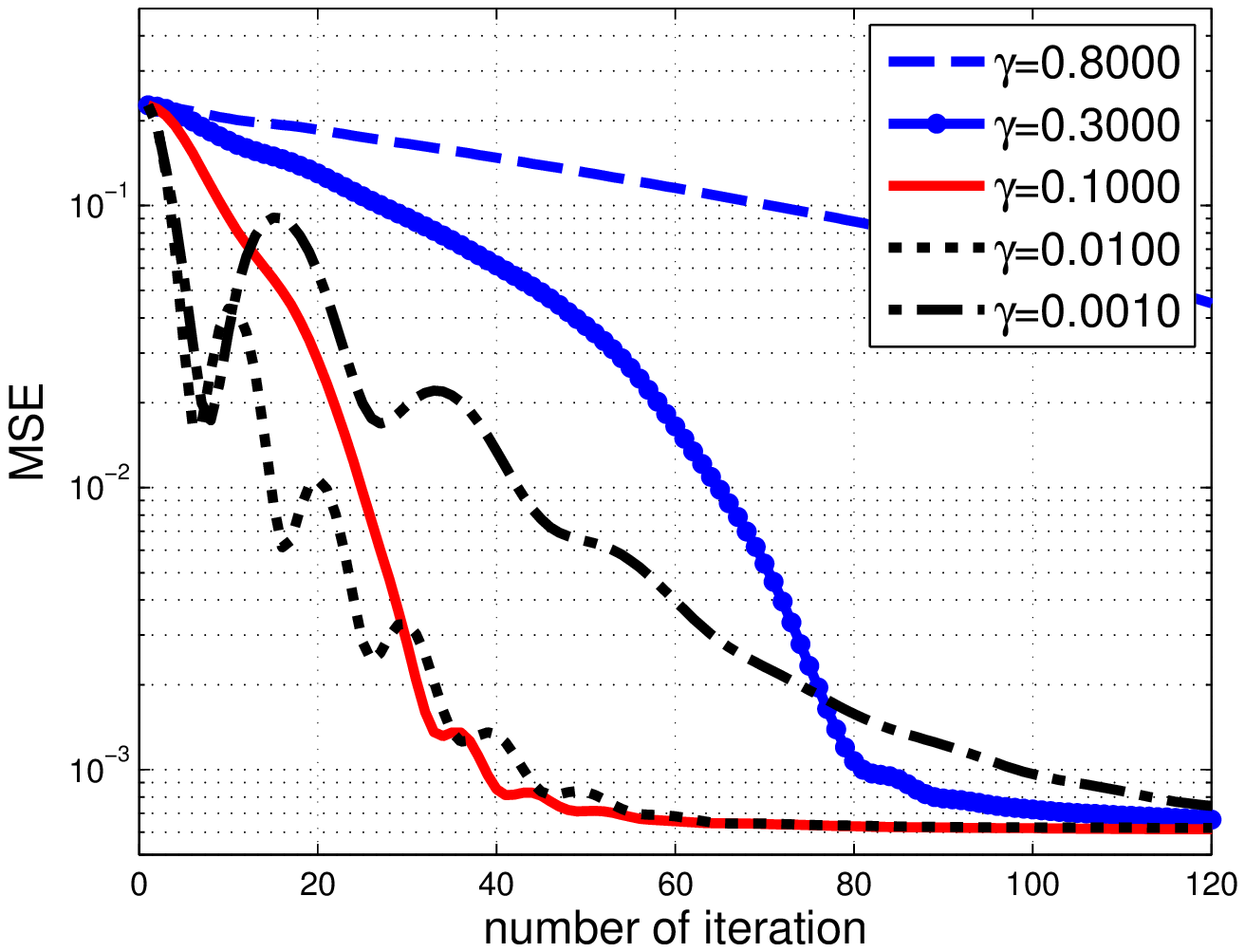}\\
$\rho_1$ \hspace{-15pt}& $\rho_2$\hspace{-15pt} & $\mu$\hspace{-15pt} & $\gamma$\\
Baby2 & & & \\
\includegraphics[width=0.235\textwidth]{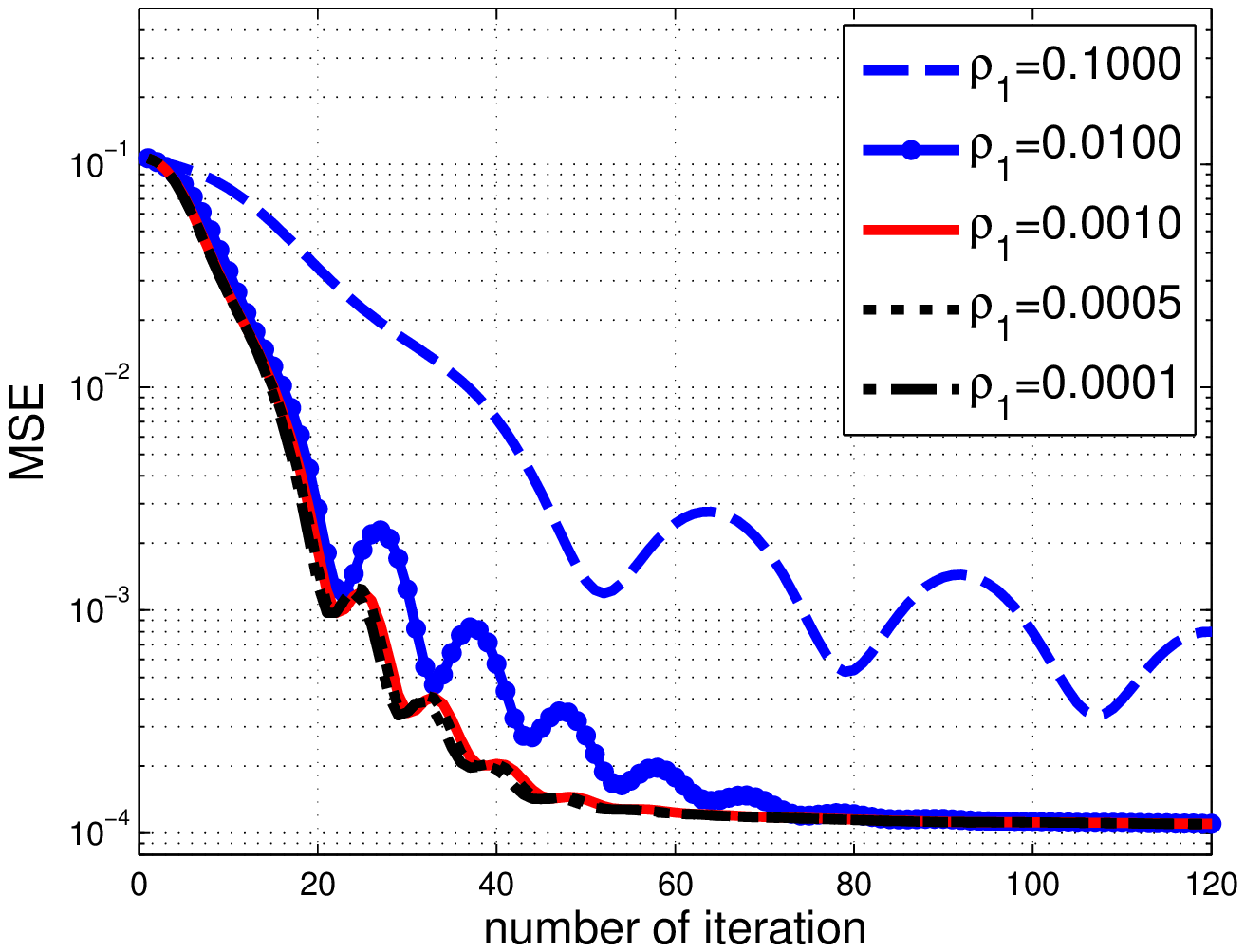}\hspace{-15pt}
&\includegraphics[width=0.235\textwidth]{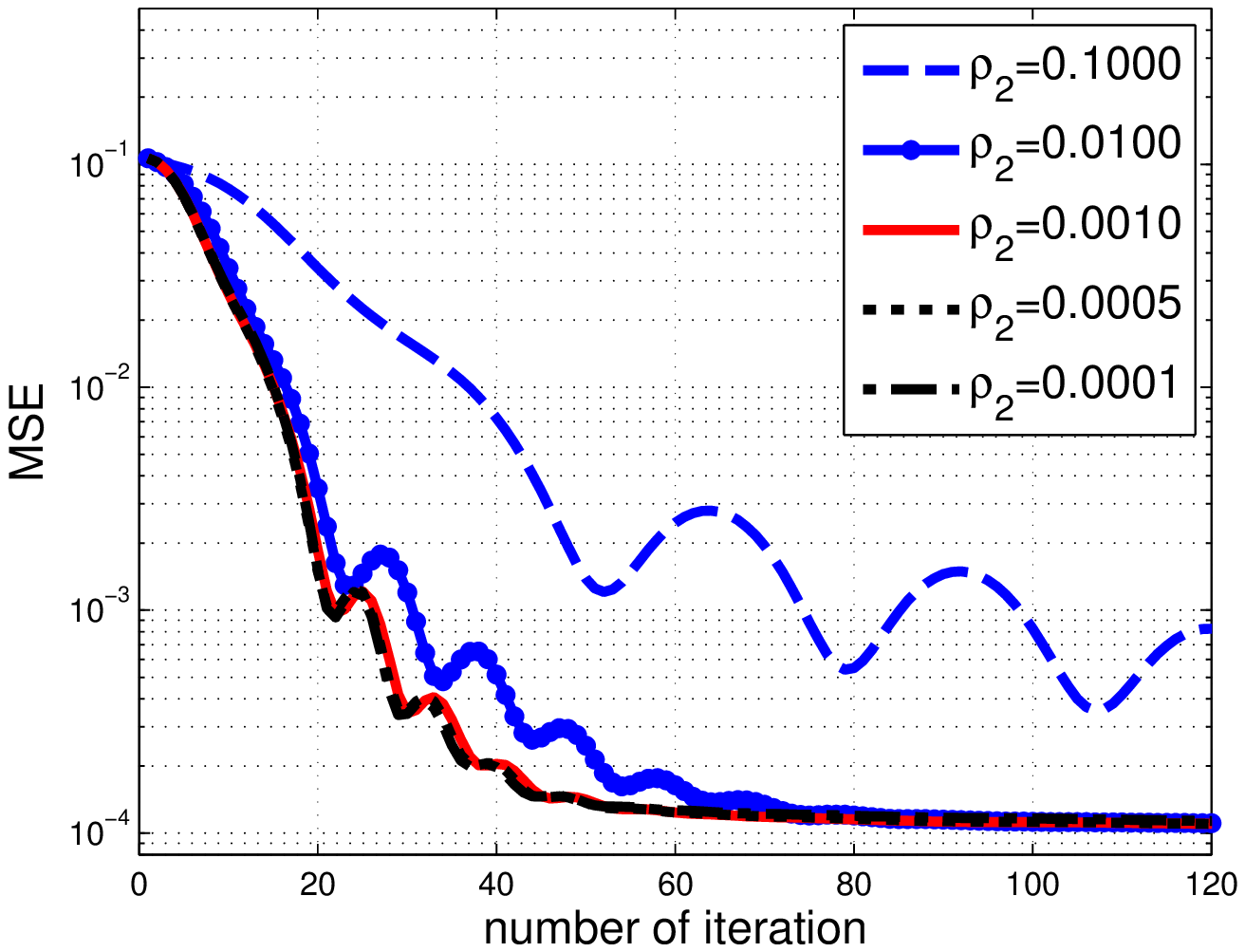}\hspace{-15pt}
&\includegraphics[width=0.235\textwidth]{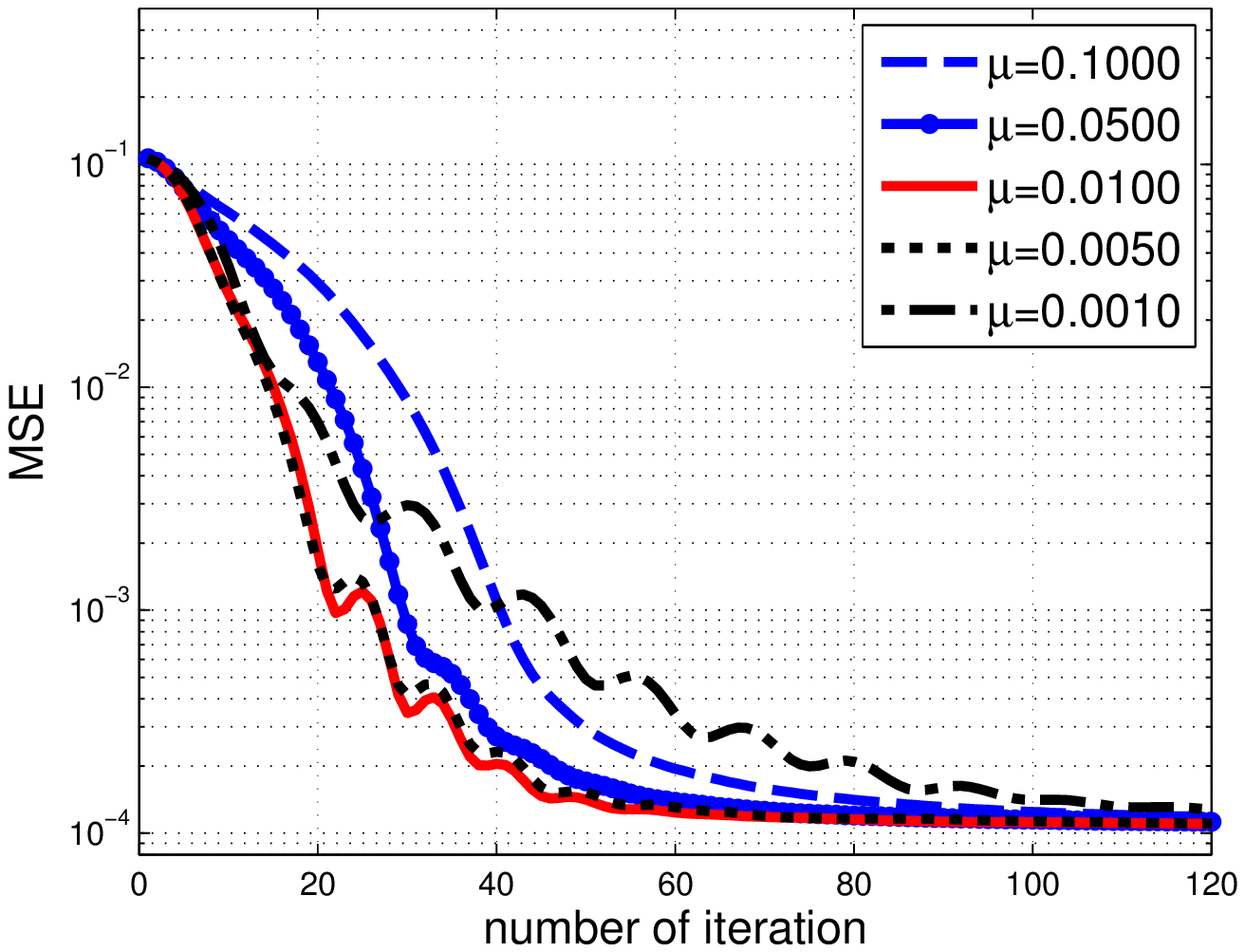}\hspace{-15pt}
&\includegraphics[width=0.235\textwidth]{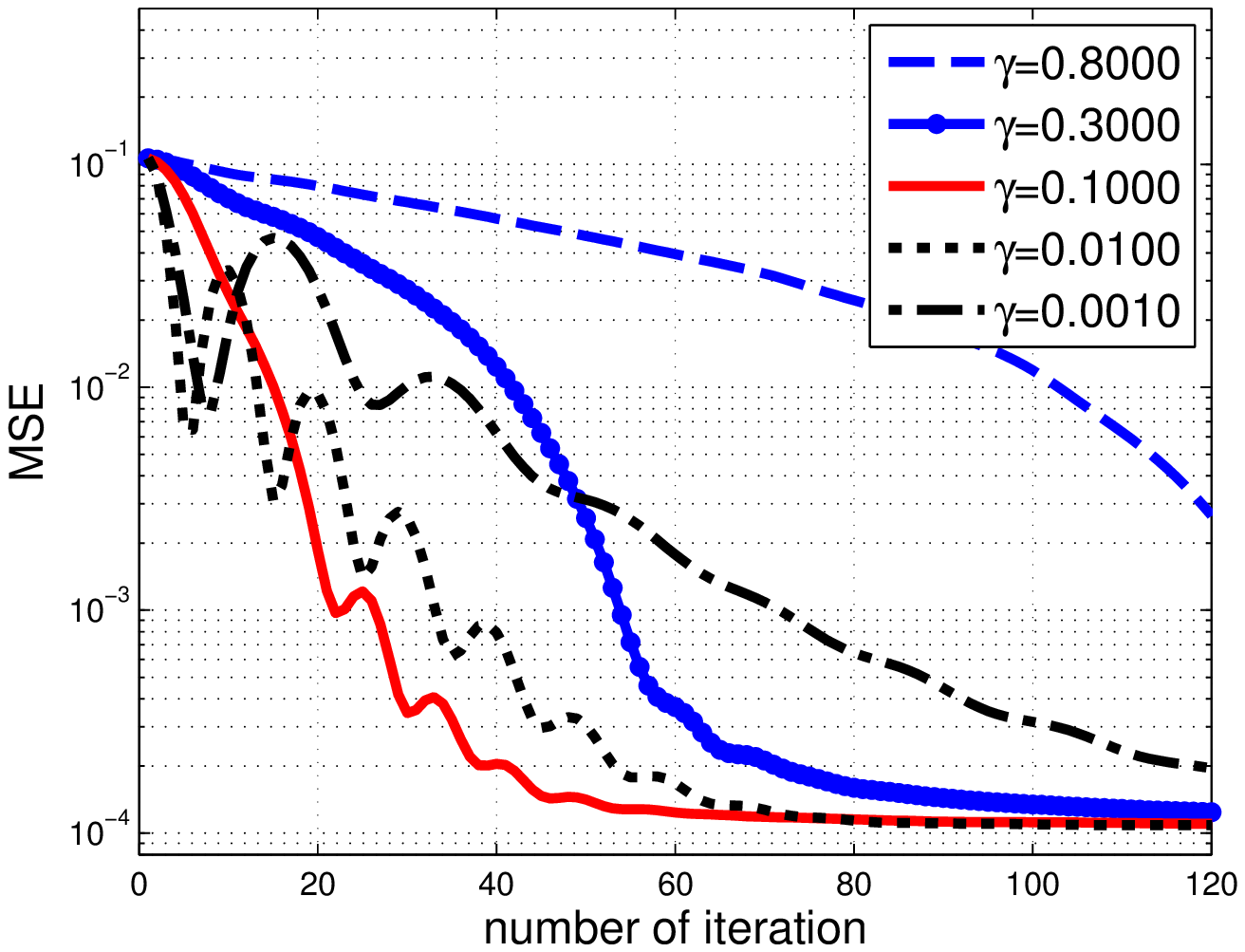}\\
$\rho_1$ \hspace{-15pt}& $\rho_2$\hspace{-15pt} & $\mu$\hspace{-15pt} & $\gamma$\\
Moebius & & & \\
\includegraphics[width=0.235\textwidth]{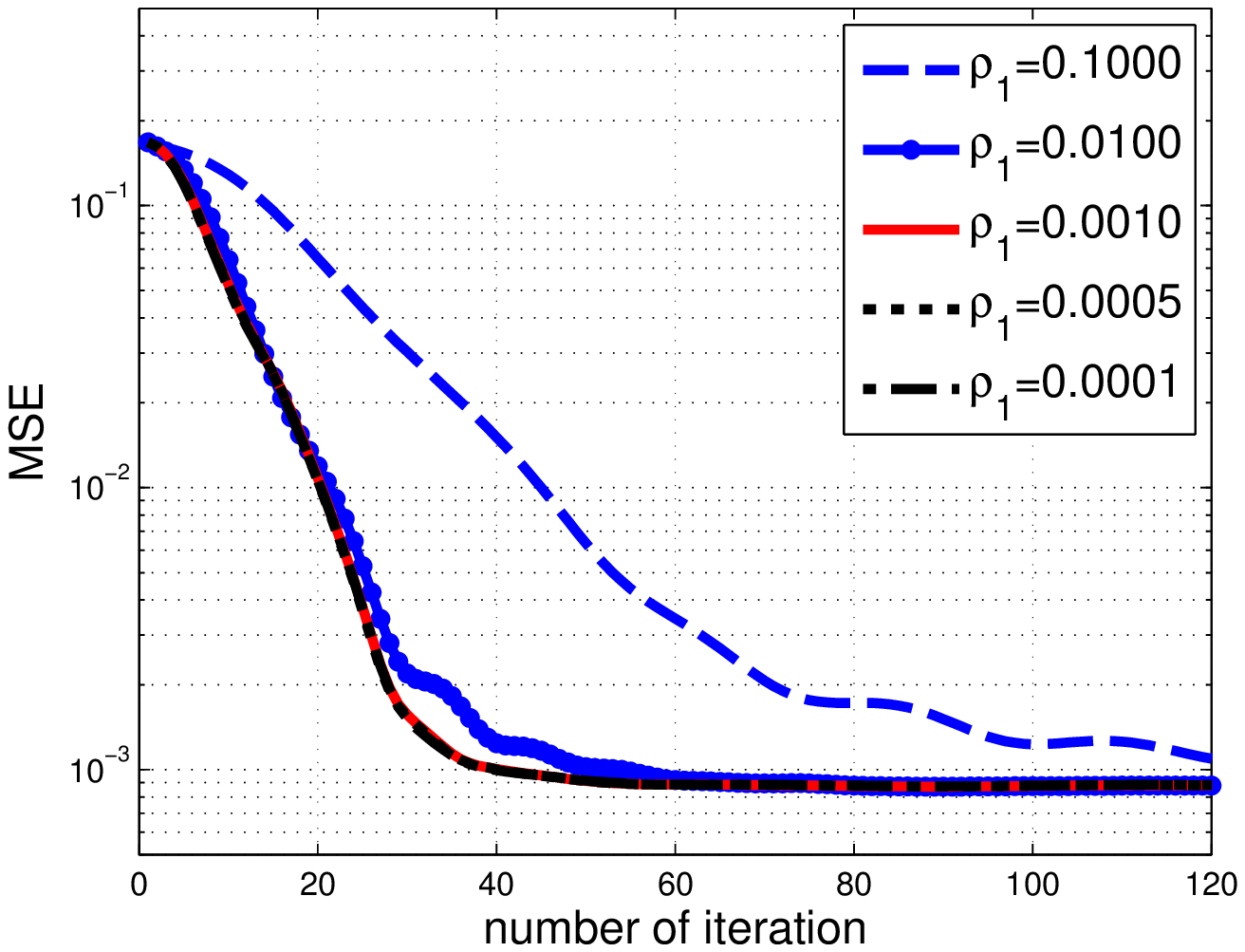}\hspace{-15pt}
&\includegraphics[width=0.235\textwidth]{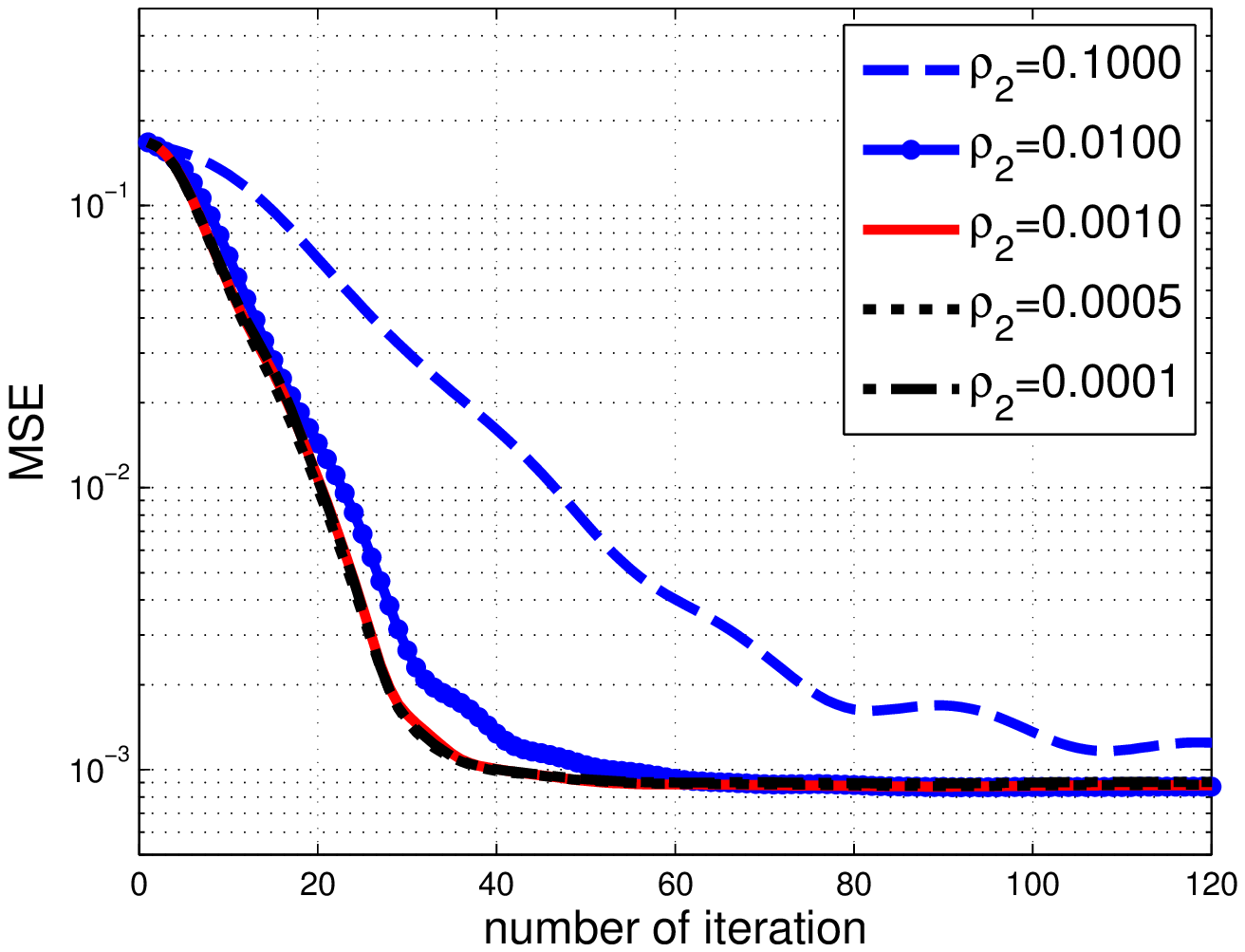}\hspace{-15pt}
&\includegraphics[width=0.235\textwidth]{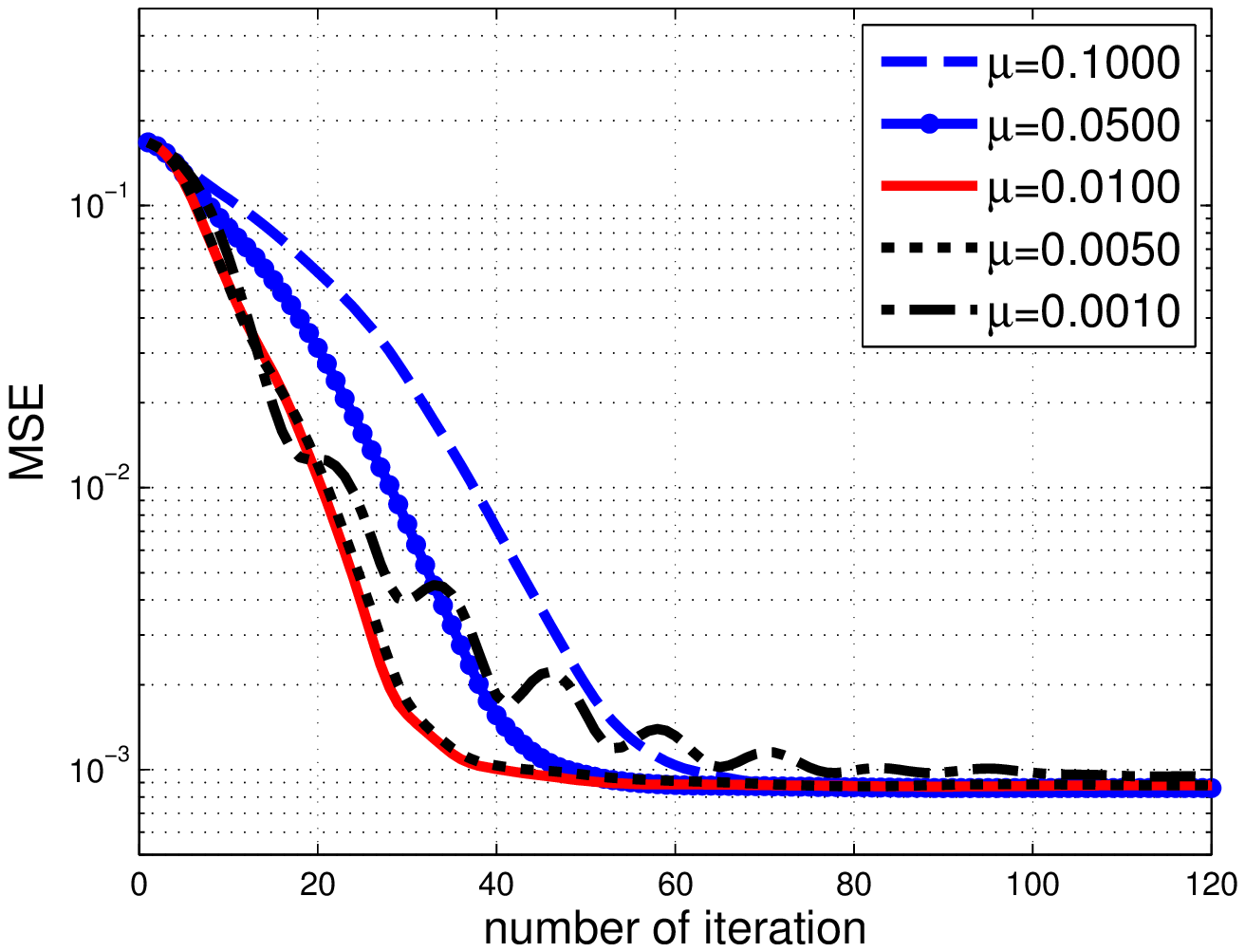}\hspace{-15pt}
&\includegraphics[width=0.235\textwidth]{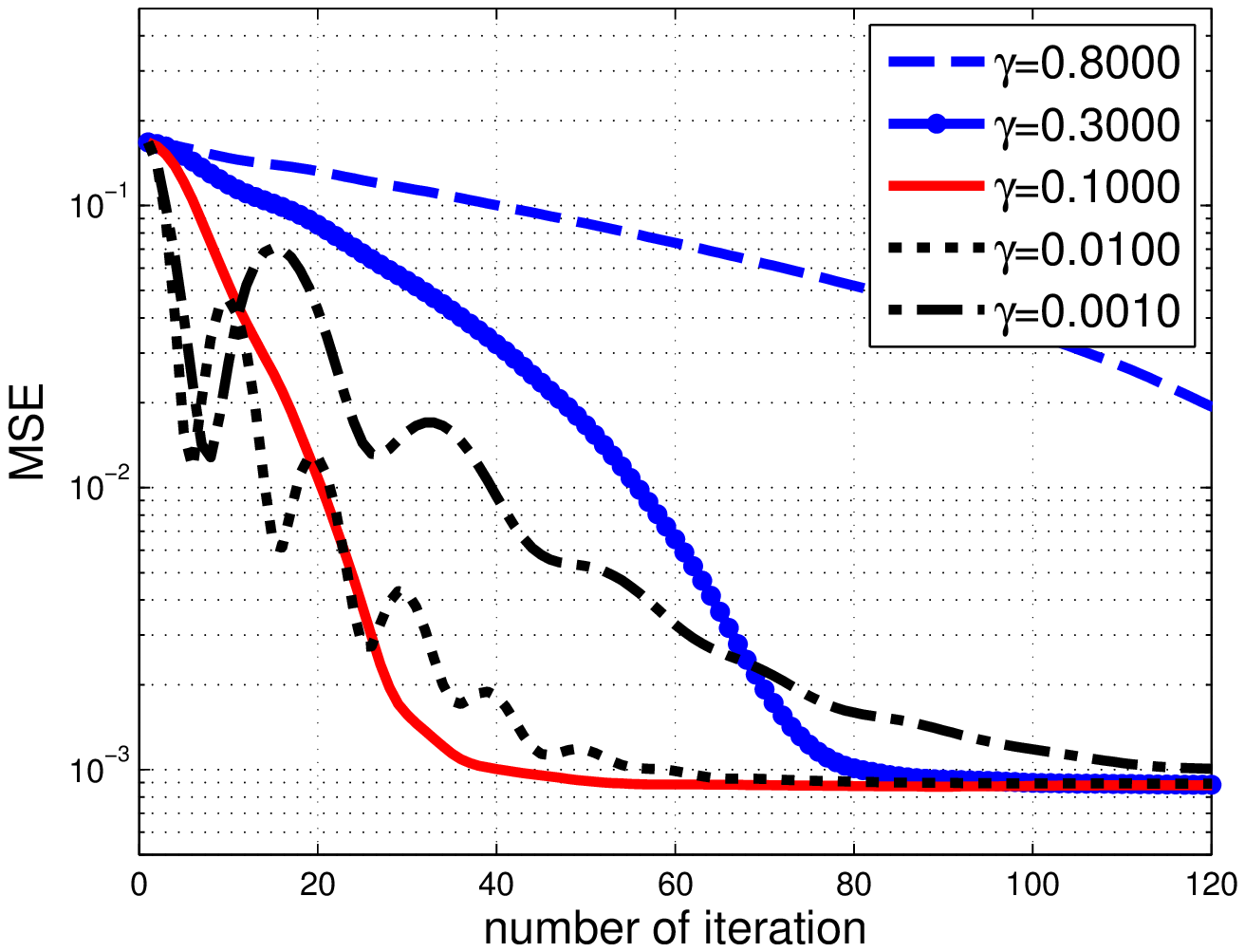}\\
$\rho_1$ \hspace{-15pt}& $\rho_2$\hspace{-15pt} & $\mu$\hspace{-15pt} & $\gamma$\\
Dolls & & & \\
\includegraphics[width=0.235\textwidth]{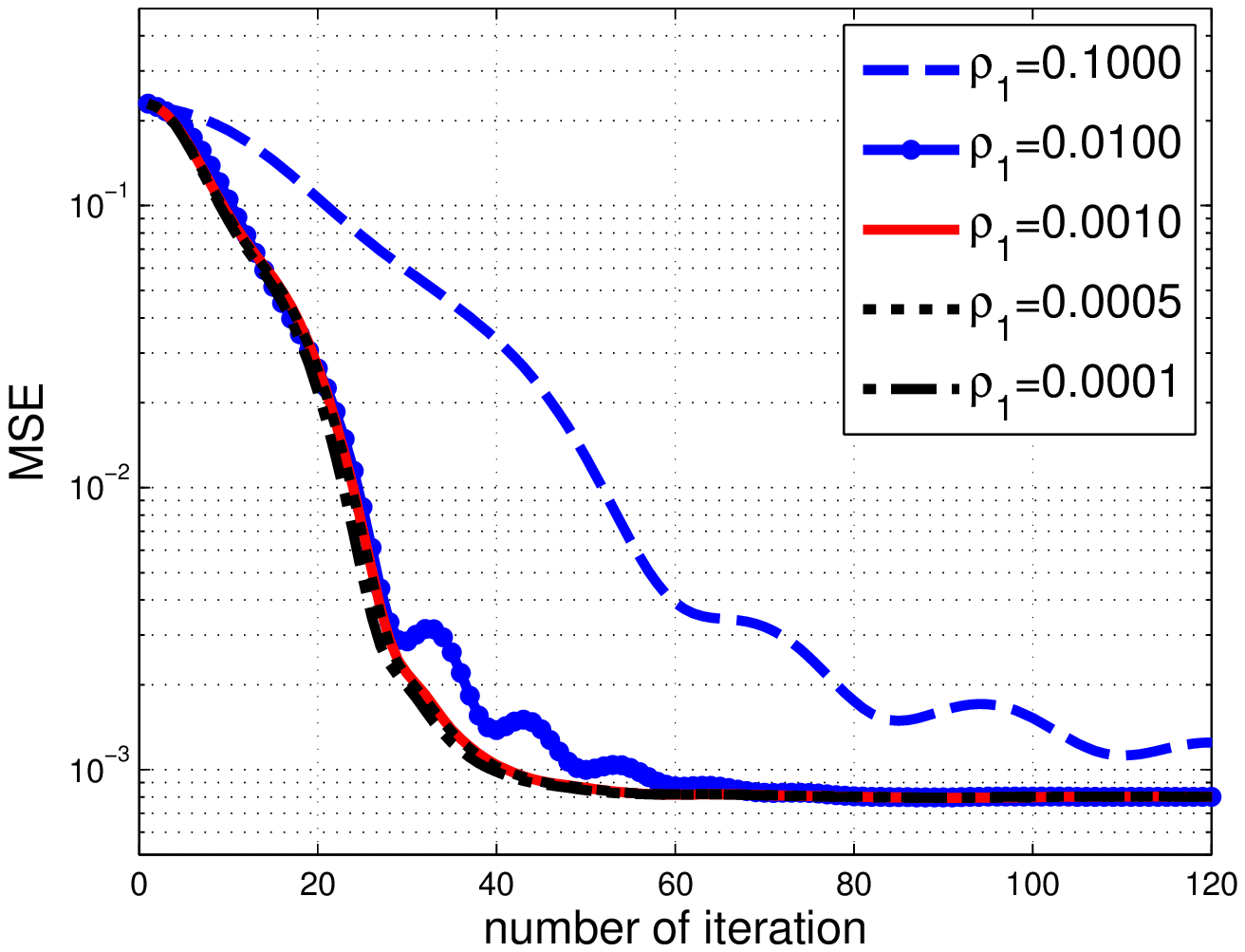}\hspace{-15pt}
&\includegraphics[width=0.235\textwidth]{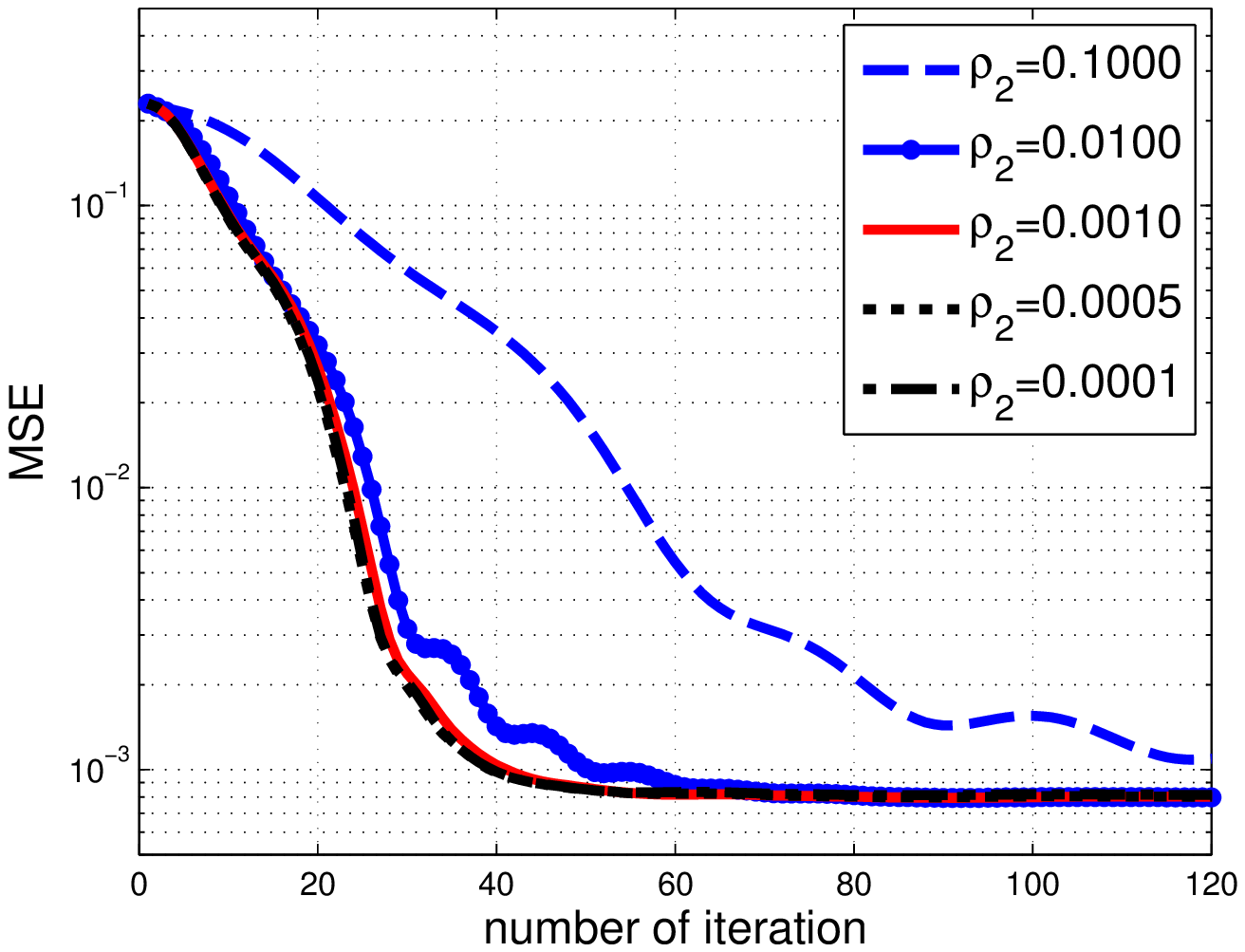}\hspace{-15pt}
&\includegraphics[width=0.235\textwidth]{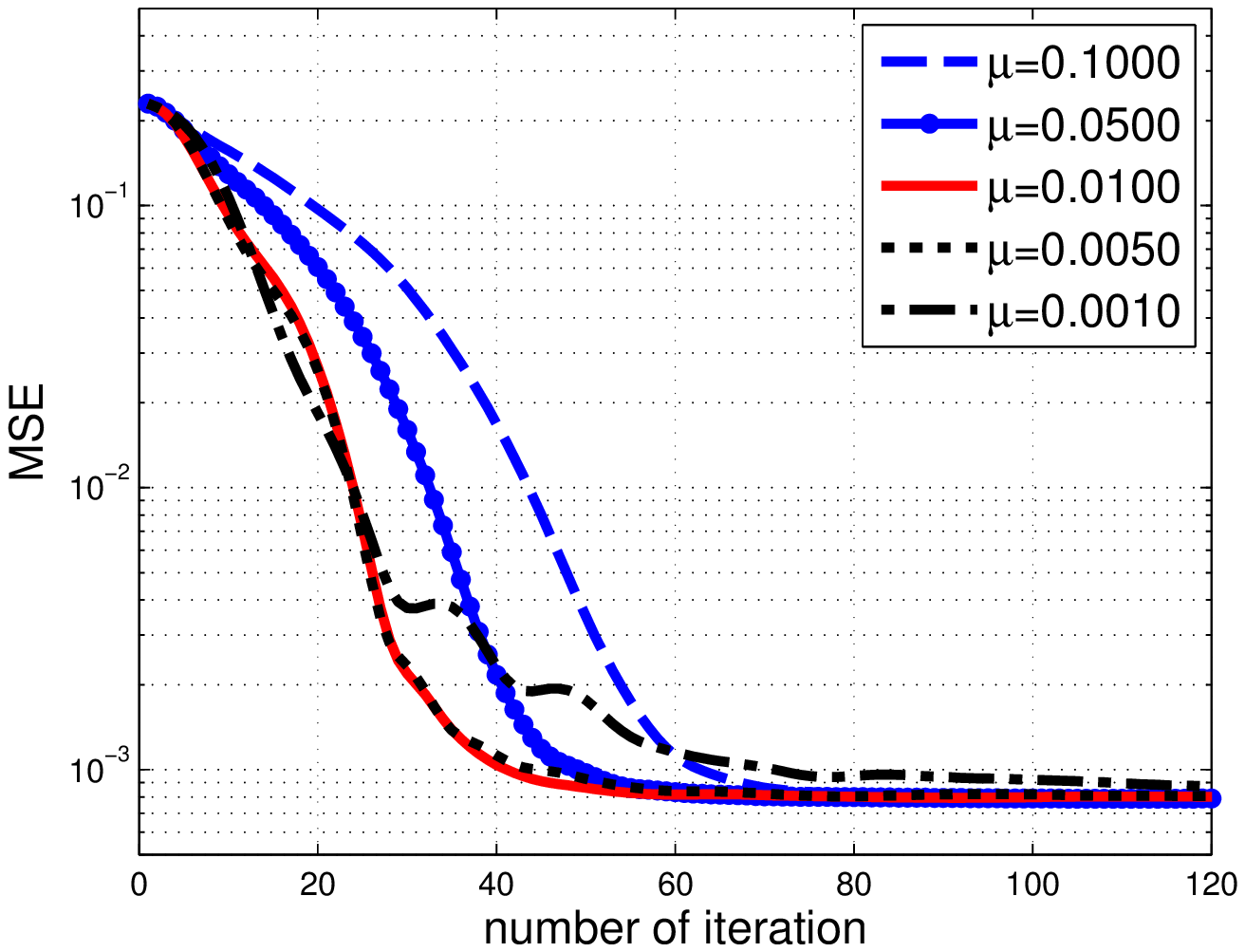}\hspace{-15pt}
&\includegraphics[width=0.235\textwidth]{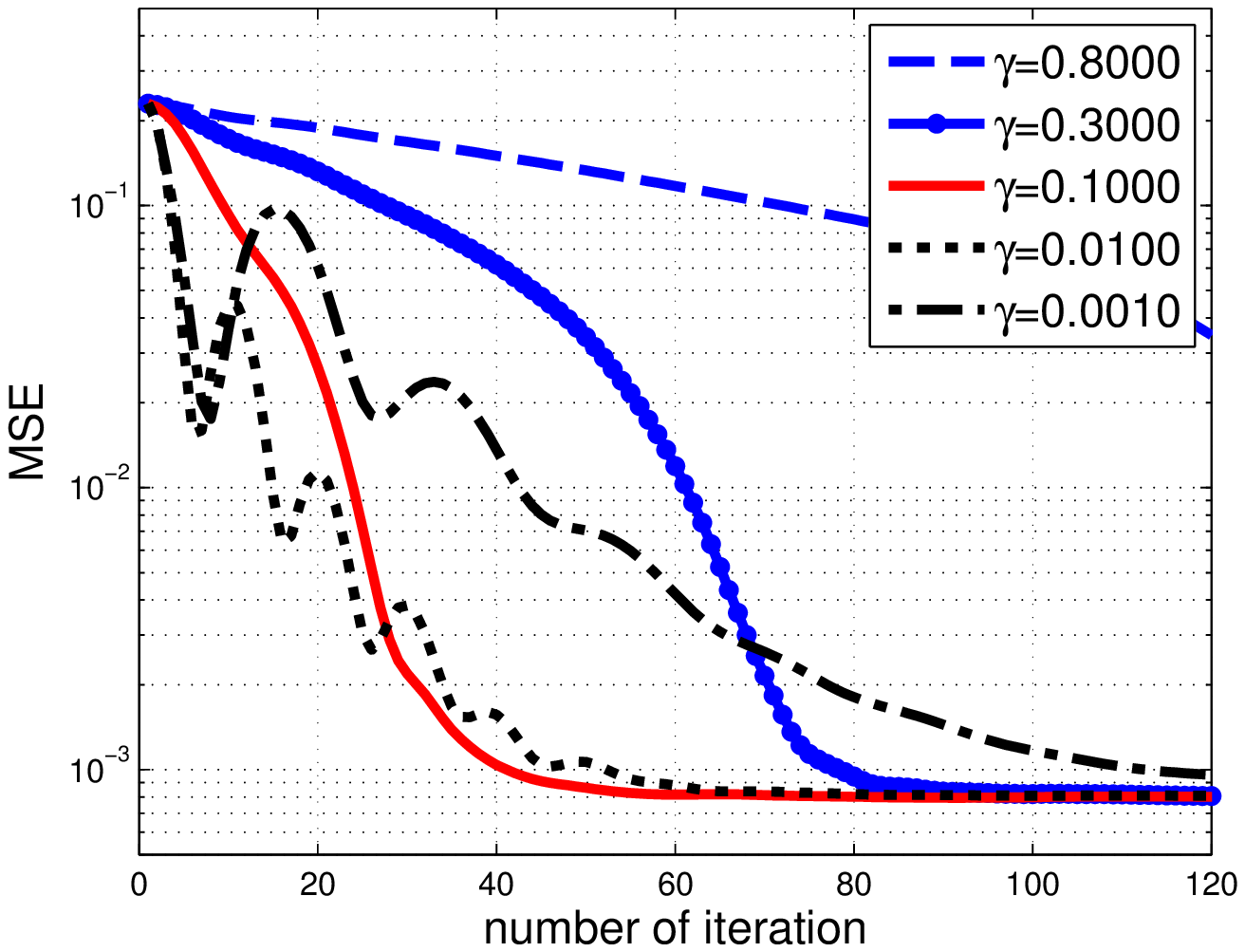}\\
$\rho_1$ \hspace{-15pt}& $\rho_2$\hspace{-15pt} & $\mu$\hspace{-15pt} & $\gamma$\\
\end{tabular}
\caption{MSE for $\xi = 0.2$.}
\label{fig:internal_parameter_sweep_20p_1}
\end{figure}

\begin{figure}[ht!]
\centering
\begin{tabular}{cccc}
Aloe & & & \\
\includegraphics[width=0.24\textwidth]{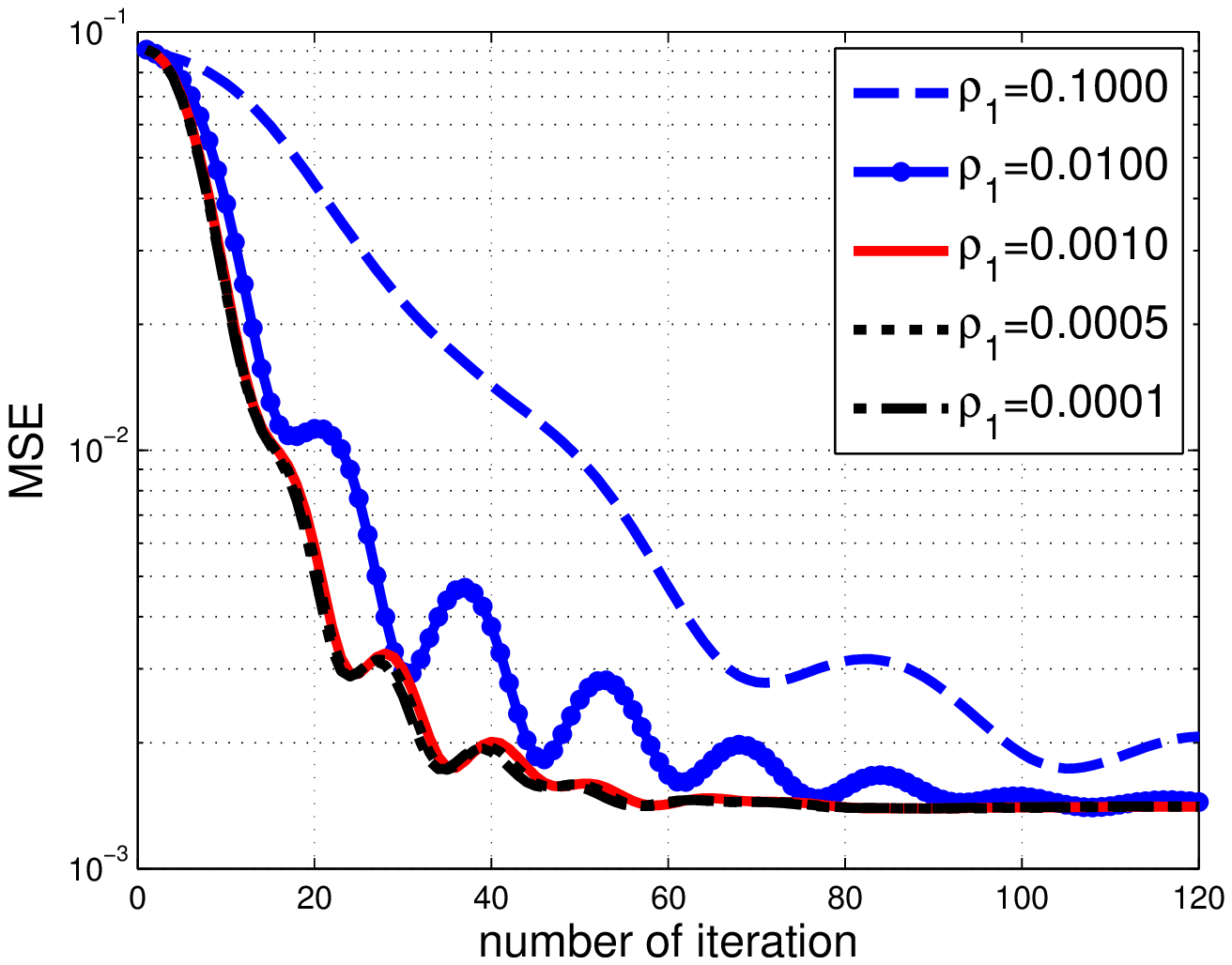} \hspace{-15pt}
&\includegraphics[width=0.24\textwidth]{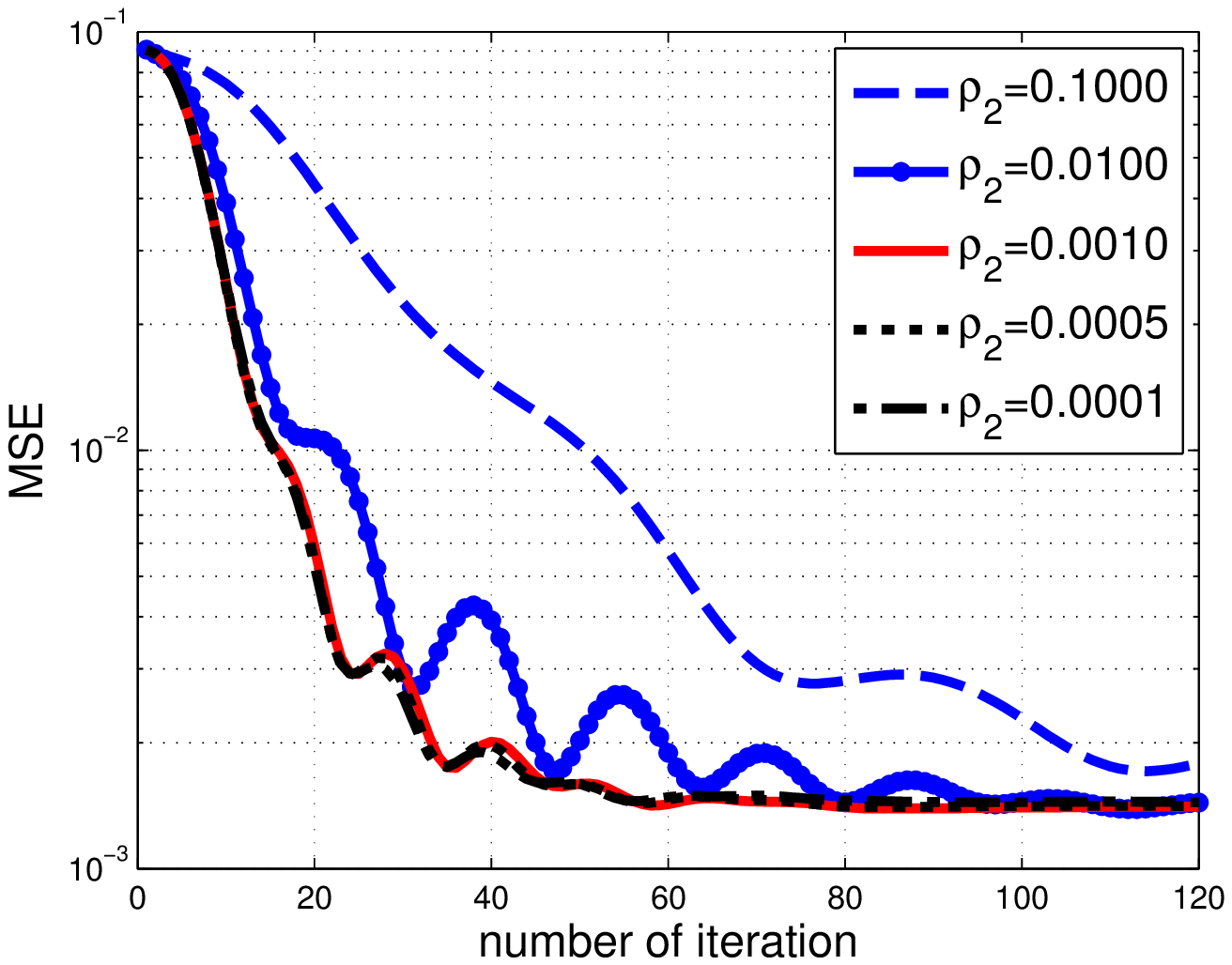} \hspace{-15pt}
&\includegraphics[width=0.24\textwidth]{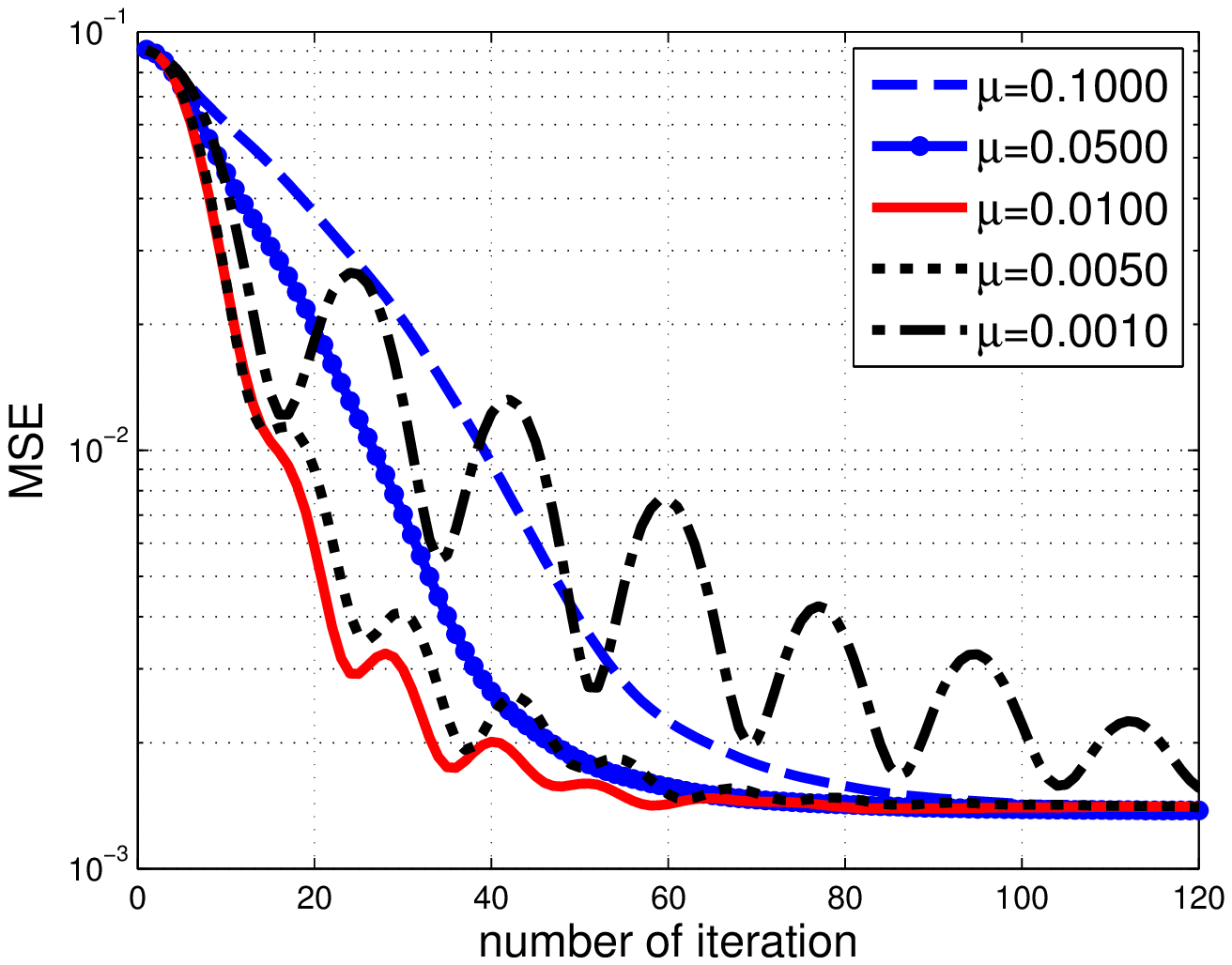} \hspace{-15pt}
&\includegraphics[width=0.24\textwidth]{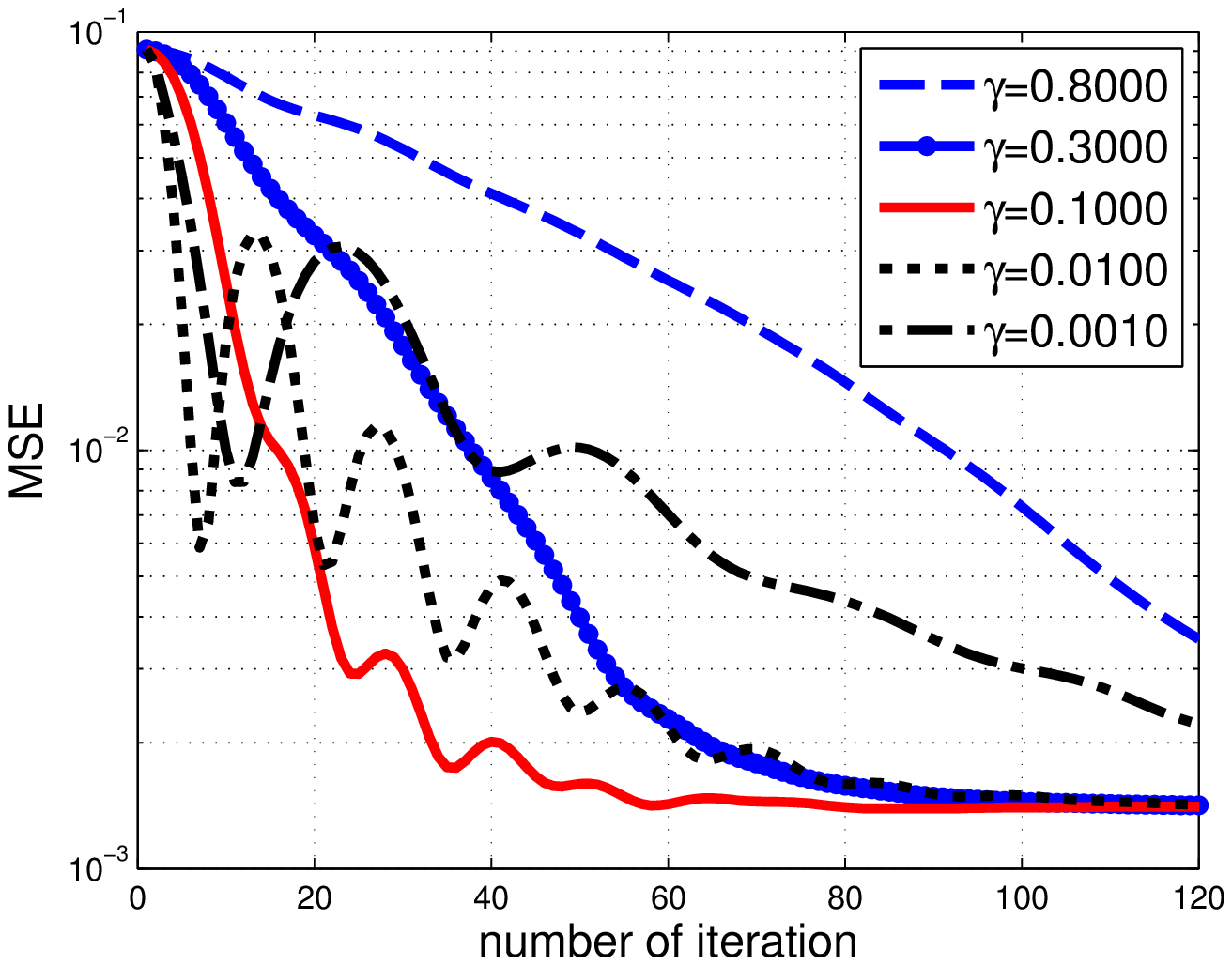}\\
$\rho_1$ \hspace{-15pt}& $\rho_2$\hspace{-15pt} & $\mu$\hspace{-15pt} & $\gamma$\\
Art & & & \\
\includegraphics[width=0.24\textwidth]{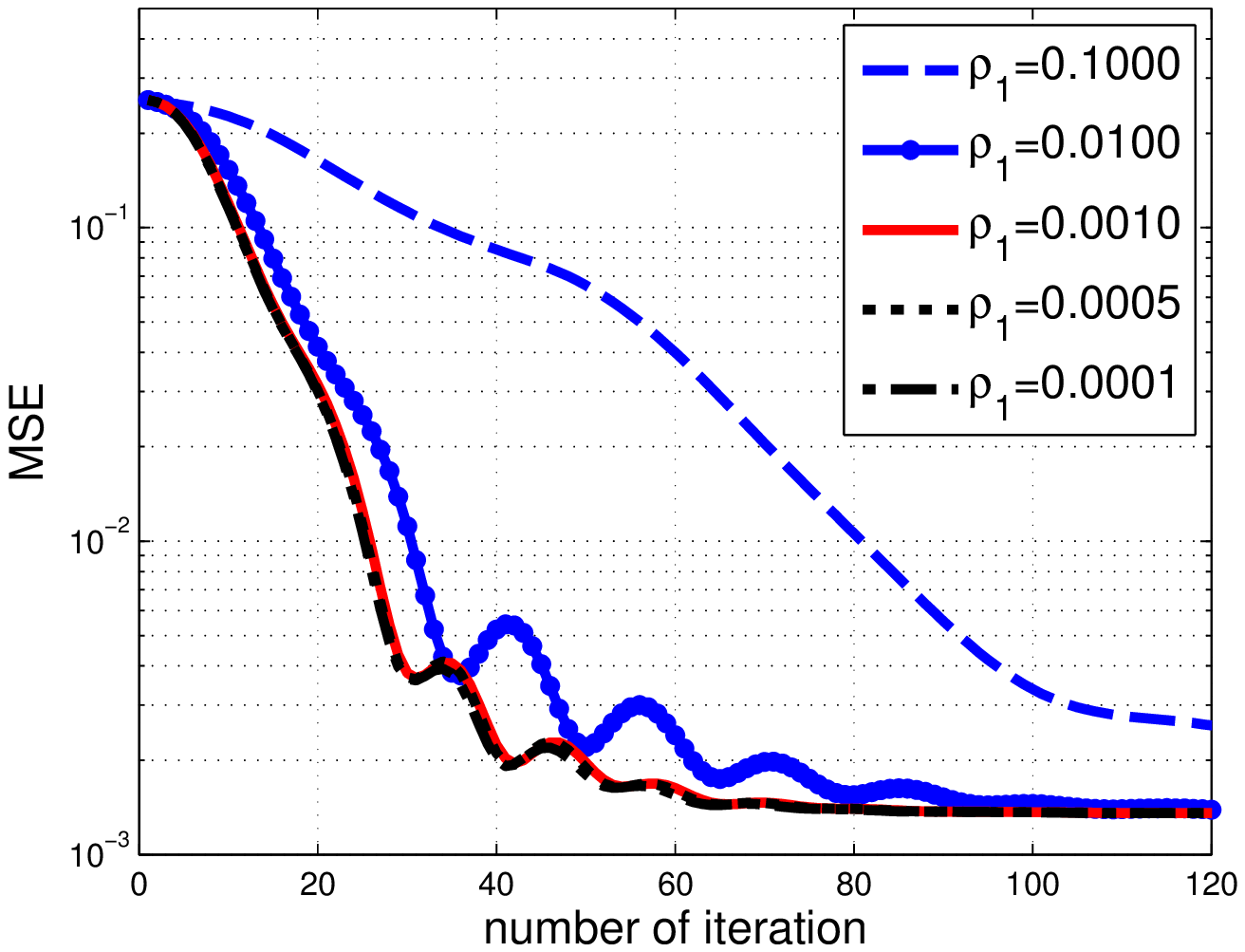} \hspace{-15pt}
&\includegraphics[width=0.24\textwidth]{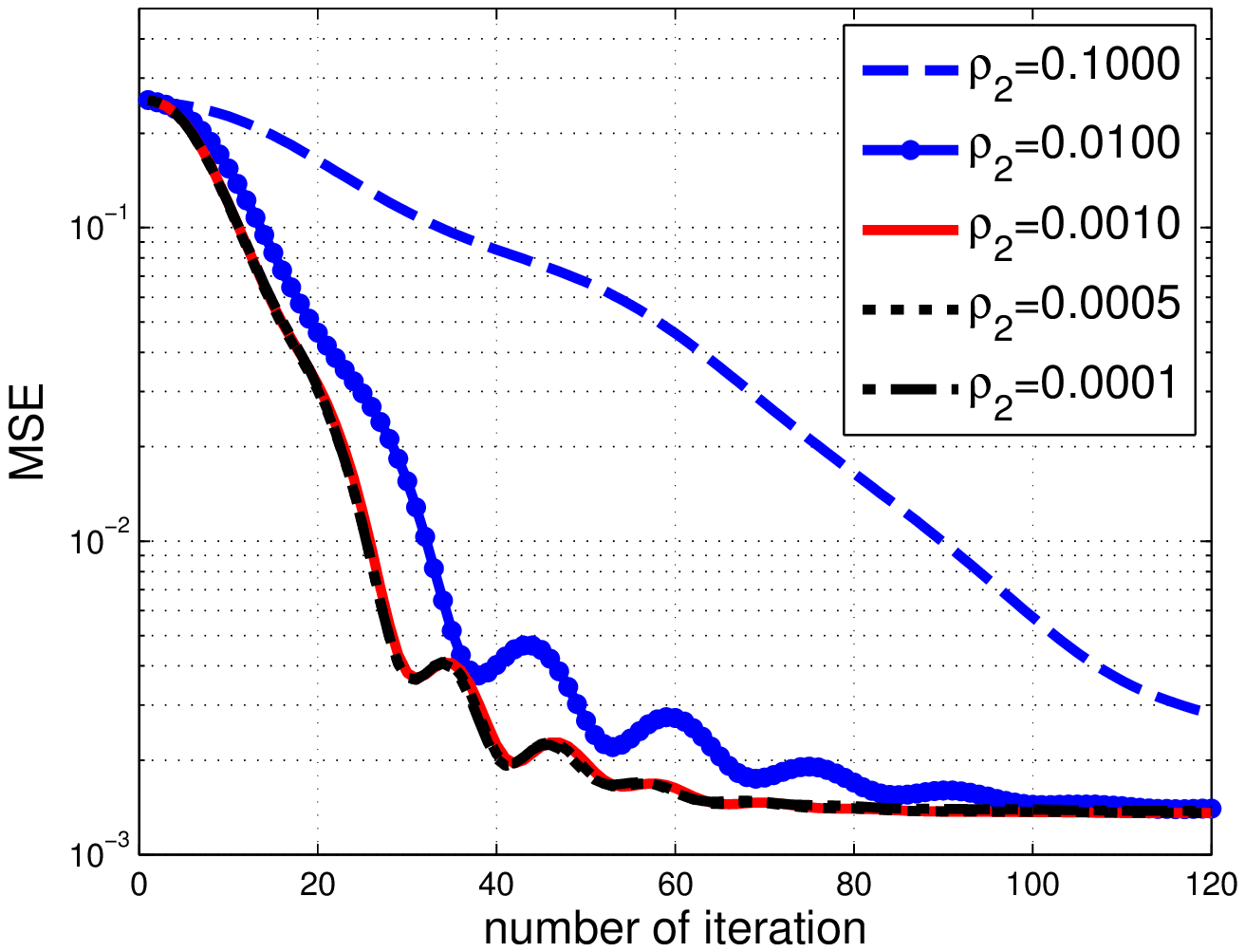} \hspace{-15pt}
&\includegraphics[width=0.24\textwidth]{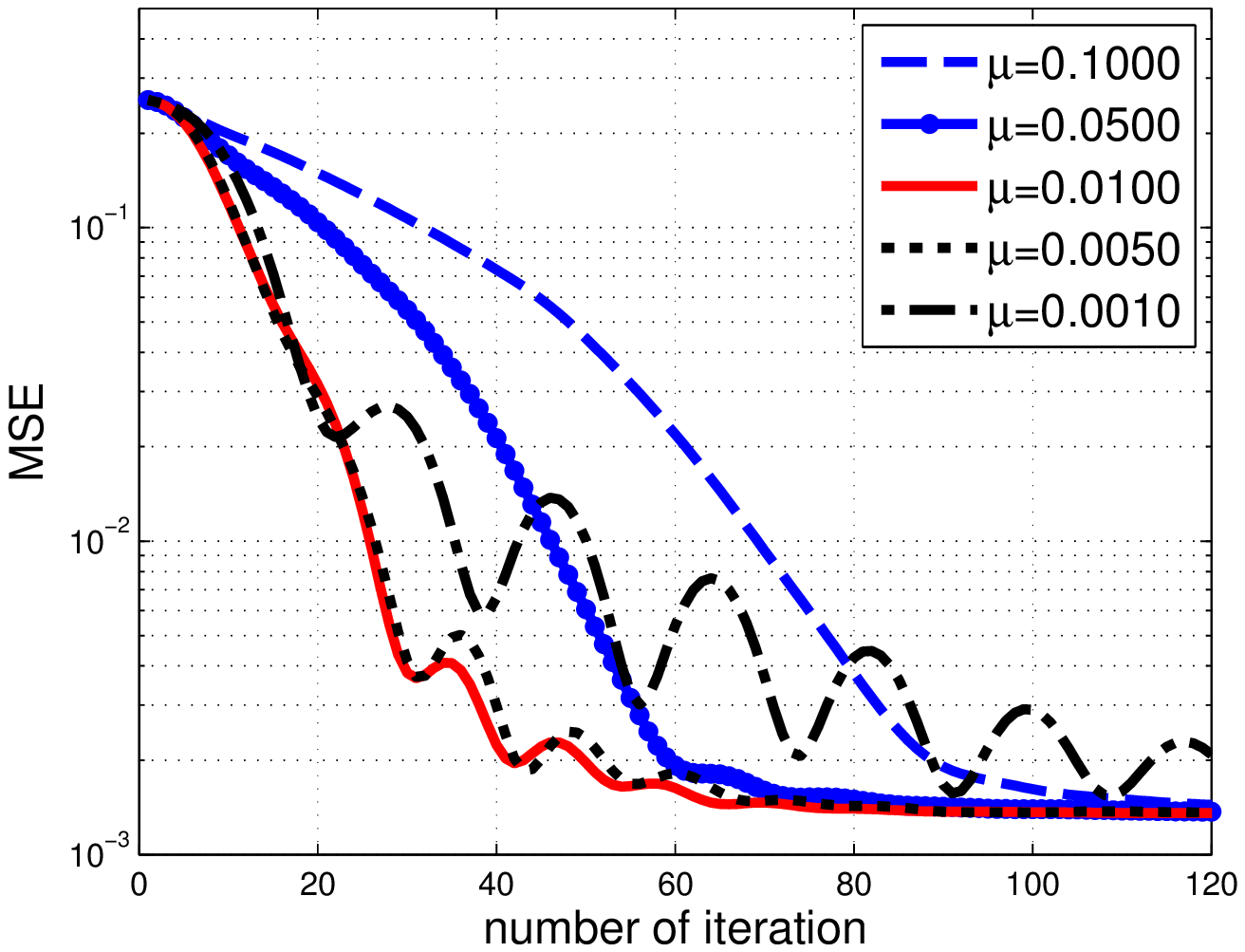} \hspace{-15pt}
&\includegraphics[width=0.24\textwidth]{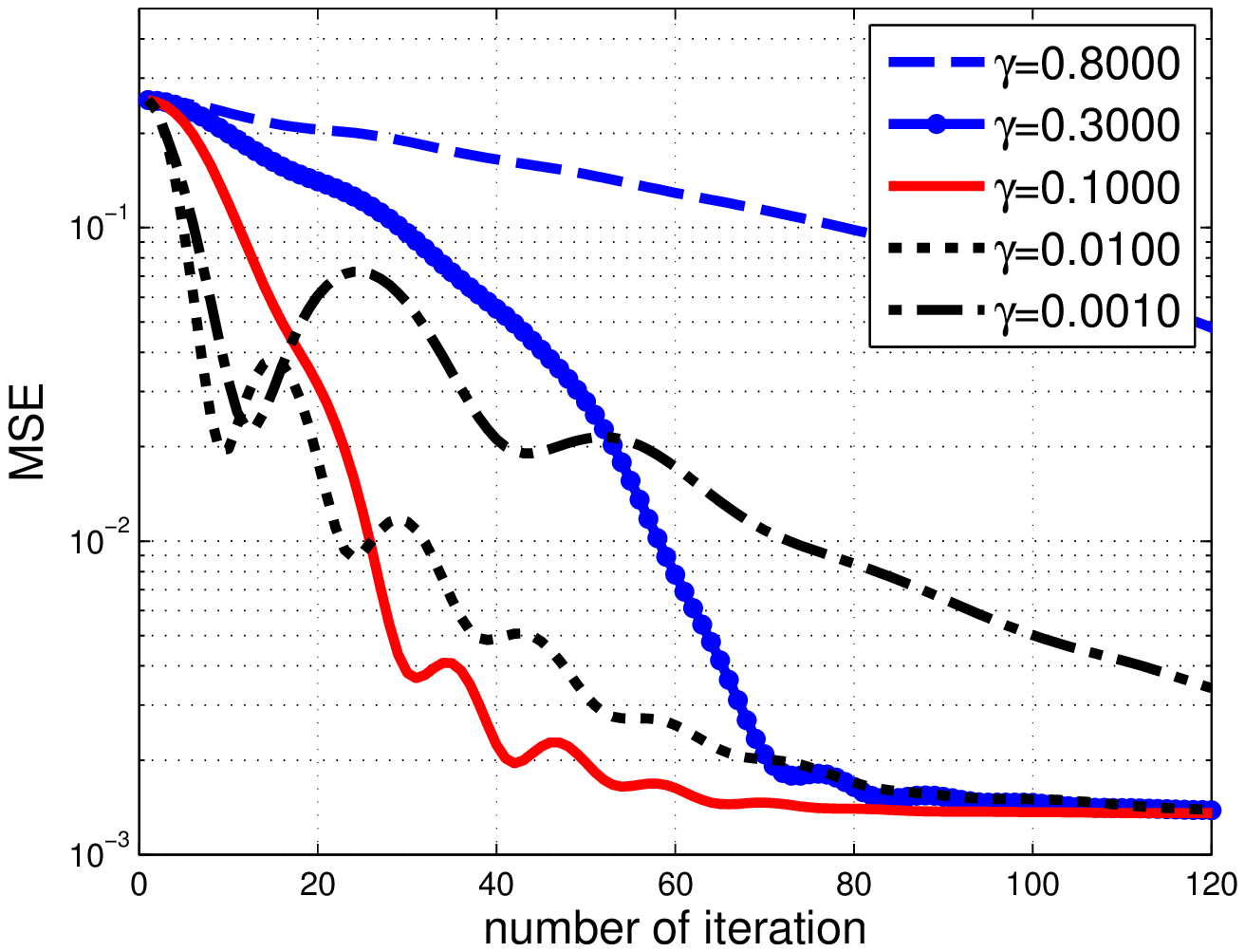}\\
$\rho_1$ \hspace{-15pt}& $\rho_2$\hspace{-15pt} & $\mu$\hspace{-15pt} & $\gamma$\\
Baby2 & & & \\
\includegraphics[width=0.24\textwidth]{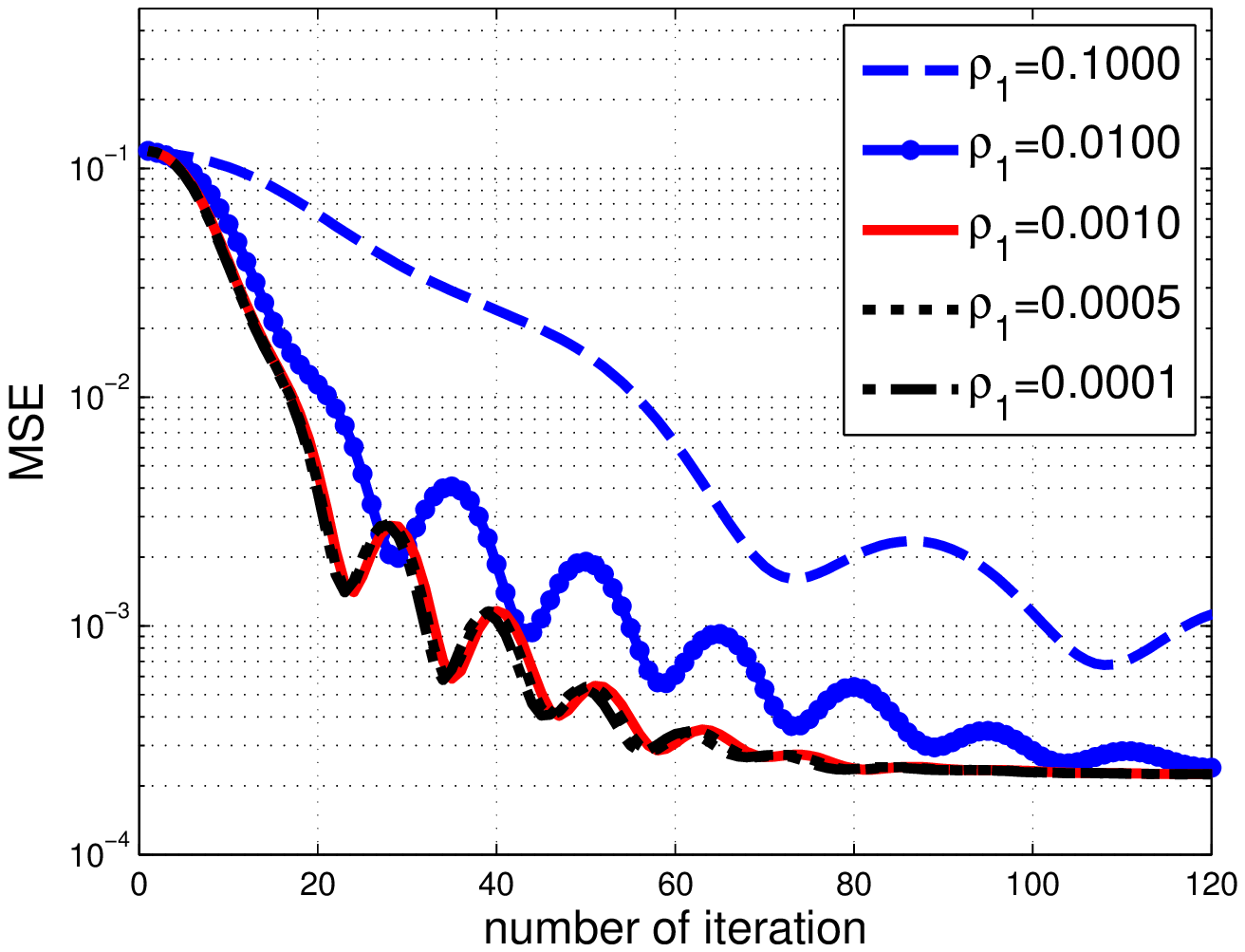}\hspace{-15pt}
&\includegraphics[width=0.24\textwidth]{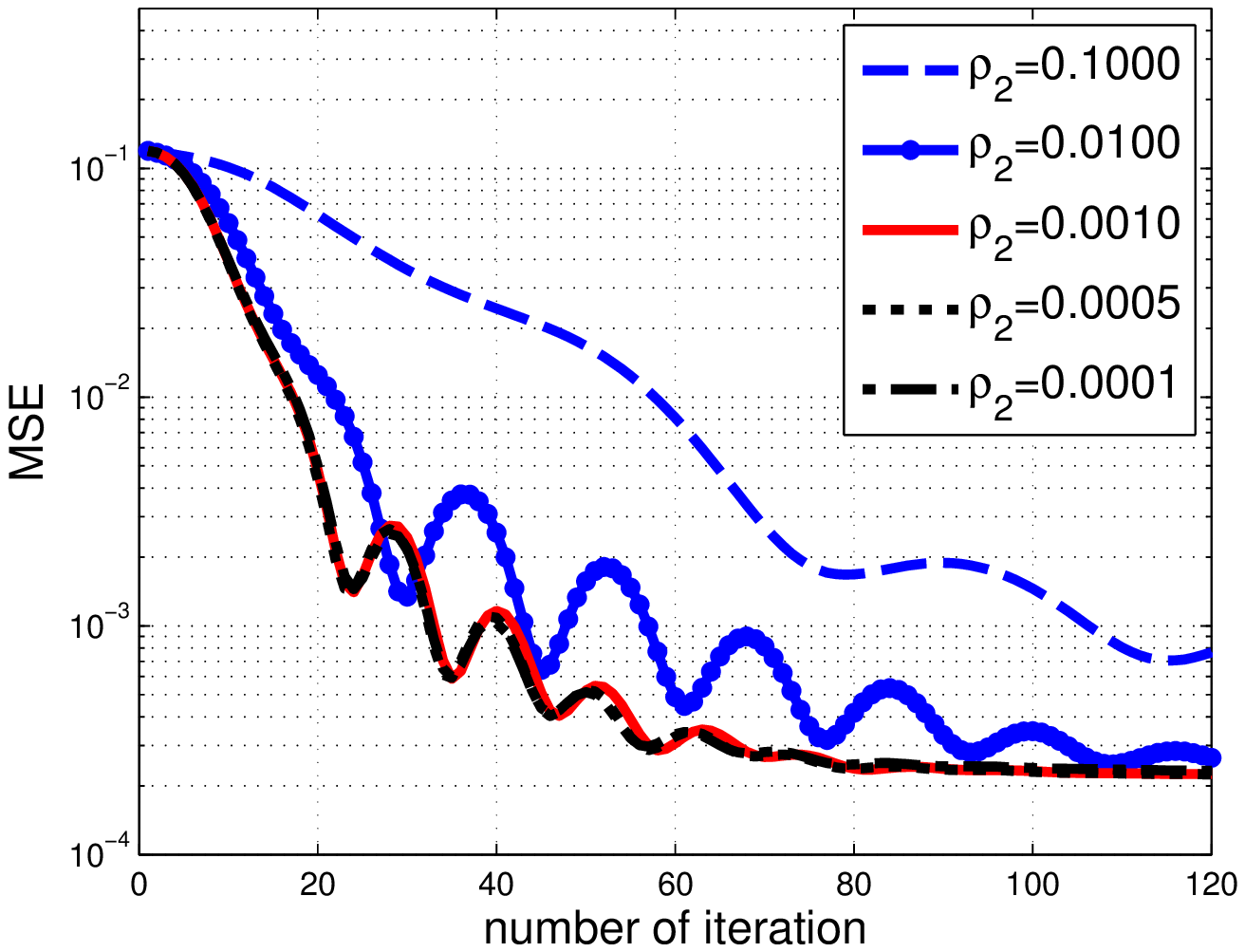}\hspace{-15pt}
&\includegraphics[width=0.24\textwidth]{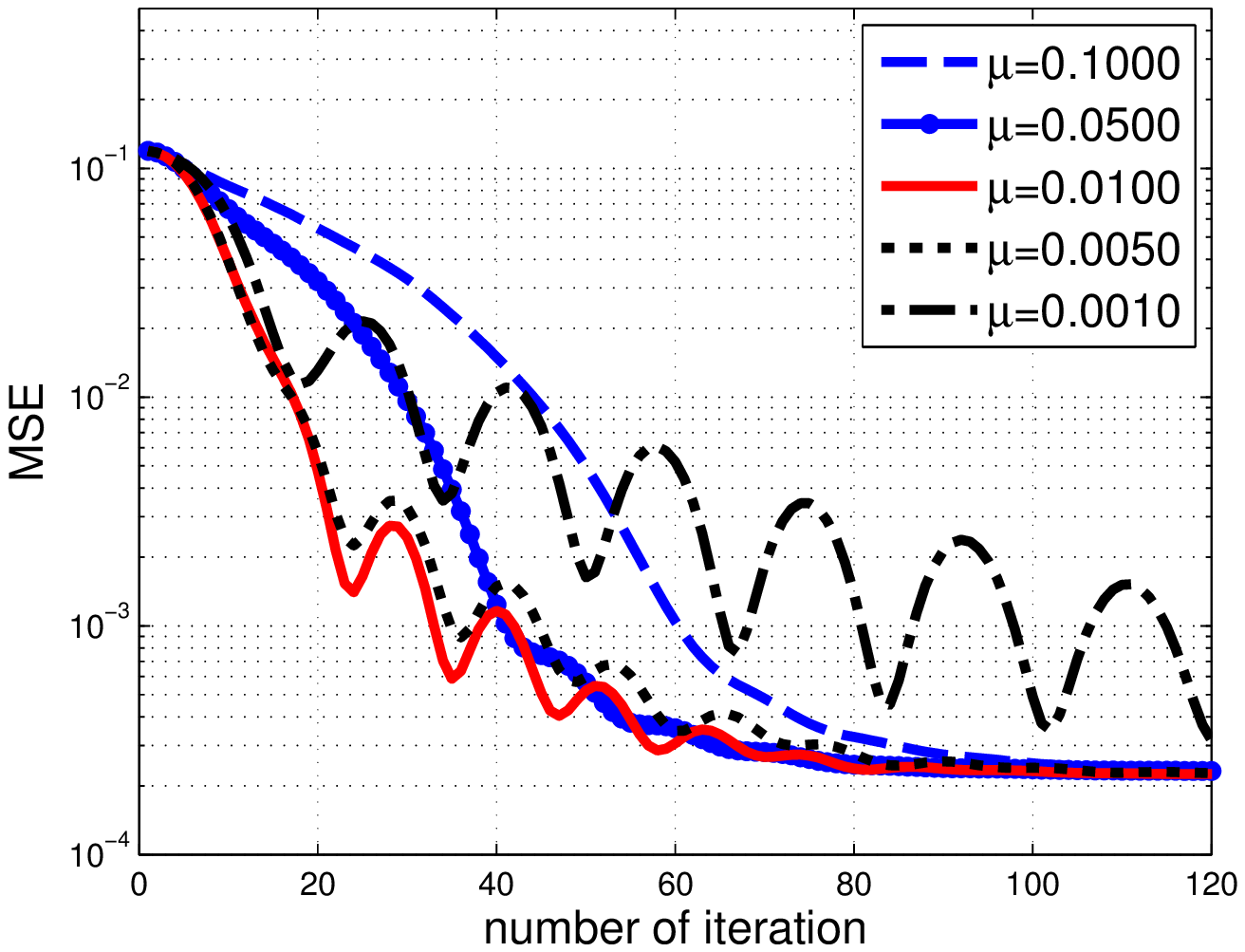}\hspace{-15pt}
&\includegraphics[width=0.24\textwidth]{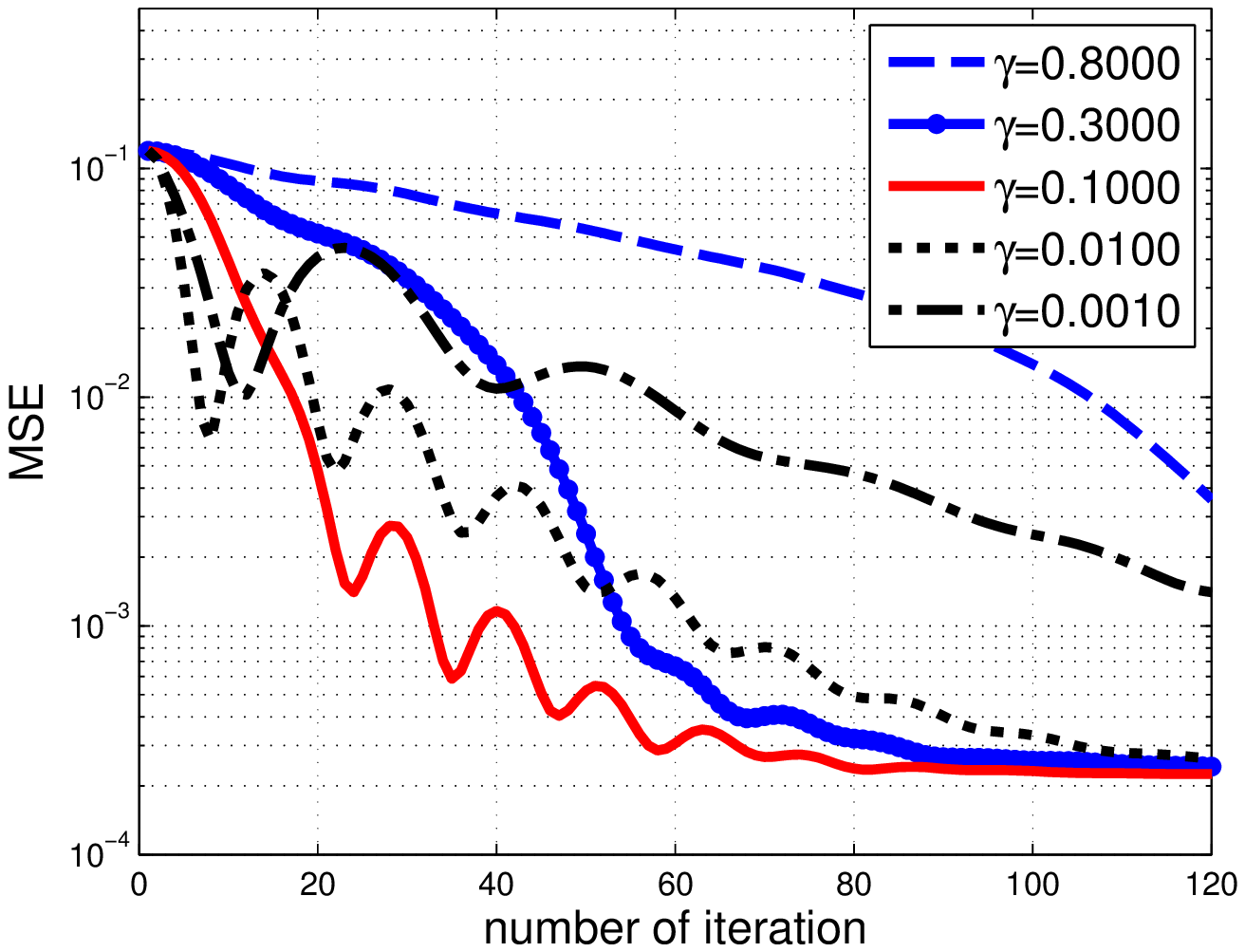}\\
$\rho_1$ \hspace{-15pt}& $\rho_2$\hspace{-15pt} & $\mu$\hspace{-15pt} & $\gamma$\\
Moebius & & & \\
\includegraphics[width=0.24\textwidth]{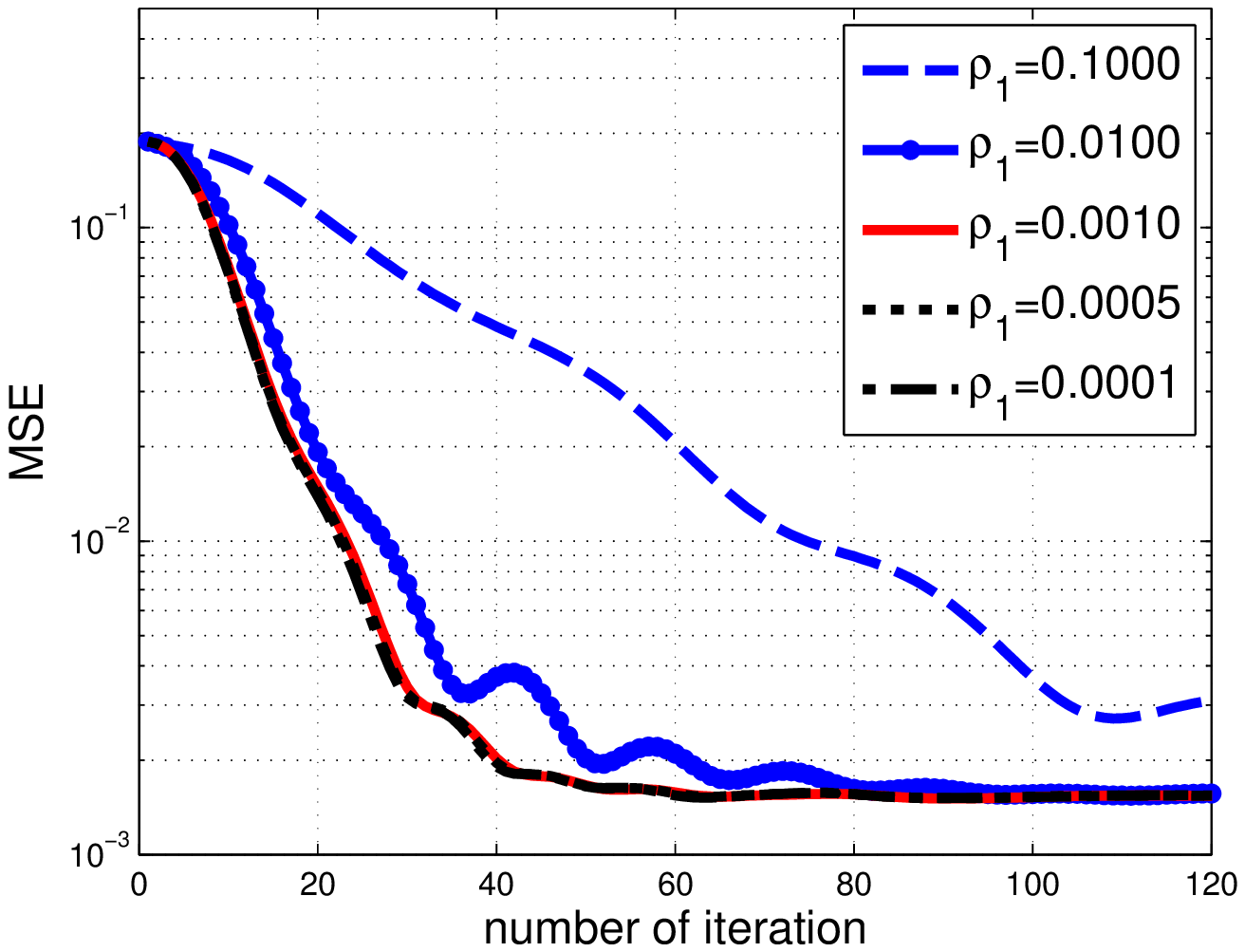}\hspace{-15pt}
&\includegraphics[width=0.24\textwidth]{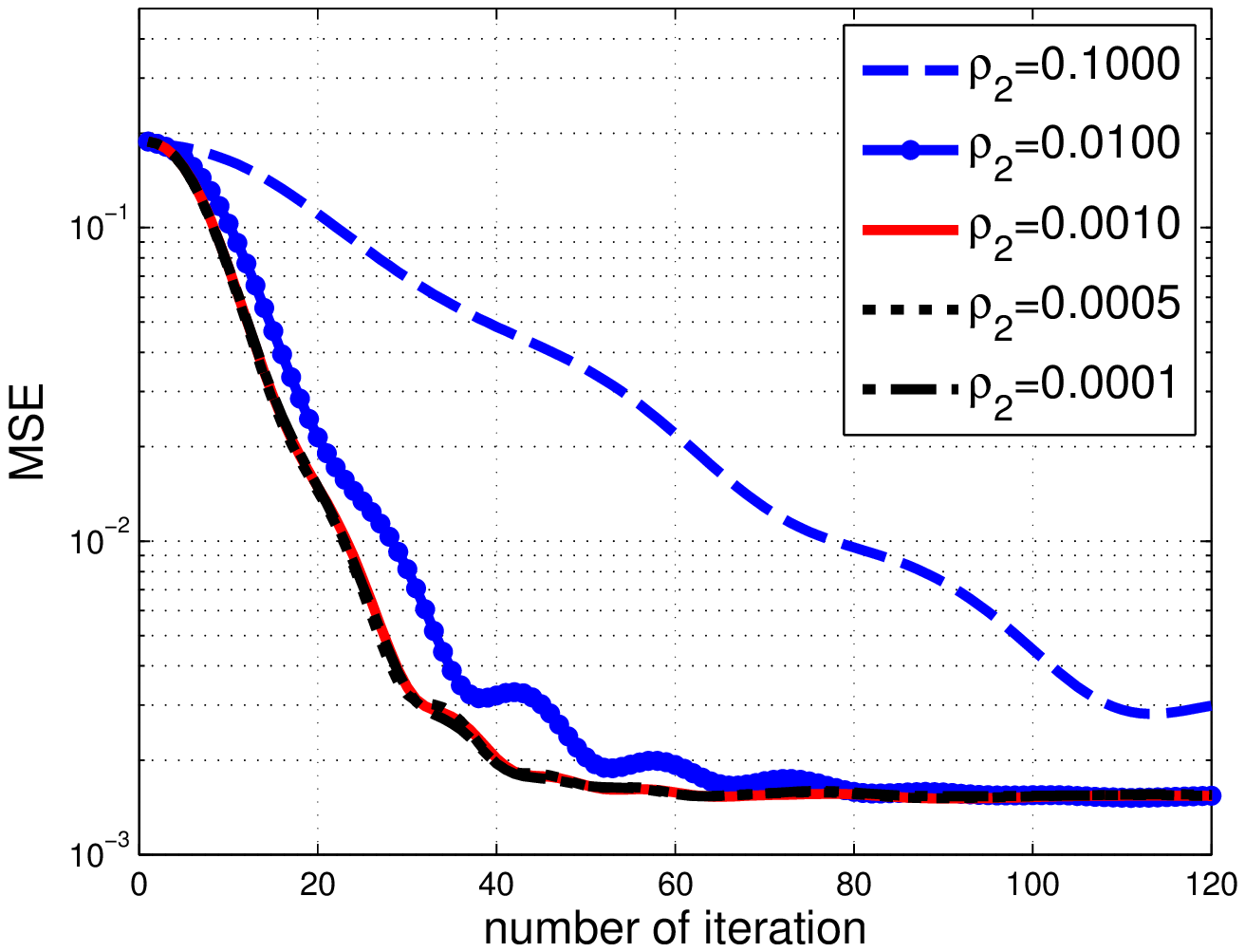}\hspace{-15pt}
&\includegraphics[width=0.24\textwidth]{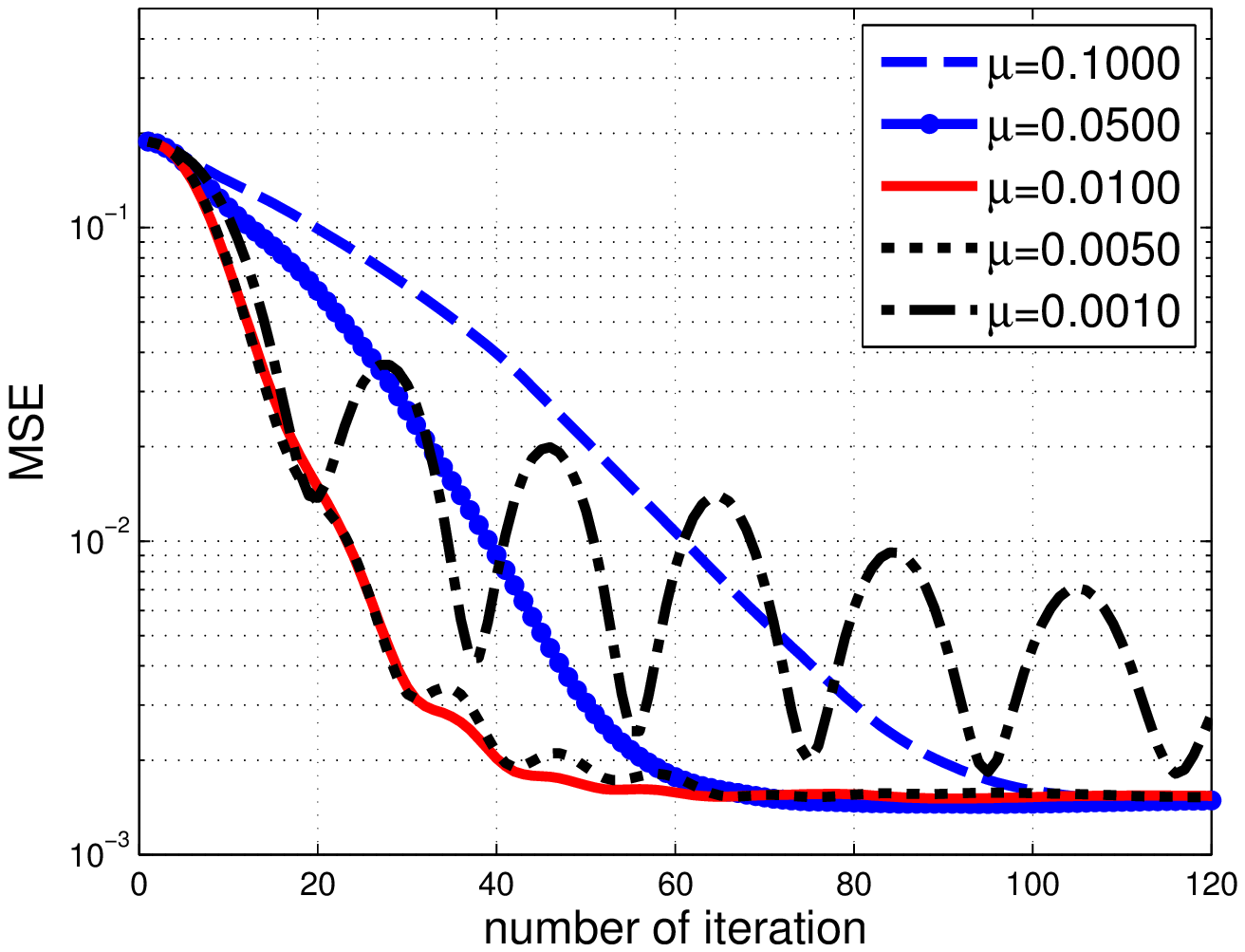}\hspace{-15pt}
&\includegraphics[width=0.24\textwidth]{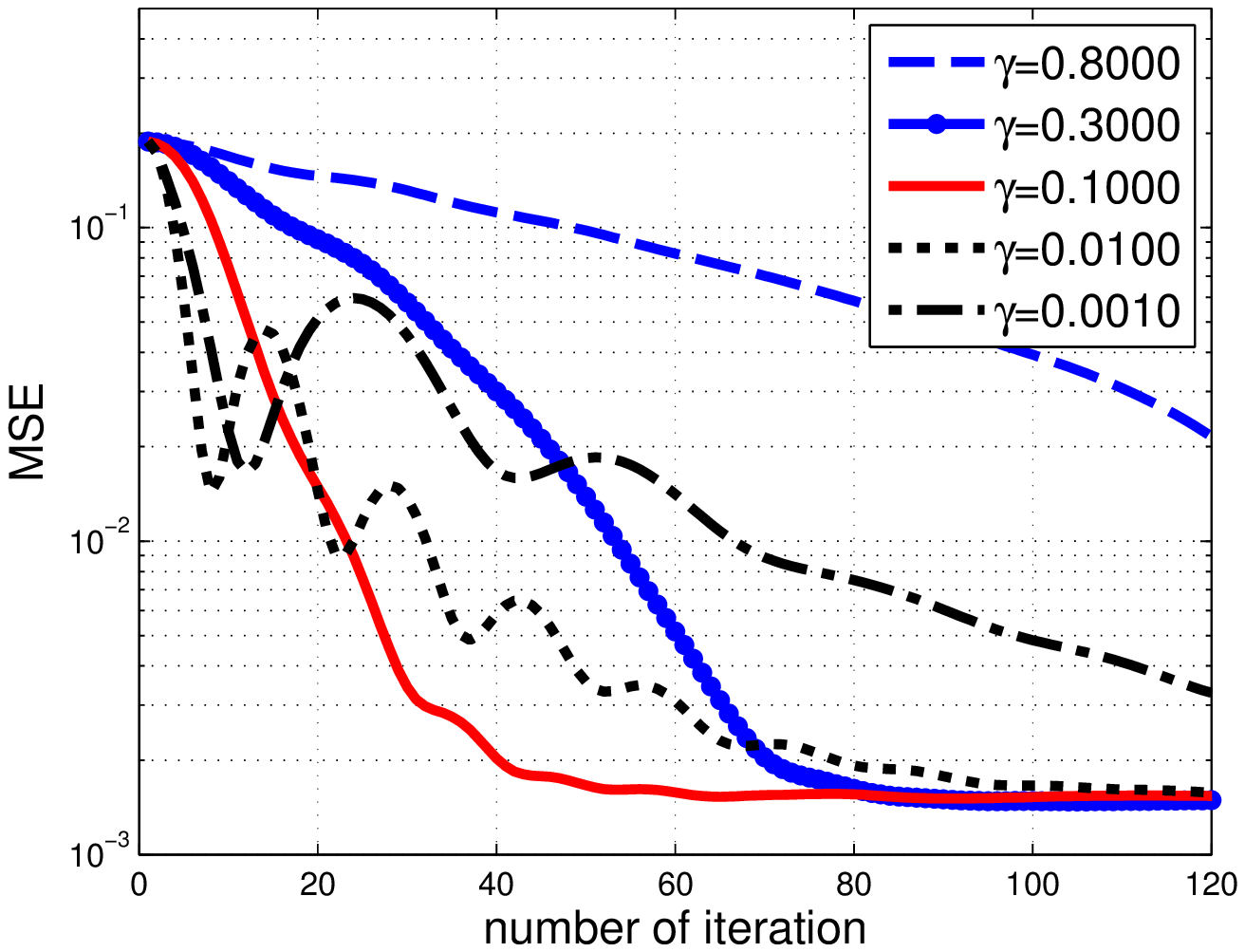}\\
$\rho_1$ \hspace{-15pt}& $\rho_2$\hspace{-15pt} & $\mu$\hspace{-15pt} & $\gamma$\\
Dolls & & & \\
\includegraphics[width=0.24\textwidth]{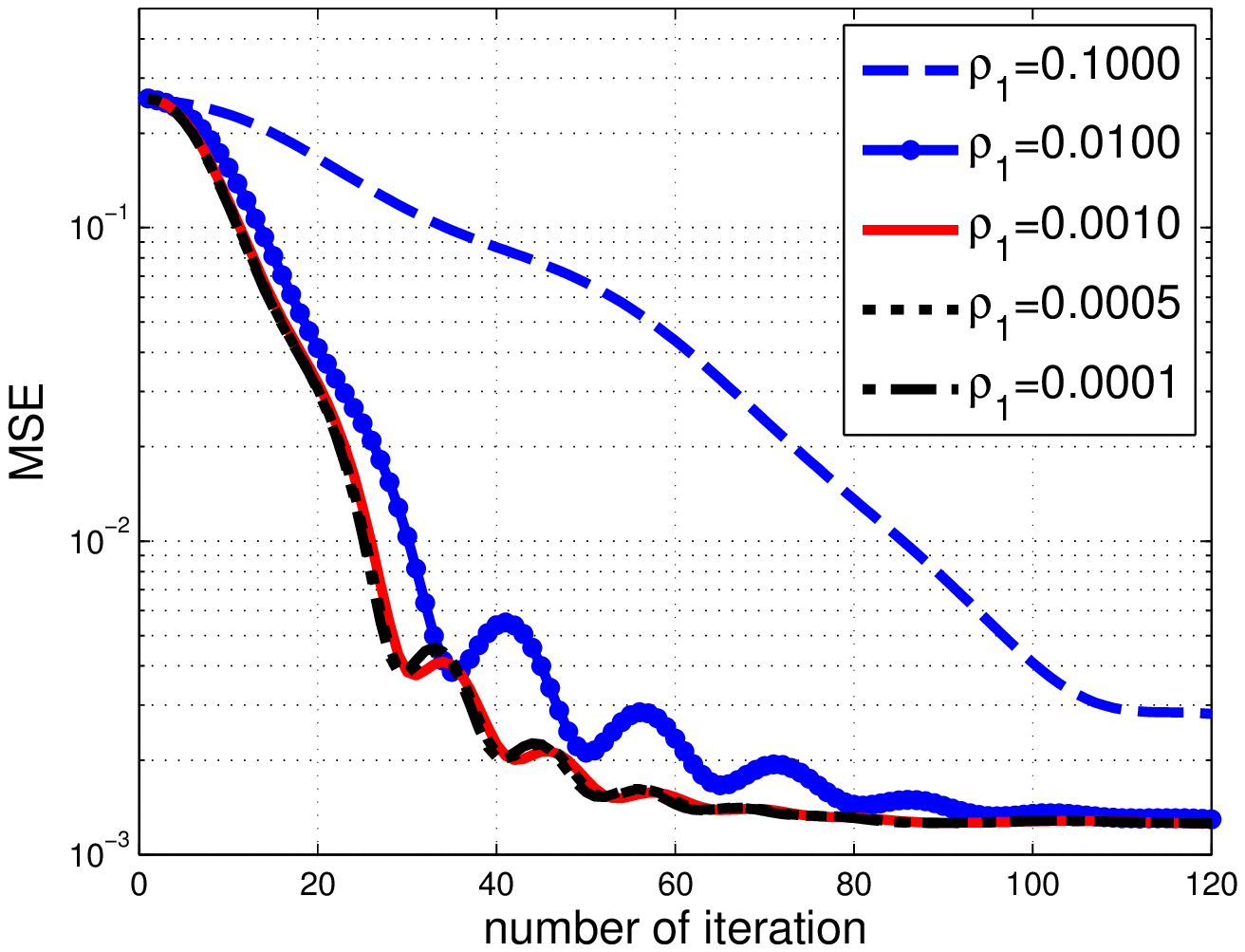}\hspace{-15pt}
&\includegraphics[width=0.24\textwidth]{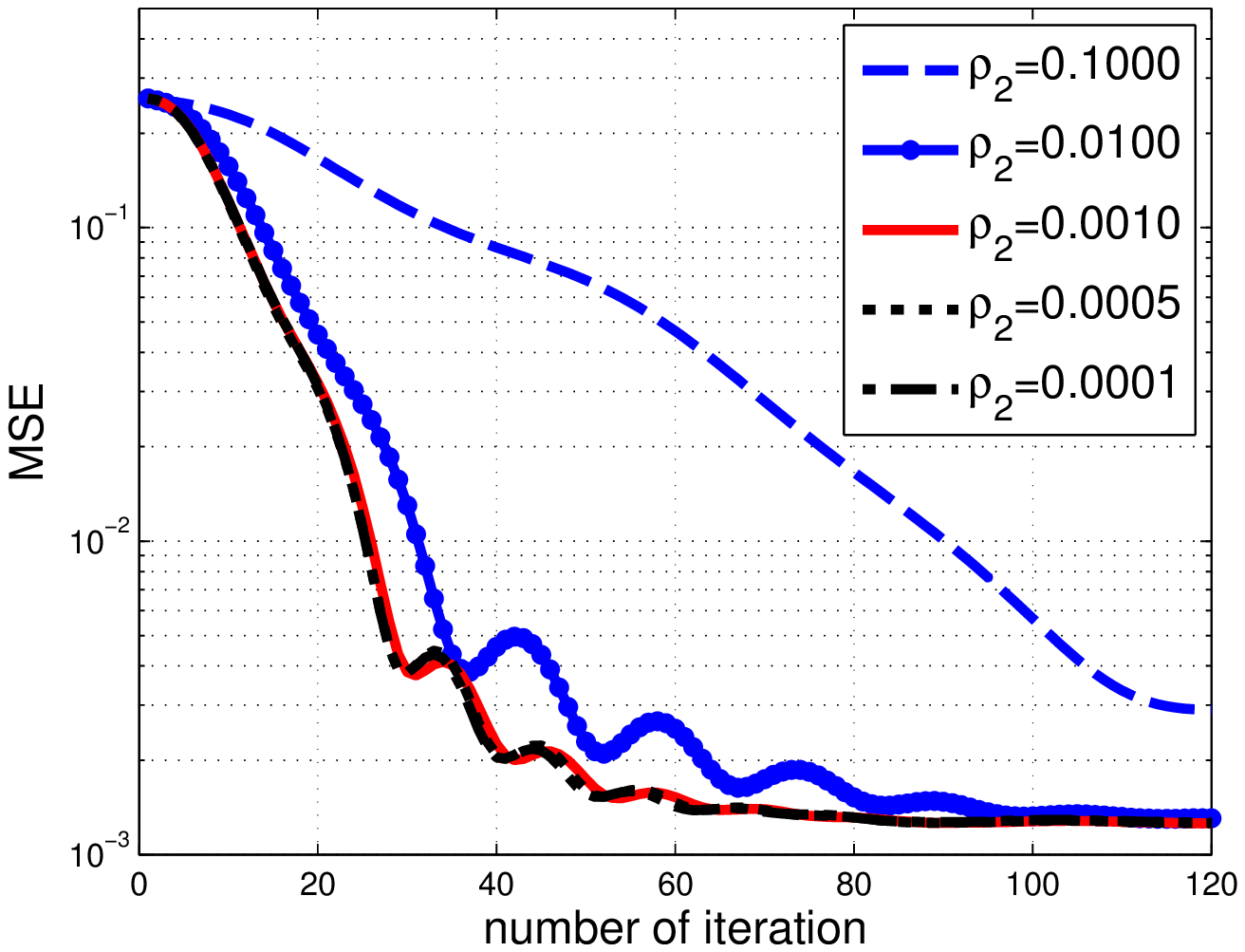}\hspace{-15pt}
&\includegraphics[width=0.24\textwidth]{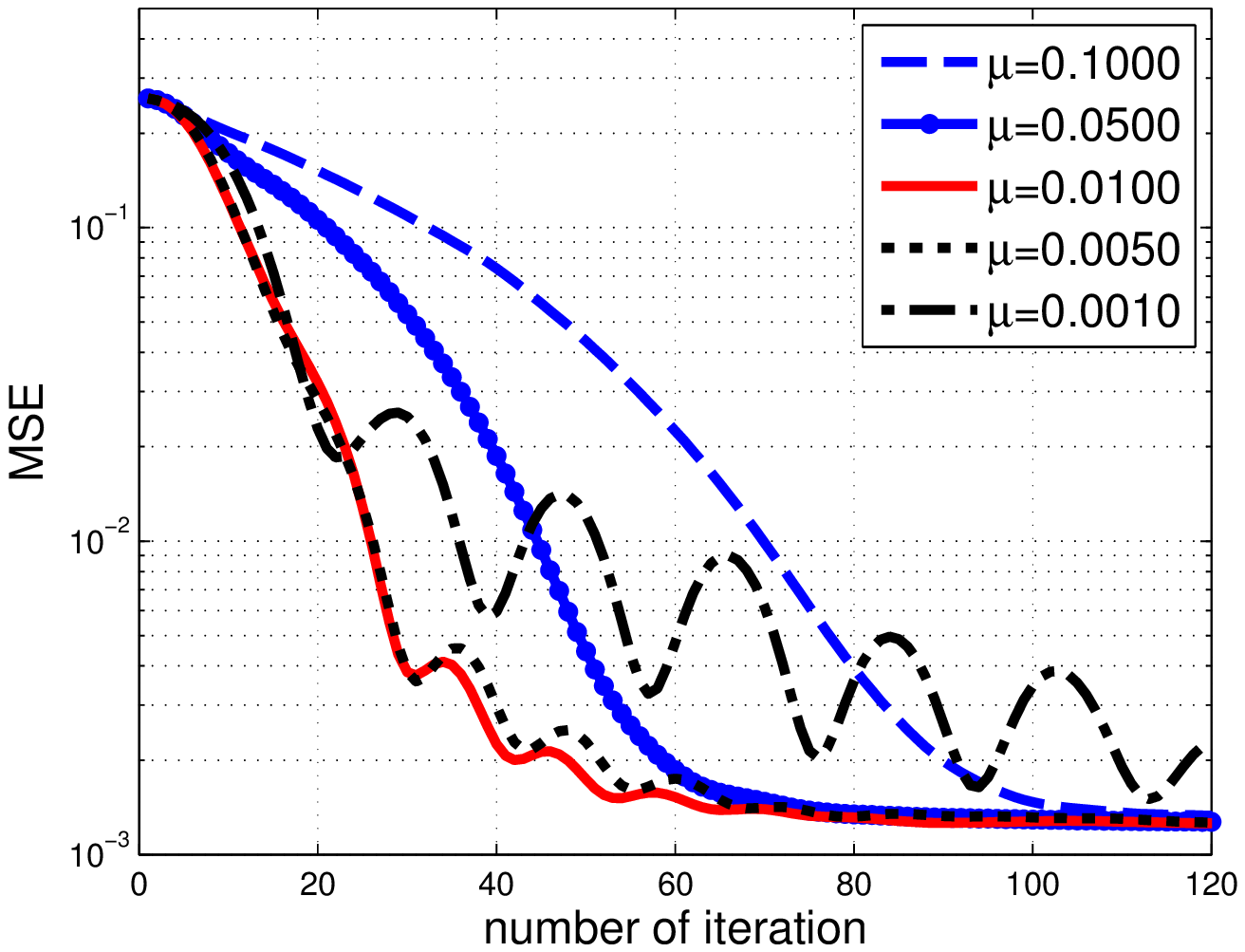}\hspace{-15pt}
&\includegraphics[width=0.24\textwidth]{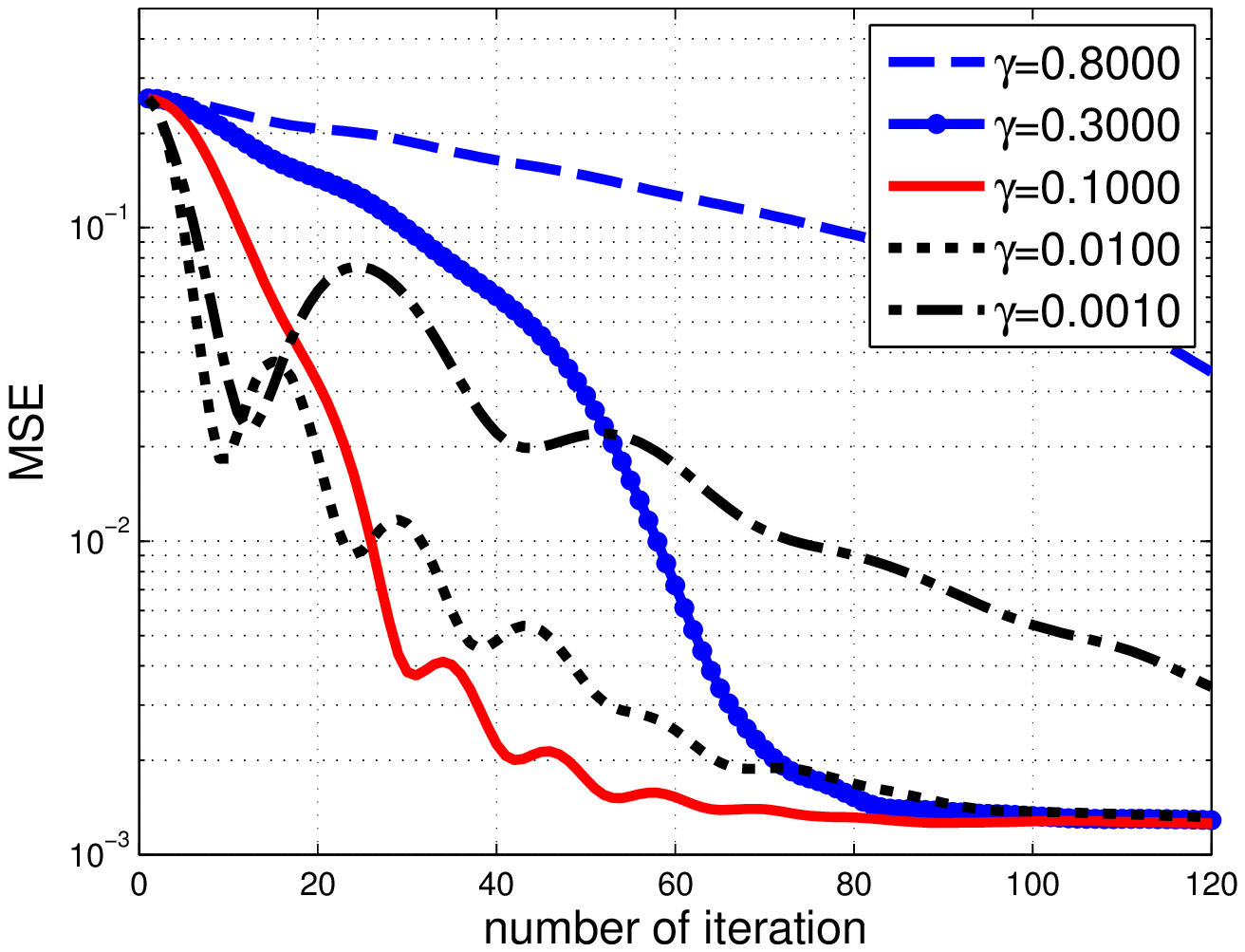}\\
$\rho_1$ \hspace{-15pt}& $\rho_2$\hspace{-15pt} & $\mu$\hspace{-15pt} & $\gamma$
\end{tabular}
\caption{MSE for $\xi = 0.1$.}
\label{fig:internal_parameter_sweep_10p_1}
\end{figure}

For $\mu,\rho_1, \rho_2, \gamma$, we conduct a set of similar experiments as before. The results are shown in \fref{fig:internal_parameter_sweep_20p_1} and \fref{fig:internal_parameter_sweep_10p_1}. The criteria to select the parameter is based on the convergence rate. This gives us $\rho_1=0.001$, $\rho_2=0.001$, $\mu=0.01$ and $\gamma=0.1$.

\newpage
\section{Summary}
We summarize our findings in \tref{table:parameters}. We remark that the values in \tref{table:parameters} are ``typical'' values that correspond to a reasonable MSE on average. Of course, for a specific problem there exists a set of optimal parameters. However, from our experience, this set of parameters seems to be robust over a wide range of problems. 

\begin{table}[h!]
\centering
\normalsize
\begin{tabular}{cll}
\hline
Parameter   & Functionality         & Values \\
\hline
$\lambda_1$ & Wavelet sparsity      &   $4\times10^{-5}$ \\
$\lambda_2$ & Contourlet sparsity   &   $2\times10^{-4}$ \\
$\beta$     & Total variation       &   $2\times10^{-3}$\\
$\rho_1$    & Half quad. penalty for Wavelet        &   $0.001$\\
$\rho_2$    & Half quad. penalty for Contourlet     &   $0.001$\\
$\mu$       & Half quad. penalty for $\vr = \vx$    &   $0.01$\\
$\gamma$    & Half quad. penalty for $\vv = \mD\vx$ &   $0.1$\\
\hline
\end{tabular}
\caption{Summary of Parameters and typical values.}
\label{table:parameters}
\vspace{-2ex}
\end{table}